\documentclass[lettersize,journal]{IEEEtran}
\usepackage{amsmath,amsfonts}
\usepackage{algorithmic}
\usepackage{algorithm}
\usepackage{array}
\usepackage[caption=false,font=normalsize,labelfont=sf,textfont=sf]{subfig}
\usepackage{textcomp}
\usepackage{stfloats}
\usepackage{url}
\usepackage{verbatim}
\usepackage{graphicx}
\usepackage{cite}
\usepackage{pifont}
\usepackage{bm}
\newcommand{\cmark}{\ding{51}}%
\newcommand{\xmark}{\ding{55}}%

\usepackage{amssymb}
\usepackage{multirow}
\DeclareMathOperator*{\argmax}{argmax}
\DeclareMathOperator*{\argmin}{argmin}

\usepackage{fontawesome5}

\usepackage{multirow}
\usepackage{tabularx}
\usepackage{longtable}
\usepackage{booktabs}
\usepackage{lscape}

\newcolumntype{R}[1]{>{\raggedleft\arraybackslash}m{#1}}
\newcolumntype{C}[1]{>{\centering\arraybackslash}m{#1}}

\usepackage{orcidlink}

\begin{document}

\title{Semantic Visual Simultaneous Localization and Mapping: A Survey on State of the Art, Challenges, and Future Directions}

\author{Thanh Nguyen Canh~$^{\orcidlink{0000-0001-6332-1002}}$~\IEEEmembership{Student Member,~IEEE}, Haolan Zhang~$^{\orcidlink{0009-0007-1742-3754}}$, Xiem HoangVan~$^{\orcidlink{0000-0002-7524-6529}}$, and Nak Young Chong~$^{\orcidlink{0000-0001-5736-0769}}$~\IEEEmembership{Senior Member,~IEEE}
\thanks{This work was supported by JST SPRING, Japan Grant Number JPMJSP2102. \textit{(Corresponding author: Thanh Nguyen Canh.)}}
\thanks{Thanh Nguyen Canh, Haolan Zhang and Nak Young Chong are with the School of Information Science, Japan Advanced Institute of Science and Technology, Nomi 923-1292, Japan (e-mail: thanhnc@jaist.ac.jp; haolan.z@jaist.ac.jp; nakyoung@jaist.ac.jp).}
\thanks{Xiem HoangVan are with the Vietnam National University, University of Engineering and Technology, Hanoi 10000, Vietnam (e-mail: xiemhoang@vnu.edu.vn).}
}

\markboth{Journal of \LaTeX\ Class Files,~Vol.~14, No.~8, August~2021}%
{Shell \MakeLowercase{\textit{et al.}}: A Sample Article Using IEEEtran.cls for IEEE Journals}


\maketitle

\begin{abstract}
Semantic Simultaneous Localization and Mapping (SLAM) is a critical area of research within robotics and computer vision, focusing on the simultaneous localization of robotic systems and associating semantic information to construct the most accurate and complete comprehensive model of the surrounding environment. Since the first foundational work in Semantic SLAM appeared more than two decades ago, this field has received increasing attention across various scientific communities. Despite its significance, the field lacks comprehensive surveys encompassing recent advances and persistent challenges. In response, this study provides a thorough examination of the state-of-the-art of Semantic SLAM techniques, with the aim of illuminating current trends and key obstacles. Beginning with an in-depth exploration of the evolution of visual SLAM, this study outlines its strengths and unique characteristics, while also critically assessing previous survey literature. Subsequently, a unified problem formulation and evaluation of the modular solution framework is proposed, which divides the problem into discrete stages, including visual localization, semantic feature extraction, mapping, data association, and loop closure optimization. Moreover, this study investigates alternative methodologies such as deep learning and the utilization of large language models, alongside a review of relevant research about contemporary SLAM datasets. Concluding with a discussion on potential future research directions, this study serves as a comprehensive resource for researchers seeking to navigate the complex landscape of Semantic SLAM.

\end{abstract}

\begin{IEEEkeywords}
Semantic SLAM, Semantic Mapping, Visual SLAM, Visual Localization.
\end{IEEEkeywords}

\section{Introduction} \label{sec:intro}

\IEEEPARstart{A}{utonomous} robotic systems play a vital role in diverse applications such as search and rescue, exploration, augmented reality, and autonomous navigation. These systems must possess a comprehensive understanding of their environment, which entails creating detailed maps, localizing themselves within these maps, and interpreting semantic information about their surroundings. Semantic SLAM addresses these challenges by integrating traditional SLAM capabilities with semantic perception, thereby enabling robots to construct detailed, high-level representations of their environment. Such semantic maps are essential for executing complex tasks with efficiency and accuracy, while balancing computational and memory constraints.

Semantic SLAM has gained increasing attention in the robotics and computer vision communities over the past decades. This interest stems from the growing demand for autonomous systems capable of operating in dynamic, unstructured, and complex environments. The field has expanded significantly, driven by advances in sensor technologies, computational capabilities, and artificial intelligence (AI). While this growth has broadened the scope of research, it has also introduced challenges related to integrating diverse methodologies and ensuring cross-disciplinary collaboration. This survey seeks to address these gaps by providing a unified perspective on Semantic SLAM, analyzing key advancements, and identifying critical research challenges.

Semantic SLAM is currently at a transformative juncture, propelled by novel developments in spatial perception and AI. Breakthroughs in deep learning have facilitated the extraction of high-level semantic features, enabling robots to recognize objects, infer relationships, and interact more intelligently with their surroundings. These advances include neural network models for beyond-line-of-sight prediction, reasoning over dynamic environments, and processing deformable scenes. By leveraging these technologies, Semantic SLAM has the potential to transcend traditional limitations and offer robust solutions for real-world applications.

However, the integration of semantic information introduces significant challenges. These include maintaining accuracy in dynamic settings, achieving real-time processing efficiency, and developing robust data fusion algorithms. Moreover, the field lacks standardized benchmarks and reproducible research practices, which are essential for evaluating and comparing different approaches. This survey aims to highlight these challenges, propose potential research directions, and emphasize the importance of interdisciplinary collaboration to advance the state-of-the-art in Semantic SLAM.

\subsection{Visual SLAM Evolution}
\label{sec:vslam}

The concept of Simultaneous Localization and Mapping (SLAM) was first introduced by Smith and Cheeseman in the 1980s \cite{smith1986representation}. In addition, since the advent of mobile robots in the late 1960s, the goal of enabling them to execute tasks autonomously has remained a central theme in robotics research~\cite{placed2023survey}. Since then, SLAM has become a cornerstone technology in robotics, enabling autonomous systems to navigate and interact with their environments. Inspired by human spatial perception—the ability to localize, map, and adapt to unfamiliar settings—robots equipped with SLAM algorithms can perform similar tasks using various sensors, such as cameras, LiDAR, and inertial measurement units (IMUs). This ability can be enhanced with acquired training and plays a crucial role in human cognition and robot control development.

SLAM technology has found applications across diverse domains, including underwater vehicles (UWVs) \cite{zhang2022visual}, unmanned aerial vehicles (UAVs) \cite{canh2024s3m}, autonomous driving \cite{cheng2022review}, service robots \cite{lee2018monocular}, augmented reality (AR) \cite{munoz2018augmented}, and virtual reality (VR) \cite{jiang2023slam}. These applications underscore the versatility and importance of SLAM in enabling autonomy.

Based on the type of sensors, traditional SLAM systems can be broadly categorized into LiDAR-based SLAM (L-SLAM) and vision-based SLAM (vSLAM). LiDAR sensors excel at precise distance measurements and high-frequency updates, making them ideal for applications requiring low-drift motion estimation \cite{hess2016real, zou2021comparative}. However, LiDAR systems are costly and computationally intensive. In contrast, vSLAM leverages cameras to capture rich visual information, allowing for feature extraction and environmental understanding at a lower cost \cite{kazerouni2022survey, campos2021orb}. Despite its advantages, vSLAM is sensitive to lighting conditions and struggles in textureless or dynamic environments. In recent days, various SLAM technologies 
have focused on vSLAM due to its low hardware cost, high accuracy in small scenes, and the ability to obtain rich environmental information. On the other hand, there are still many challenges, especially that come with dynamic object movements and environments lacking textures. 

To overcome these limitations, researchers have explored multi-sensor fusion approaches, combining cameras, IMUs, and GPS to improve localization robustness. Recent advancements in vSLAM have emphasized graph-based methodologies \cite{lu1997globally}, where the front end constructs graphs from sensor data and the back end optimizes these graphs to determine the most probable pose configuration. 

While traditional vSLAM focuses on geometric map construction and localization, it lacks semantic understanding. The integration of \textit{semantic} information—enabled by advancements in deep learning—has opened new avenues for vSLAM. By extracting semantic features, researchers have improved pose estimation, map accuracy, and environmental understanding \cite{ma2017multi, mccormac2017semanticfusion, sunderhauf2017meaningful}. Semantic vSLAM systems enable robots to perform tasks such as semantic localization and mapping, significantly enhancing their adaptability to complex and dynamic environments. The rise of these systems represents a paradigm shift, bridging the gap between geometric and semantic approaches in SLAM. In tandem with its geometric foundation, SLAM has steadily evolved to incorporate higher-level information about the environment—termed \emph{semantic information}—giving rise to the field of \emph{Semantic SLAM}.

Early SLAM solutions largely centered on geometric cues, such as points, lines, and planar structures, using various sensor modalities like cameras or LiDAR for feature extraction \cite{chen2025semantic, kazerouni2022survey}. This approach, often referred to as traditional or \emph{geometric SLAM}, excels when the environment remains static and well-defined. However, real-world settings inherently involve illumination changes, dynamic obstacles, and complex scenes with sparse textures. These conditions can undermine the robustness and accuracy of purely geometric methods, which typically rely on handcrafted features or geometric primitives. Consequently, the research community has increasingly focused on more holistic solutions that integrate vision sensors ({\textit{e.g.}, monocular, stereo, RGB-D cameras) and other sensor data, thereby enhancing SLAM’s adaptability and reliability across diverse and challenging operational domains \cite{placed2023survey, canh2022multisensor, chen2025semantic}.

In parallel with these sensor-fusion strategies, breakthroughs in deep learning have led to significant leaps in \emph{semantic understanding}. By leveraging neural networks for object detection, semantic segmentation, and scene recognition, SLAM systems can move beyond constructing raw geometric maps to encoding rich descriptive information about encountered objects and scene layouts~\cite{dellaert2004semantic, sunderhauf2017meaningful}. This synergy—combining geometric estimation with semantic cues—paves the way for systems that can better handle moving objects, reinterpret ambiguous features, and enable robust long-term mapping in dynamic or partially known environments~\cite{ma2017multi, mccormac2017semanticfusion}. Notably, \emph{Semantic SLAM} not only locates and maps entities in the scene but also categorizes and attributes them with meaningful labels, thereby supporting higher-level tasks such as scene understanding, path planning, human–robot interaction, and task-oriented manipulation. In instances where the aim is to enhance localization, mapping, or both, these challenges are known as semantic localization, semantic mapping, and Semantic SLAM, respectively.

\textit{Semantic Mapping} augments traditional SLAM by embedding high-level contextual and categorical information about the environment into the spatial map, enabling a shift from purely geometric representations to cognitively meaningful scene understanding. This concept was first introduced by Dellaert and Bruemmer~\cite{dellaert2004semantic}, highlighting the importance of associating semantic entities—such as object classes, scene layouts, and terrain types—with spatial locations. Unlike conventional occupancy grids or point clouds, semantic maps provide symbolic abstractions that facilitate complex reasoning, task planning, and human-robot interaction~\cite{zhao2019lidar, jin2021semantic, canh2023object, canh2024s3m}. Recent advances in deep learning, particularly in semantic segmentation and instance-level object detection, have significantly improved the granularity and accuracy of semantic labels. This has enabled near-real-time semantic map construction, even in dynamic environments, using data from RGB, RGB-D, or multimodal sensors. Techniques such as Bayesian fusion of per-frame segmentation~\cite{mccormac2017semanticfusion} and learned feature embeddings for instance tracking~\cite{mccormac2018fusion++} have further enhanced map coherence and consistency over time.

However, despite its potential, semantic mapping introduces several open challenges that hinder scalable deployment in real-world robotics. One major concern is the increased computational complexity associated with maintaining and updating semantic labels, especially under limited hardware and power constraints~\cite{placed2023survey, chen2025semantic, hempel2022online}. Moreover, inaccurate or incomplete semantic predictions can introduce drift and inconsistency into the SLAM pipeline, particularly when these labels are used for loop closure or data association. Handling dynamic objects adds further complexity, as semantic information must be continuously validated and updated to prevent map corruption. Another critical limitation lies in the scarcity of standardized benchmarks and evaluation protocols focused specifically on the quality and consistency of semantic maps. This gap restricts fair comparison and reproducibility across approaches~\cite{dellaert2004semantic}. Additionally, memory-efficient representations, such as OctoMap-based semantic trees~\cite{ruan2022semantic, liu2023data} or voxel hashing~\cite{li2022vox, matez2024voxeland}, must balance rich semantic encoding with storage constraints, especially for long-term autonomy or large-scale environments.

\textit{Semantic Localization} aims to enhance pose estimation accuracy and robustness by incorporating high-level semantic information such as object classes, spatial relationships, and scene context, moving beyond low-level geometric features. This paradigm enables more reliable localization in perceptually ambiguous or dynamically changing environments, where traditional SLAM approaches based on handcrafted features often degrade. Semantically-aware localization leverages advances in deep neural networks (DNNs) for object detection, semantic segmentation, and scene classification to extract robust, discriminative features that are invariant to viewpoint and appearance changes. Early works such as SuMa++~\cite{zaganidis2018integrating, chen2019suma++} integrated semantic segmentation into the SLAM pipeline using a semantic ICP framework, demonstrating improved alignment accuracy in cluttered and unstructured environments. Probabilistic approaches have also been explored, where semantic information augments traditional observation models. For instance, extensions of the Markov localization framework~\cite{yi2009active, atanasov2016localization} incorporated object-level semantics into the sensor likelihood model to improve robustness against perceptual aliasing. 

A significant limitation of these early models lies in their deterministic treatment of semantic observations, which disregards the inherent uncertainty in perception systems. To mitigate this, Akai \textit{et al.}~\cite{akai2020semantic} proposed a Bayesian localization approach that integrates supervised object recognition into a probabilistic graphical model. Their framework models classification uncertainty using a Dirichlet distribution, allowing for a more principled fusion of semantic cues. Subsequently, Akai~\cite{akai2022mobile} introduced a probabilistic treatment of depth regression errors, improving pose inference in scenes with noisy depth observations. 

Recent efforts~\cite{toft2018semantic, schonberger2018semantic, xiao2019dynamic, wu2022learning} have focused on leveraging dense semantic maps to enhance global consistency and robustness under appearance changes. These approaches often encode semantic priors into factor graph-based SLAM systems or utilize metric-semantic fusion for long-term localization. Nevertheless, many of these methods still depend on planar scene assumptions and handcrafted features, which limit their generalization to complex, 3D-structured environments. As highlighted in~\cite{zhang2025bev}, such limitations can result in projection inconsistencies and scale-dependent feature degradation, underscoring the need for learning-based spatial representations capable of modeling geometric and semantic uncertainty jointly.

\textit{Semantic SLAM} unifies the principles of semantic mapping and semantic localization into a cohesive framework, enabling intelligent robotic systems. By fusing low-level geometric data with high-level semantic cues, Semantic SLAM systems can achieve more accurate and robust localization based on object-level features, generate high-resolution and context-aware maps, and maintain compact and efficient storage representations. These systems exhibit improved robustness to occlusions, lighting variations, and viewpoint changes, while also facilitating semantic-level interaction with dynamic and unstructured environments~\cite{yang2019cubeslam, bowman2017probabilistic}. Recent works have demonstrated the utility of Semantic SLAM for active exploration~\cite{asgharivaskasi2021active}, object-centric localization~\cite{chen2020sloam}, and robust relocalization under appearance and structural changes~\cite{miller2022stronger}. In general, Semantic SLAM can be decomposed into four primary sub-problems that define its core functionality:

\begin{enumerate} 
    \item \textbf{Semantic extraction and mapping}: Leveraging semantic segmentation, object detection, and scene parsing to extract meaningful features and incrementally integrate them into a globally consistent map. 
    \item \textbf{Semantic data association}: Establishing correspondences between semantic entities across time and viewpoints, often using probabilistic models, learned feature descriptors, or graph-based optimization to maintain label consistency. 
    \item \textbf{Semantic uncertainty and representation}: Modeling perceptual uncertainty from semantic predictions (\textit{e.g.}, classification confidence, depth variance) using probabilistic frameworks such as Bayesian filters, Dirichlet distributions, or Gaussian Mixtures to ensure reliable integration. 
    \item \textbf{Semantic cost function}: Designing loss functions or objective terms that encode semantic priors and constraints—such as object geometry, label consistency, or topology—for use in optimization-based SLAM back-ends. 
\end{enumerate}

Finally, \textit{Semantic SLAM} extends traditional SLAM by simultaneously estimating the 3D geometry of a scene while attaching semantic labels to observed objects and structures. This unified framework integrates both metric and symbolic understanding of the environment, enabling robots not only to map where things are but also what they are. One of the earliest systems to leverage both spatial and semantic representations was proposed by Galindo \textit{et al.}~\cite{galindo2005multi}, which introduced a layered cognitive map architecture to incorporate conceptual knowledge into the mapping process. Contemporary Semantic SLAM approaches typically extend the SLAM state vector to include semantic entities or incorporate semantic constraints into the optimization process. Some works introduce novel object-level representations, such as ellipsoids (quadrics)\cite{nicholson2018quadricslam}, planar landmarks\cite{hosseinzadeh2019structure}, or mesh models~\cite{wang2022drg}, allowing compact and expressive 3D scene encoding. Others focus on semantic data association strategies to maintain map consistency under ambiguity and perceptual noise. Probabilistic models based on maximum likelihood estimation~\cite{bowman2017probabilistic, doherty2019multimodal}, max-mixture frameworks~\cite{doherty2020probabilistic}, and multi-hypothesis k-assignment methods~\cite{michael2022probabilistic} have been developed to handle uncertain object detection and multi-instance association. Approaches can also be categorized based on how semantics are modeled: explicitly through symbolic labels or object poses, or implicitly via learned feature embeddings or volumetric priors. While explicit~\cite {hornung2013octomap, pagliari2014kinect, canh2024s3m} representations facilitate interpretability and planning, but they often demand increased storage and suffer from brittleness under perceptual uncertainty. Implicit representations, learned through DNNs~\cite{zhu2022nice, yang2022vox}, offer improved robustness but pose challenges for explainability and optimization within SLAM back-ends.

Recently, 3D Gaussian Splatting has emerged as a promising representation for capturing scene geometry and appearance. By modeling the scene as a collection of anisotropic 3D Gaussians, this approach supports continuous view synthesis and efficient rendering while offering potential as a compact SLAM representation. Its integration into SLAM frameworks remains an active area of research, with early efforts exploring its suitability for semantic reconstruction and localization~\cite{matsuki2024gaussian, yan2024gs, li2024sgs, li2025hier}. A detailed comparison of these Semantic SLAM systems is presented in Table~\ref{tab:vslam}, summarizing core characteristics such as sensor modality, representation type, semantic modeling technique, back-end architecture, and suitability for dynamic or large-scale environments.

\subsection{Previous Works}

Several recent surveys~\cite{kostavelis2015semantic, sualeh2019simultaneous, garg2020semantics, wen2021semantic, lai2022review, chen2022overview, chen2025semantic, pu2023visual, bavle2023slam, wang2024survey, georgevichsemantic, song2025semantic} have investigated the progression of visual SLAM, particularly highlighting its evolution from traditional geometric approaches to those that incorporate deep learning and semantic scene understanding. However, the majority of these studies emphasize the importance and the limitations of classical vSLAM, especially in dynamic environments, and the potential of semantic cues to enhance scene interpretation. Few offer a systematic breakdown of Semantic SLAM as a unified framework, and most do not examine its subcomponents in detail, such as semantic data association or loop closure optimization. Table~\ref{tab:surveypaper} summarizes the themes covered by each study and contrasts them with the scope of this survey.

The earliest among them, by Kostavelis and Gasteratos~\cite{kostavelis2015semantic}, offers a comprehensive survey on “semantic mapping for mobile robots.” This work analyzes the trends and main aspects of semantic mapping and identifies three core research challenges: (1) defining the minimal criteria for a map to be considered semantic, (2) developing semantic mapping evaluation metrics, and (3) integrating semantic maps for knowledge representation. However, this review predominantly focuses on augmenting traditional 2D or 3D maps (\textit{e.g.}, topological maps) with semantic labels and does not address components like semantic localization, data association, or back-end optimization. More recent works~\cite{lai2022review, chen2022overview, bavle2023slam} acknowledge the limitations of geometric SLAM and advocate for deep learning as a data-driven paradigm for sensor interpretation and scene understanding. These surveys trace the chronological evolution of SLAM technologies and paradigms but stop short of delving into the architecture or mathematical modeling of Semantic SLAM systems. Other surveys~\cite{garg2020semantics, georgevich_ferreira_semantic_2025, song2025semantic} approach semantic mapping as a strategy to improve perceptual awareness in autonomous systems. These papers typically divide the semantic mapping pipeline into three stages: data acquisition, semantic and spatial fusion, and symbolic knowledge representation. However, they generally treat mapping and perception as decoupled modules rather than parts of an integrated SLAM framework.

Sualeh \textit{et al.}\cite{sualeh2019simultaneous} introduces the fundamentals of SLAM and provides a high-level overview of Semantic SLAM. Their work is introductory and does not explore a formalized model or detailed submodules. Similarly, Pu \textit{et al.}\cite{pu2023visual} highlight the challenges of vSLAM in dynamic environments, reviewing how deep learning improves the front-end (\textit{e.g.}, dynamic object segmentation, feature robustness under varying illumination) and the back-end (\textit{e.g.}, place recognition and loop detection). However, their work lacks a unified mathematical formulation of Semantic SLAM and does not cover semantic data association or map optimization in depth. Wen \textit{et al.}\cite{wen2021semantic} study the influence of dynamic objects on visual SLAM and leverage Mask R-CNN to segment dynamic regions and initialize camera poses. Their approach combines photometric, geometric, and depth consistency to differentiate between static and dynamic features. Meanwhile, Chen \textit{et al.}\cite{chen2025semantic} surveys semantic object association methods—both probabilistic and non-probabilistic—and briefly mentions semantic localization in indoor and outdoor environments, but only at a high level. Most recently, Wang \textit{et al.}~\cite{wang2024survey} categorize Semantic SLAM in dynamic scenes into four key areas: (1) dynamic object exclusion, (2) semantic tracking for localization, (3) semantic mapping under dynamic conditions, and (4) multi-sensor fusion. While this survey emphasizes the challenges of real-world deployment, it does not offer a unified treatment of Semantic SLAM nor a detailed breakdown of how semantics are integrated into specific SLAM components such as data association, loop closure, or cost modeling.

In summary, existing surveys underscore the significance of integrating semantics into SLAM and outline historical trends and potential applications. However, they fall short of presenting Semantic SLAM as a unified probabilistic framework or exploring the four critical components—semantic extraction and mapping, semantic data association, semantic uncertainty modeling, and semantic-aware cost functions—in a structured manner. Therefore, this paper aims to fill this gap by formulating Semantic SLAM probabilistically and providing an in-depth discussion of its modular architecture and open challenges.

\subsection{Paper Structure}
The remainder of this paper is structured as follows. Section~\ref{sec:problem} introduces the Semantic SLAM problem, beginning with a unified probabilistic formulation and followed by the decomposition into key subproblems. Section~\ref{sec:extraction} discusses semantic feature extraction from raw sensor data, including object detection, segmentation, and scene understanding. Section~\ref{sec:localization} focuses on semantic localization methods that leverage object-level and high-level contextual cues for robust pose estimation. Section~\ref{sec:mapping} covers semantic mapping techniques, including map fusion strategies and handling dynamic environments. Section~\ref{sec:association} explores semantic data association and
\onecolumn
\clearpage
\begin{landscape}
\footnotesize
\begin{longtable}{@{\extracolsep{\fill}}R{2.4cm}||C{0.4cm} C{1.3cm}C{1.8cm} C{2.0cm} C{0.7cm} C{1.1cm} C{0.8cm} C{1cm} C{0.5cm} C{0.4cm} C{1.0cm} C{0.4cm} C{0.7cm}}
\caption{A DETAILED COMPARISON OF REPRESENTATIVE AND STATE-OF-THE-ART SEMANTIC SLAM ALGORITHMS ACROSS MULTIPLE CRITERIA.}
\label{tab:vslam} \\
\toprule
\textbf{Reference} & \textbf{Year} & \textbf{SLAM Approach} & \textbf{Semantic Perception} & \textbf{Map Repr.} & \textbf{Dynamic Aware} & \textbf{Open Vocab.} & \textbf{Archi.} & \textbf{Library} & \textbf{Env.} & \textbf{Online} & \textbf{Loop Closure} & \textbf{Large Scale} & \textbf{Public Avai.} \\
\toprule 
\endfirsthead

\multicolumn{14}{c}%
{{\tablename\ \thetable{} -- continued from previous page}} \\
\hline
\textbf{System} & \textbf{Year} & \textbf{SLAM Approach} & \textbf{Semantic Perception} & \textbf{Map Repr.} & \textbf{Dyn.} & \textbf{Open Vocab} & \textbf{Arch.} & \textbf{Library} & \textbf{Env.} & \textbf{Online} & \textbf{Loop Closure} & \textbf{Large Scale} & \textbf{Publicly Ava.} \\
\hline 
\endhead

\hline \multicolumn{14}{r}{{Continued on next page}} \\ 
\endfoot

\endlastfoot

\multicolumn{14}{c}{\textbf{Classic, Object-based, and Early Dense SLAM}} \\
\hline
CNN-SLAM~\cite{tateno2017cnn}                        & 2017 & DD & CNN DP     & DM         & - & - & PG      & g2o        & \faHome     & + & + & - & \cmark \\
SemanticFusion~\cite{mccormac2017semanticfusion}     & 2017 & DF & CNN SS     & SM         & - & - & F2M     & \_\_       & \faHome     & + & - & - & \cmark \\
Co-Fusion~\cite{runz2017co}                          & 2017 & OF & CNN SS     & SM         & + & - & F2M     & \_\_       & \faHome     & + & - & - & \cmark \\
VSO~\cite{lianos2018vso}                             & 2018 & F  & SS         & SM         & - & - & BA      & g2o        & \faDoorOpen & + & + & + & \xmark \\
DS-SLAM~\cite{yu2018ds}                              & 2018 & F  & SegNet     & SP         & + & - & PG+BA   & ceres/g2o  & \faTree     & + & + & + & \cmark \\
MaskFusion~\cite{runz2018maskfusion}                 & 2018 & OF & Mask R-CNN & SM         & + & - & F2M     & \_\_       & \faHome     & + & - & - & \cmark \\
CubeSLAM~\cite{yang2019cubeslam}                     & 2019 & OBA& O          & 3D Cuboids & + & - & PG+BA   & GTSAM      & \faDoorOpen & + & + & + & \cmark \\
QuadricSLAM~\cite{nicholson2018quadricslam}          & 2019 & OBA& O          & DQ         & - & - & FG      & GTSAM      & \faDoorOpen & - & + & + & \cmark \\
EAO-SLAM~\cite{wu2020eao}                            & 2020 & OBA& O          & 3D Cuboids & - & - & JO      & GTSAM      & \faHome     & + & - & + & \cmark \\
Fusion++~\cite{mccormac2018fusion++}                 & 2018 & OF & Mask R-CNN & TSDF       & - & - & PG      & g2o        & \faHome     & + & + & - & \xmark \\
PanopticFusion~\cite{narita2019panopticfusion}       & 2019 & DF & P          & PV         & - & - & F2M     & \_\_       & \faHome     & + & - & - & \xmark \\
Kimera~\cite{rosinol2020kimera}                      & 2020 & F  & P          & 3D SeM     & + & - & FG      & GTSAM & \faDoorOpen & + & - & + & \cmark \\
Hydra~\cite{hughes2022hydra}                         & 2022 & F  & SS         & 3D SG      & - & - & FG      & GTSAM      & \faDoorOpen & + & + & + & \cmark \\
\hline
\multicolumn{14}{c}{\textbf{Continuous Representation SLAM}} \\
\hline
iMAP~\cite{sucar2021imap}                            & 2021 & I  & C          & MLP         & - & - & JO     & PyTorch    & \faHome     & + & - & - & \cmark \\
NICE-SLAM~\cite{zhu2022nice}                         & 2022 & I  & C          & FeG+MLP     & - & - & JO     & PyTorch    & \faHome     & + & - & - & \cmark \\
Point-SLAM~\cite{sandstrom2023point}                 & 2023 & I  & C          & FeG+MLP     & - & - & JO     & PyTorch    & \faHome     & + & - & - & \cmark \\
H2-Mapping~\cite{jiang2023h}                         & 2023 & I  & C          & FeG+MLP     & - & - & JO     & PyTorch    & \faHome     & + & - & + & \cmark \\
SNI-SLAM~\cite{zhu2024sni}                           & 2024 & I  & SS         & VGM+MLP     & - & - & \_\_   & \_\_       & \faHome     & + & - & - & \cmark \\
H3-Mapping~\cite{jiang2024h3}                        & 2024 & I  & C          & FeG+MLP     & - & - & JO     & PyTorch    & \faHome     & + & - & + & \cmark \\
DynaMoN~\cite{schischka2024dynamon}                  & 2024 & I  & SS         & FeG+MLP     & + & - & JO     & PyTorch    & \faHome     & + & - & - & \cmark \\
SplaTAM~\cite{keetha2024splatam}                     & 2024 & GS & C          & 3D G        & - & - & JO     & PyTorch    & \faHome     & + & - & - & \cmark \\
NEDS-SLAM~\cite{ji2024neds}                          & 2024 & GS & SS         & SeG         & - & - & JO     & PyTorch    & \faHome     & + & - & + & \xmark \\
GS-SLAM~\cite{yan2024gs}                             & 2024 & GS & C          & 3D G        & - & - & JO+BA  & PyTorch    & \faHome     & + & - & + & \cmark \\
SGS-SLAM~\cite{li2024sgs}                            & 2025 & GS & S          & SeG         & + & - & JO     & PyTorch    & \faHome     & + & - & - & \cmark \\
SemGauss-SLAM~\cite{zhu2025semgauss}                 & 2025 & GS & SS         & SeG         & - & - & JO     & PyTorch    & \faHome     & + & - & + & \xmark \\
\hline
\multicolumn{14}{c}{\textbf{Foundation Model-based SLAM Systems}} \\
\hline
ConceptFusion~\cite{jatavallabhula2023conceptfusion} & 2023 & DF & CLIP       & DFF         & - & + & F2M    & PyTorch    & \faHome     & - & - & - & \cmark \\
FM-Fusion~\cite{liu2024fm}                           & 2024 & DF & SAM + CLIP & VGM         & - & + & F2M    & PyTorch    & \faHome     & + & - & - & \cmark \\
ConceptGraphs~\cite{gu2024conceptgraphs}             & 2024 & SG & CLIP       & SM          & - & + & PG     & PyTorch    & \faHome     & + & - & - & \cmark \\
LOSS-SLAM~\cite{singh2024loss}                       & 2024 & OF & DINO       & OM          & - & + & FG     & GTSAM      & \faHome     & + & - & - & \xmark \\
LEXIS~\cite{kassab2024language}                      & 2024 & LD & VLM        & TM          & - & + & PG     & PyTorch    & \faHome     & + & + & - & \xmark \\
Hier-SLAM++~\cite{li2025hier++}                      & 2025 & GS & LLM+GM     & SeG         & - & + & JO+PG  & PyTorch    & \faHome     & + & - & + & \xmark \\
\hline
\multicolumn{14}{c}{\textbf{Multi-Robot Semantic SLAM Systems}} \\
\hline
Kimera-Multi~\cite{tian2022kimera}                   & 2022 & F   & SS        & D SeM       & - & - & PG     & GTSAM      & \faDoorOpen & + & + & + & \cmark \\
SlideSLAM~\cite{liu2024slideslam}                    & 2024 & OF  & IS        & D OM        & - & - & PG     & GTSAM      & \faDoorOpen & + & + & + & \cmark \\
HAMMER~\cite{yu2025hammer}                           & 2025 & GS  & CLIP      & SeG         & - & + & PG     & BA         & \faDoorOpen & + & - & + & \xmark \\
\bottomrule
\end{longtable}
\begin{tabular}{p{4.5cm}|p{4.5cm}|p{5cm}|p{4cm}|p{3.5cm}}
F: Feature based              & SM: Sparse Map              & DP: Depth Prediction                      & PG: Pose Graph           & + : Present                  \\
DD: Dense Direct              & DM: Dense Map               & SS: Semantic Segmentation                 & FG: Factor Graph         &  - : Absent                   \\
DF: Dense Fusion              & SM: Surfer Map              & O: Object Detection                       & F2M: Frame to Model      & \faHome: Indoor              \\
LD: Language Driven           & SP: Sparse Point            & P: Panoptic Segmentation                  & BA: Bundle Adjustment    & \faTree: Outdoor             \\
OF: Object Fusion             & TM: Topological Map         & DQ: Dual Quadrics                         & JO: Joint Optimization   & \faDoorOpen: Indoor/Outdoor  \\
SeM: Semantic Mesh            & OM: Object Map              & TSDF: Truncated Signed Distance Function  & FeG: Feature Grid        & D: Distributed\\
I: Neural Implicit            &  VLM: Vision Language Model & IS: Instance Segmentation                 & G: Gaussians             & VGM: Voxel Grid Map\\
SG: Scene Graph               &  LLM: Large Language Model  & MLP: Multi-Layer Perceptron               & C: Color Only            & SeG: Semantic Gaussian\\
GS: Gaussian Splatting        &  GM: Generation Model       & PV: Panoptic Voxels                       & DFF: Dense Feature  Field& SfM: Structure from Motion\\
\bottomrule
\end{tabular}
\end{landscape}
\twocolumn

\begin{table*}[ht] 
\centering
\caption{Comparison of topic coverage across related works.} ~\label{tab:surveypaper}
\footnotesize  
\begin{tabular}{p{1.2cm} p{3.7cm}||c|c|c|c|c|c|c|c}
\multicolumn{2}{c||}{\textbf{Topic}} &  
\textbf{\textit{Kostavelis}~\cite{kostavelis2015semantic}} & 
\textbf{\textit{Sualeh}~\cite{sualeh2019simultaneous}} & 
\textbf{\textit{Chen}~\cite{chen2022semantic}} & 
\textbf{\textit{Chen}~\cite{chen2022overview}} & 
\textbf{\textit{Pu}~\cite{pu2023visual}} & 
\textbf{\textit{Wang}~\cite{wang2024survey}} &
\textbf{\textit{Chen}~\cite{chen2025semantic}} & 
\textbf{\textit{Our}}\\
\hline
\multirow{2}{=}{Introduction} 
& Historical review             & Briefly & Yes     & Yes      & Yes     & Yes     & Briefly & Yes      & Yes \\
& Problem formulation           & No      & No      & No       & No      & No      & No      & Briefly  & Yes\\
\hline

\multirow{5}{=}{Modular scheme} 
& Semantic Extraction                   & Briefly & Briefly & Briefly  & Yes     & Yes     & No      & Yes      & Yes\\
& Semantic Localization                 & No      & No      & No       & No      & Briefly & Briefly & No       & Yes\\
& Semantic Mapping                      & Yes     & No      & Yes      & Briefly & Yes     & Yes     & Briefly  & Yes\\
& Semantic Data Association             & No      & No      & No       & No      & Briefly & No      & Yes      & Yes\\
& Semantic Back-end Optimization        & No      & Briefly & Yes      & No      & Briefly & No      & Yes      & Yes\\
\hline

\multirow{3}{=}{Alternative approaches} 
& Learning-based and Foundation Model   & No      & No      & No       & No      & No      & No      & No       & Yes\\
& Continuous Representations            & No      & No      & No       & No      & No      & No      & No       & Yes\\
& Multi-Robot Systems                   & No      & No      & Briefly  & No      & Briefly & No      & No       & Yes\\
\hline

\multirow{7}{=}{Open problems} 
& Semantic benchmarks                   & Briefly & No      & No       & Briefly & No      & No      & No       & Yes\\
& Active Semantic SLAM                  & No      & No      & No       & No      & No      & No      & No       & Yes\\
& Lifelong learning                     & No      & No      & No       & No      & No      & Yes     & Briefly  & Yes\\
& Generalization \& robustness          & No      & No      & Briefly  & No      & Briefly & Briefly & Briefly  & Briefly\\
& Reproducible research                 & No      & No      & No       & No      & No      & No      & No       & Briefly\\
& Practical applications                & Yes     & No      & No       & Briefly & No      & Briefly & No       & Briefly\\
\hline
\end{tabular}
\end{table*}

\noindent  integration, emphasizing probabilistic models and multi-instance tracking. Section~\ref{sec:optimization} addresses loop closure detection and global optimization incorporating semantic constraints. Section~\ref{sec:learning} presents learning-based approaches, including Deep Learning and foundation models. Section~\ref{sec:continuous} reviews recent advances in continuous and implicit scene representations, such as neural fields and 3D Gaussian splatting. Section~\ref{multirobot} extends the discussion to multi-robot Semantic SLAM, focusing on collaborative semantic mapping and distributed optimization. Section~\ref{sec:question} outlines open research challenges, including lifelong learning, zero-shot generalization, and standardization of semantic benchmarks. Finally, Section~\ref{sec:conclusion} concludes the paper with a summary of key insights and future directions for the field.

\section{The Semantic SLAM Problem}
\label{sec:problem}

\subsection{Problem Formulation}
    
Consider the Semantic SLAM problem, where the robot's state is represented as $\mathbf{\mathcal{X}} \triangleq \{\textbf{X}_t\}_{t=1}^T \in SE(3)$ for $t = 1, \ldots, T$, and evolves according to deterministic discrete-time kinematics:
\begin{equation}
    \mathbf{X}_t := \begin{bmatrix}
        \mathbf{R}_t & \mathbf{p}_t \\
        0^T & 1
    \end{bmatrix},
\end{equation}
where $\mathbf{R}_t \in SO(3)$ denotes the rotation matrix and $\mathbf{p}_t \in \mathbb{R}^3$ the translation vector of the robot pose.

Semantic SLAM extends classical SLAM by jointly estimating the robot trajectory and a map enriched with semantic information. The goal is to estimate the trajectory $\mathcal{X}$, a set of landmark states $\mathcal{L} = \{L_m\}_{m=1}^M$, and semantic information $\mathcal{S} = \{\mathbf{S}_c\}_{c=1}^C$ corresponding to $C$ semantic classes. We define the semantic map as $\mathcal{M} = \{ \mathbf{M}_t \}_{t=1}^T$. These estimates are conditioned on a sequence of observation measurements $\mathcal{Z} = \{ \mathbf{Z}_t \}_{t=1}^T$ and control inputs $\mathcal{U} = \{\mathbf{U}_t\}_{t=1}^T$.

To infer this state, the system maintains an internal \textit{belief} or \textit{information state} \cite{thrun2002probabilistic, thrun2005probabilistic, sigaud2013markov}, denoted by $b_t(\mathbf{X}_t)$, representing the posterior probability distribution at time $t$:
\begin{equation}
    b_t(\mathbf{X}_t) \triangleq p \left( \mathbf{X}_t, \underbrace{L_{1:t}, \mathbf{S}_{1:t}}_{\text{descriptor } \mathbf{\Theta}}, \mathbf{M}_{1:t} \mid \mathbf{Z}_{1:t}, \mathbf{U}_{1:t-1} \right),
\end{equation}
where $\mathbf{Z}_{1:t}$ and $\mathbf{U}_{1:t-1}$ denote the sequences of observations and controls, respectively. The \textit{belief space} of probability destiny functions (pdf) over the set $\mathcal{X}$ is defined as:
\begin{equation}
    \mathcal{B}(\mathcal{X}) \triangleq \left\{ b : \mathcal{X} \rightarrow \mathbb{R} \mid \int b(\mathbf{X}) d\mathbf{X} = 1,\, b(\mathbf{X}) \geq 0 \right\}.
\end{equation}

In order to estimate the robot's pose and establish a map, agents must be capable of predicting posterior belief distributions, that is the pdf over $\mathcal{X}$ after taking observation $\mathbf{Z}_{t+1}$ and performing a control input $\mathbf{U}_t$:
\begin{equation}
    b_{t+1} (\mathbf{X}_{t+1}) \triangleq p(\mathbf{X}_{t+1}, \Theta_{t+1}, \mathbf{M}_{t+1} \mid \mathbf{Z}_{t+1}, \mathbf{U}_t, b_t(\mathbf{X}_t)).
\end{equation}

This problem can be factorized using Bayes’ rule and the Markov assumption as:
\begin{equation}
\begin{aligned}
    p(\mathbf{X}_{1:t}, \mathbf{M}, \Theta_{1:t} & \mid \mathbf{Z}_{1:t}, \mathbf{U}_{1:t}) \propto \prod_{\tau=1}^t p(\mathbf{Z}_\tau \mid \mathbf{X}_\tau, \Theta_\tau, \mathbf{M})  \\
    & \cdot p(\Theta_\tau \mid \mathbf{X}_\tau, \mathbf{M}) \cdot p(\mathbf{X}_\tau \mid \mathbf{X}_{\tau-1}, \mathbf{U}_\tau),
\end{aligned}
\end{equation}
where, $p(\mathbf{X}_\tau \mid \mathbf{X}_{\tau-1}, \mathbf{U}_\tau)$ is the motion model, $p(\mathbf{Z}_\tau \mid \mathbf{X}_\tau, \Theta_\tau, \mathbf{M})$ is the sensor likelihood (observation model), and $p(\Theta_\tau \mid \mathbf{X}_\tau, \mathbf{M})$ represents the semantic likelihood conditioned on the pose and map.

In Semantic SLAM, the map \( \mathbf{M} \) is characterized by a collection of semantic landmarks.
\begin{equation}
    \mathbf{M} = \{ (l_i, c_i) \mid l_i \in \mathbb{R}^3, s^c_i \in \mathcal{C} \},
\end{equation}
where $l_i$ is the 3D location and $s^c_i$ is the semantic class of landmark $i$, and $\mathcal{C} = \{1, \dots, C \}$ is the set of possible semantic labels.

The belief can be updated recursively via a Bayesian filter:
\begin{equation}
\begin{aligned}  
    b_t(\mathbf{X}_t, \mathbf{M}, \Theta_t) \propto &p(\mathbf{Z}_t \mid \mathbf{X}_t, \mathbf{M}, \Theta_t) \int p(\mathbf{X}_t \mid \mathbf{X}_{t-1}, \mathbf{U}_t) \\
    & \cdot b_{t-1}(\mathbf{X}_{t-1}, \mathbf{M}, \Theta_{t-1}) \, d\mathbf{X}_{t-1}.
\end{aligned}
\end{equation}

In addition, Semantic SLAM must also reason about \textit{semantic uncertainty} of the system. A piece of semantic information can be extracted from keyframe $t$ as $\mathbf{s}_k = (s_k^c, s_k^s, s_k^b) \in \mathbf{S}_t$~\cite{bowman2017probabilistic}, consisting of a categorical label $s_k^c \in \mathcal{C}$, a score confidence $s_k^s$, and a bonding box $s_k^b$. We may model its uncertainty using a distribution such as:
\begin{equation}
    p(\mathbf{S}_t \mid \mathbf{Z}_t) = \text{Categorical}(\pi_t), \quad \pi_t = \text{softmax}(f_\theta(\mathbf{Z}_t)),
\end{equation}
where $f_\theta(\cdot)$ is a deep network predicting semantic logits from sensor data \cite{long2015fully, he2017mask, chen2020sloam}. 
The robot state is commonly assumed Gaussian with a pdf $b(\mathbf{X})$ having mean $\hat{\sigma}$ and covariance $\sum_x$~\cite{indelman2015planning, sunderhauf2017dual}. The \textit{semantic data association} process infers a set of latent correspondences $\mathcal{D}$ linking current observations to map landmarks. The standard formulation follows a maximum likelihood objective:
\begin{equation} \label{eq:mlproblem}
    \mathcal{X}^*_{ml}, \Theta^*_{ml} = \argmax_{\mathcal{X}, \Theta} p(\mathcal{Z} \mid \mathcal{X}, \Theta).
\end{equation}

Some works also assume \textit{maximum likelihood} (ML) observations~\cite{salas2013slam++, bowman2017probabilistic, sunderhauf2017dual, mccormac2018fusion++, nicholson2018quadricslam, yang2019cubeslam, doherty2020probabilistic}, \textit{i.e.} the semantic measurement data associations are independent across keyframes:
\begin{equation}
\begin{aligned}
    \mathcal{D^*} &= \argmax_{\mathcal{D}} p(\mathcal{D} \mid \mathcal{X}^0, \Theta^0, \mathcal{Z}), \\
    D_{i+1} &=  \argmax_{\mathcal{D}} p(\mathcal{D} \mid \mathcal{X}^i, \Theta^i, \mathcal{Z}).
\end{aligned}
\end{equation}
An Expectation-Maximization (EM) approach is often applied:
\begin{equation}
\begin{aligned}
    \mathcal{X}^{i+1}, \Theta^{i+1} &= \argmax_{\mathcal{X, D}, \Theta} p(\mathcal{Z} \mid \mathcal{X, D}, \Theta), \\
    \mathcal{X}^{i+1}, \Theta^{i+1} &= \argmax_{\mathcal{X}, \Theta, \mathcal{D}} \mathbb{E}_\mathcal{D}[\log p(\mathcal{Z} \mid \mathcal{X}, \Theta, \mathcal{D}) \mid \mathcal{X}^i, \Theta^i, \mathcal{Z}] \\
    &= \argmax_{\mathcal{X}, \Theta, \mathcal{D}} \sum_{\mathcal{D} \in \mathbb{D}} p(\mathcal{D} \mid \mathcal{X}^i, \Theta^i, \mathcal{Z}) \log p(\mathcal{Z} \mid \mathcal{X}, \Theta, \mathcal{D})
\end{aligned}
\end{equation}

\subsection{Main Subproblems}

Although Semantic SLAM is initially articulated as a unified approach, it is frequently split into a modular sequence for the sake of manageability and the development of the system. The five main constituent subproblems include:

\begin{enumerate}
    \item \textit{Semantic extraction}: this stage involves extracting high-level semantic information from raw sensory inputs $\mathbf{Z}_t$. Deep learning models such as convolutional neural networks (CNNs), semantic segmentation networks (\textit{e.g.}, Mask R-CNN~\cite{he2017mask}, DeepLab~\cite{chen2017rethinking}), or transformers are typically used to generate semantic labels, object masks, or scene descriptions $\mathbf{S}_t$. The output includes categorical distributions, pixel-wise segmentations, and object-level bounding boxes.

    \item \textit{Semantic localization}: in this stage, semantic features $\mathbf{S}_t$ are used alongside geometric measurements to enhance the robot's pose estimation $\mathbf{X}_t$. Techniques such as semantic re-weighting of features, use of object-level landmarks, or probabilistic pose refinement using semantic masks are commonly applied.

    \item \textit{Semantic mapping}: this stage constructs and maintains a geometric-semantic map $\mathcal{M}$, where each element combines a 3D location $l_i$ and semantic class $c_i$. The map can be built as sparse object-level landmarks, dense semantic voxel grids, or mesh-based reconstructions. Semantic fusion techniques update the map incrementally by aggregating multi-view observations and resolving class conflicts.

    \item \textit{Semantic data association}: the final stage addresses the temporal consistency of semantic landmarks across frames. This includes matching detected objects to prior map entities, resolving ambiguity in class predictions. Probabilistic frameworks such as maximum likelihood estimation or expectation-maximization are often employed.
    \item \textit{Semantic 
    closed loop and Optimization}: past frames are revisited and semantic landmarks are aligned to correct accumulated drift. Optimization is often done via bundle adjustment (BA) or factor graph smoothing.
\end{enumerate}

To provide a clear and well-organized presentation, 
given that a substantial amount of existing literature breaks down Semantic SLAM into five distinct stages, we have opted to review each of these stages individually in Sections~\ref{sec:extraction} to~\ref{sec:optimization}. In addition, this structure supports a robust theoretical foundation and allows extensions such as learning-based methods (Section~\ref{sec:learning}), continuous scene representation (Section~\ref{sec:continuous}), and multi-robot collaboration (Section~\ref{multirobot}). It also creates a pathway for incorporating active decision-making for semantic exploration, bridging toward Active Semantic SLAM.

\section{Stage I: Semantic Extraction in Semantic SLAM} \label{sec:extraction}

Semantic extraction is the first and foundational stage in the semantic vSLAM pipeline. It encompasses a suite of techniques designed to identify, classify, and delineate meaningful entities and contexts from raw sensor inputs, typically visual data from cameras or depth information from RGB-D sensors. Effective semantic extraction aims to transform noisy and ambiguous sensor data into structured semantic information that is robust, accurate, and relevant for downstream tasks like mapping, localization, and decision-making. This section delves into the core components of semantic extraction: discerning individual objects through object detection (A), achieving pixel-level understanding via object and instance segmentation (B), recognizing broader environmental contexts through place categorization (C), and refining object representations using contextual priors (D).

\subsection{Object Detection}

Object detection is a fundamental computer vision task that involves identifying and localizing instances of predefined object categories within an environment. Typically, this task involves predicting bounding boxes around each object and associated semantic labels (\textit{e.g.,} car, human, chair). Before the dominance of deep learning, several traditional methods were prominent, such as the Histogram of Oriented Gradient (HOG) descriptor and Deformable Part Model (DPM). However, these methods are generally less accurate and less adaptable to diverse object classes compared to modern deep learning approaches, particularly in complex scenes. Deep learning, especially CNNs, has dramatically advanced object detection. These methods typically optimize a multi-task loss function, which combines losses for object classification and bounding box regression. Semantic vSLAM systems often utilize single-stage detectors due to their real-time inference capabilities. Among the most widely used are the YOLO series~\cite{redmon2016you, redmon2017yolo9000, redmon2018yolov3, bochkovskiy2020yolov4, yolov5, li2022yolov6, wang2023yolov7, yolov8, wang2024yolov9}, which unify object classification and bounding box regression into a single forward pass. Another efficient variant is the SSD (Single Shot MultiBox Detector)~\cite{liu2016ssd}, which uses multi-scale feature maps for detection.

To improve detection accuracy, especially in cluttered or dynamic scenes, many systems adopt two-stage architectures such as R-CNN~\cite{girshick2014rich}, Fast R-CNN~\cite{girshick2015fast}, and Faster R-CNN~\cite{ren2015faster}. These methods use region proposal networks (RPNs) followed by refinement networks for accurate object localization and classification. Mask R-CNN~\cite{he2017mask} further extends Faster R-CNN to include object instance segmentation, which is especially beneficial for overlapping or deformable objects. Recent advances incorporate transformer architectures, exemplified by DETR~\cite{carion2020end} and DINO~\cite{zhang2022dino}, which model long-range dependencies in the image and perform set-based global object prediction. These models have demonstrated strong performance in complex environments with occlusions and semantic ambiguity. Some approaches extend object detection into 3D by incorporating depth sensing or monocular depth estimation. Models such as VoxelNet~\cite{zhou2018voxelnet}, Frustum PointNets~\cite{qi2018frustum}, and DenseFusion~\cite{wang2019densefusion} integrate RGB-D  to localize objects in 3D space. 

In the context of Semantic SLAM, detected objects serve as semantic landmarks, offering more stable and interpretable features than low-level keypoints~\cite{soares2021crowd, wu2022yolo, xia2023yolo, gong2024real}. Formally, given an input image $I_t$ at time $t$ over $N_t$ objects in observation $t$, the object detector outputs a set of detections:
\begin{equation}
     \mathbf{s}_k = (s_k^c, s_k^s, s_k^b)_{k=1}^{N_t} \in \mathbf{S}_t,
\end{equation}
where $s_k^c$ is the semantic class label, $s_k^s$ is the confidence score, and $s_k^b \in \mathbb{R}^4$ denotes the shape information. These outputs are used to initialize or update object-level landmarks in the semantic map. Hence, several semantic vSLAM works combine YOLO with some state-of-the-art vSLAM methods like ORB-SLAM, RGBD-SLAM for obstacle avoidance and UAV tracking~\cite{rui2021multi}, navigation guidance system~\cite{xie2022multi}, and real-time decision-making in autonomous driving scenarios~\cite{li2023rgbd, canh2023object}. On the other hand, some works~\cite{shao2021faster, zhang2022dynamic} integrate Mask R-CNN or Faster R-CNN as a semantic filter, which provides different semantic labels for the image context. Soares \textit{et al.}~\cite{soares2019visual} explored the trade-off between detection accuracy and inference speed by fusing YOLO and Mask R-CNN outputs in crowded environments. Shao \textit{et al.}~\cite{shao2021faster} and Zhang \textit{et al.}~\cite{zhang2022dynamic} proposed systems where Faster R-CNN filters spurious keypoints for robust localization in dynamic environments. Furthermore, systems such as QuadricSLAM~\cite{nicholson2018quadricslam} and CubeSLAM~\cite{yang2019cubeslam} represent objects as parameterized shapes (quadrics, cuboids), optimizing their poses alongside the camera trajectory.

In summary, object detection is the critical front-end of Semantic SLAM, bridging raw sensor data with symbolic world understanding. It is essential for semantic landmark identification~\cite{zhang2018semantic, islam2025advancing}, dynamic object handling~\cite{soares2019visual, wu2024dyn}, constructing object-aware semantic maps~\cite{canh2023object, li2023rgbd}, and enabling task-specific applications~\cite{rui2021multi,xie2022multi}.

\subsection{Object/Instance Segmentation}

 While object detection provides coarse bounding box localization of objects, \textit{object} and \textit{instance segmentation} provide a more granular scene understanding by assigning a class label to every pixel (semantic segmentation) or to individual object instances (instance segmentation). This pixel-level classification is invaluable for Semantic SLAM for several reasons: (1) it enables more accurate geometric modeling of the environment, (2) it improves data association by reducing ambiguities in cluttered or overlapping scenarios, and (3) it provides richer semantic context for tasks such as dynamic object filtering and scene understanding. Popular semantic segmentation networks include U-Net~\cite{ronneberger2015u}, Bayesian SegNet~\cite{kendall2015bayesian}, SegNet~\cite{badrinarayanan2017segnet}, PSPNet~\cite{zhao2017pyramid} and DeepLabv3+~\cite{chen2018encoder}. For instance segmentation, approaches like Mask R-CNN~\cite{he2017mask}, SOLO~\cite{wang2020solo}, and YOLACT~\cite{bolya2019yolact} are commonly used. Other key methods include PANet~\cite{liu2018path}, which improves information flow between layers, and HTC~\cite{chen2019hybrid}, which introduces multi-stage cascaded heads for refined segmentation. Recently, transformer-based models such as Mask2Former~\cite{cheng2022masked} and QueryInst~\cite{fang2021instances} have demonstrated improved performance by leveraging global context and set-based reasoning for mask prediction. 

One of the major challenges in SLAM is coping with dynamic objects. Instance segmentation allows for the explicit identification and masking of dynamic regions. For instance, DynaSLAM employs Mask R-CNN for foreground-background separation, ensuring that only static background points contribute to the SLAM backend optimization~\cite{bescos2018dynaslam}. Similarly, DS-SLAM~\cite{yu2018ds} uses semantic masks to segment out people and other moving objects, thus enhancing pose estimation stability. Formally, for a given image $I_t$ at time $t$ over $N_t$ object in observation $t$, the segmentation model produces:
\begin{equation}
    \mathbf{s}_k = (s_k^c, s_k^s, s_k^m)_{k=1}^{N_t} \in \mathbf{S}_t,
\end{equation}
where $s_k^c$ is the class label, $o_k^s$ is the confidence score, and $s_k^m \in \{0,1\}^{H \times W}$ is the binary mask for the $k$-th instance. Segmentation information can be fused with depth maps or point clouds to generate 3D semantic reconstructions. SemanticFusion~\cite{mccormac2017semanticfusion}, Fusion++~\cite{mccormac2018fusion++}, MID-Fusion~\cite{xu2019mid}, and Voxblox~\cite{oleynikova2017voxblox} demonstrate how semantic masks can be incrementally fused with geometric data. Such dense semantic mapping enhances the robot’s ability to reason about spatial relationships and object permanence in dynamic scenes. Beyond 2D segmentation, 3D instance segmentation is increasingly explored. Approaches like PointGroup~\cite{jiang2020pointgroup} and 3D-SIS~\cite{hou20193d} predict 3D instance masks from RGB-D data or point clouds, providing direct volumetric segmentation useful for dense semantic mapping.

Object and instance segmentation enhance the granularity and reliability of Semantic SLAM. By providing detailed object contours and instance-level masks, these methods bridge the gap between visual perception and 3D mapping, dynamic object management, enabling the system to reason about both the geometry and semantics of its environment at a fine-grained level. Semantic map construction benefits from segmentation through voxel-based fusion, as in Bayesian updates by Stückler \textit{et al.}~\cite{stuckler2012semantic} or the label-oriented voxelgrid filter by Shi \textit{et al.}~\cite{shi2021rgb}, and surfel (surface element)-based or segment-based maps~\cite{canh2024s3m}.

In addition, panoptic segmentation unifies these two tasks, providing a more holistic scene understanding~\cite{kirillov2019panoptic, mohan2021efficientps, de2020fast}. It assigns both a semantic label and a unique instance ID (if applicable) to every pixel in the image. This allows it to jointly represent both amorphous ``stuff'' categories (like road, sky, vegetation) and discrete ``thing'' instances (like cars, people). The performance of panoptic segmentation is often measured by the panoptic quality metric and Intersection-over-Union (IoU) of predicted segment $q$ and ground truth segment $g$, which captures both recognition and segmentation quality:
\begin{equation}
\mathcal{A}_{PQ} = \frac{\sum_{(q, g)\in TP} IoU(q,g)}{|TP| + \frac{1}{2} |FP| + \frac{1}{2}|FN|},
\end{equation}
where TP, FP, and FN are the set of true positives, false positives, and false negatives, respectively. Conceptually, PQ can be seen as the product of Segmentation Quality (SQ) and Recognition Quality (RQ): $\mathcal{A}_{PQ} = \mathcal{A}_{SP} \times \mathcal{A}_{RQ}$.
Several SLAM systems now leverage panoptic segmentation for their comprehensive output. Panoptic-SLAM~\cite{abati2024panoptic} and PS-SLAM~\cite{li2025ps} are recent visual SLAM systems designed for dynamic environments that use panoptic segmentation networks to identify both static background ``stuff'' and dynamic ``thing'' instances. This allows for more precise dynamic feature filtering and the creation of more detailed semantic maps

\subsection{Place Categorization}

Place categorization is a pivotal function within Semantic SLAM, responsible for assigning high-level semantic labels (\textit{e.g.} ``corridor'', ``kitchen'', ``office'', ``street scene'') to distinct areas within the robot's perceived environment~\cite{sunderhauf2016place, garg2017improving}. This capability extends beyond individual object recognition to encompass a global, scene-level understanding of spatial context. Such contextual awareness is invaluable for tasks like semantic loop closure, hierarchical mapping for efficient planning, and topological localization. It also facilitates the incorporation of semantic priors, enabling more informed robot behavior and modular SLAM processing~\cite{arshad2023visem}. 

Robust place categorization remains challenging due to environmental variability, including dynamic lighting, occlusions, and seasonal or viewpoint changes. Early techniques relied on holistic image features, attempting to capture the ``gist'' of a scene using global descriptors. The Bag-of-Words (BoW) model, which quantizes local image features (\textit{e.g.,} SIFT, SURF, ORB ) into a ``visual vocabulary'', became a foundational technique for compact place representation and recognition~\cite{woo2024context}. However, these handcrafted approaches lack invariance to real-world scene diversity. With the advent of deep learning, CNNs and, more recently, transformers have become the backbone of place categorization. Models like AlexNet~\cite{krizhevsky2012imagenet}, VGGNet~\cite{simonyan2014very}, GoogLeNet~\cite{szegedy2015going}, and ResNet~\cite{he2016deep}, often pre-trained on large-scale datasets such as ImageNet~\cite{deng2009imagenet}, Places365~\cite{zhou2017places}, and SUN~\cite{patterson2014sun}, serve as powerful feature extractors or can be fine-tuned for specific scene recognition tasks. Scene-specific classifiers like PlaceNet~\cite{arandjelovic2016netvlad}, Places365 CNN~\cite{zhou2017places}, or transformer-based models such as ViT~\cite{dosovitskiy2020image}
and TransVPR~\cite{wang2022transvpr} have improved the robustness of categorization across different lighting, seasons, and occlusions. 

Formally, given an input image $I_t$ to a probability distribution $\pi_t$ over $N_t$ objects in observation $t$, the place categorization output is:
\begin{equation}
    \mathbf{s}_k = (\text{Categorical}(\pi_t))_{k=1}^{N_t} \in \mathbf{S}_t, \quad \pi_t = \text{softmax}(g_\phi(I_t)),
\end{equation}
where $g_\phi$ is a deep scene recognition model, typically the output of the final layers of the CNN. Some systems fuse object-level and scene-level to jointly enhance contextual reasoning and 
contextual understanding~\cite{kostavelis2017semantic}. This multi-scale fusion supports discriminative feature learning and enables robust global localization even under perceptual aliasing.

In Semantic SLAM, place categorization thus complements object-level observations by enabling semantic anchoring at the scene level, promoting spatial consistency and high-level task planning~\cite{kostavelis2015semantic}. Its integration with geometric reasoning and memory models remains an active area of research. Several studies incorporate place recognition into SLAM, either as a topological constraint for global localization~\cite{lin2018topological}, or to augment loop closure~\cite{zimmerman2022long} and support long-term navigation~\cite{suomela2024placenav}. In addition, hierarchical Semantic SLAM systems, such as those proposed by Bowman \textit{et al.}~\cite{bowman2017probabilistic} and Salas-Moreno \textit{et al.}~\cite{salas2013slam++}, often use place categories to structure spatial memory, enabling fast retrieval and reuse of semantic submaps. Some methods fuse place category with scene graphs or room templates for hierarchical map construction~\cite{muravyev2025prism}. Recent SLAM frameworks like DS-SLAM~\cite{yu2018ds}, PSPNet-SLAM~\cite{long2020pspnet}, and DRV-SLAM~\cite{ji2024drv} embed place categorization into the SLAM pipeline to provide semantic cues in dynamic and ambiguous environments. For example, semantic loop closures can be performed when the predicted place label matches historical observations, even under significant viewpoint change.

To summarize, semantic extraction in SLAM transforms raw sensor inputs $z_t$ into semantic observations $y_t$. This process is formally captured as:
\begin{equation}
    p(y_t \mid z_t) = \mathcal{S}(z_t),
\end{equation}
where $\mathcal{S}$ denotes a semantic encoder (\textit{e.g.}, CNN, transformer) producing categorical labels, segmentation masks, and confidence distributions~\cite{long2015fully, he2017mask, zhou2019semantic}.

\subsection{Object Regularization via Semantic Context Constraints}
While object detection and segmentation (\textit{e.g.} YOLO~\cite{redmon2018yolov3}, Faster R-CNN~\cite{ren2015faster}, DETR~\cite{carion2020end}) provide the initial semantic labels for entities in the environment, these raw perceptions often lack the consistency, plausibility, and adherence to real-world physical and semantic properties expected in a high-fidelity semantic map. Object regularization via semantic context constraints refers to the process of refining the representation of objects (their pose, shape, size, class, and relationships) by enforcing constraints derived from high-level semantic understanding of the scene and the objects themselves. In many practical scenarios, such geometric inaccuracies (\textit{e.g.} noisy data, segmentation error), semantic ambiguities, physical implausibility, temporal inconsistencies, and misclassifications arise when semantic context is ignored during object recognition.  To address this issue, semantic contextual constraints are introduced to regularize and refine the semantic predictions. 

A key aspect of regularization is the enforcement of geometric and shape priors. Objects are often regularized by fitting predefined geometric primitives such as cuboids for furniture, spheres for smaller objects, or planes for walls and tables to their observed point clouds or masks. The parameters of these primitives (\textit{e.g.}, dimensions, orientation) can be constrained based on typical values for an object's semantic class~\cite{civera2011towards}. This process can be formulated as an optimization problem where, for an object with parameters $\mathbf{s}_m = (s_m^p, s_m^b)$ including pose $s_m^p$ and shape $s_m^b$, and observation $\mathbf{Z}_t$, the goal is to find the maximum a posteriori (MAP) estimate:
\begin{equation}
    \mathbf{s}_m = \argmax_{\mathbf{s}_m} (log P(\mathbf{Z}_t \mid \mathbf{s}_m) + logP(\mathbf{s}_m \mid c_i)) ,
\end{equation}
where the likelihood term $P(\mathbf{Z}_t \mid \mathbf{s}_m)$ measures how well the models fit the data (\textit{e.g.}, point-to-model distance), while the prior $P(\mathbf{s}_M \mid c_i)$ penalizes deviations from expected shapes and sizes for the object's class $c_i$.

Beyond individual object geometry, topological and relational constraints define the expected spatial arrangements between entities. This includes support relationships (\textit{e.g.}, a ``monitor'' is on a ``desk''; a ``chair'' is on the ``floor'') and co-occurrence patterns. Such relationships can be modeled explicitly within a probabilistic framework or as factors in an optimization graph. Vasudevan and Siegwart~\cite{vasudevan2008bayesian} proposed a Bayesian framework where the joint probability of a place concept $\mathbf{s}_k$ depends not only on object counts $c_i$ but also on inter-object relationships $r_j$ (\textit{e.g.}, distance, orientation):
\begin{equation}
    P(\mathbf{s_k}, c_{1 \cdots n_1}, r_{1 \cdots n_2}) = P(\mathbf{s}_k) \cdot \prod_{i=1}^{n_1} P(c_i \mid \mathbf{s}_k) \cdot \prod_{j=1}^{n_2} P(r_j \mid \mathbf{s}_k).
\end{equation}
This allows the system to reason about plausible scene configurations. Similarly, graph-based representations, like the semantic graphs proposed by Kong \textit{et al.}~\cite{kong2020semantic}, Singh and Leonard~\cite{singh2024open}, and Muravyev \textit{et al.}~\cite{muravyev2025prism}, can encode these topological constraints as edges between object nodes.

These constraints are typically enforced through several computational methods. In factor graph optimization, which is central to modern SLAM, object properties and their relationships are directly incorporated into the graph~\cite{tennakoon2023factor}. An object's pose $\mathbf{X}_j$ and shape information $s_j^b$ become variables, and semantic constraints are added as factors. For instance, a relational factor between a cup $O_j$ and a table $O_k$ could be expressed as an energy term to be minimized:
\begin{equation}
    E_r (\mathbf{X}_j, \mathbf{X}_k) = f_c(IsOn(X_j, s_j^b, \mathbf{X}_k, s_k^b,)),
\end{equation}
where $f_c$ function penalizes configurations that violate the geometric ``IsOn'' predicate. Patel \textit{et al.}~\cite{patel2018semantic} describe this principle as ``semantically guided local and global pose optimization''.

Probabilistic fusion of multiple observations over time also provides powerful temporal regularization. As a robot observes an object from different viewpoints, inconsistencies in pose, shape, or semantic label can be averaged out. For example, voxel-based semantic maps can update the class probability distribution for a voxel $v$ recursively. Building on the work of Stückler \textit{et al.} (2012)~\cite{stuckler2012semantic}, Nakajima \textit{et al.}~\cite{nakajima2018fast} proposed updating the class probability $P(s_k^c \mid \mathbf{M}_k, \mathbf{Z}_{1:t})$ for a map region $\mathbf{M}_k$ given measurement $\mathbf{Z}_1:t$. For instance, using a log-odds representation $l_t(s_k^c \mid \mathbf{M}_k) = \log \frac{P(s_k^c \mid \mathbf{M}_k, \mathbf{Z}_{1:t})}{1-P(s_k^c \mid \mathbf{M}_k, \mathbf{M}_{1:t})}$, the update can be:
\begin{equation}
\begin{aligned}
    l_t(s_k^c \mid \mathbf{M}_k) &= l_{t-1}(s_k^c \mid \mathbf{M}_k) \\ &+ \log \frac{P(s_k^c \mid \mathbf{M}_k, \mathbf{Z}_{t})}{1-P(s_k^c \mid \mathbf{M}_k, \mathbf{Z}_{t})} - l_0(s_k^c \mid \mathbf{M}_k),
\end{aligned}
\end{equation}
where $l_0(s_k^c \mid \mathbf{M}_k)$ is the prior log-odds.
This iterative refinement process regularizes the semantic map by ensuring temporal consistency. A similar principle is applied in S3M-SLAM~\cite{canh2024s3m}, where semantic surfel labels are updated based on new observations.

In addition, scene parsing can be formulated as an energy minimization problem where an energy function defines the desirability of a particular labeling and geometric configuration of objects $\mathbf{L}_t = \{l_1, \cdots, l_N\}$. One effective way to impose regularization is Conditional Random Fields (CRFs)~\cite{du2020accurate, jeon2022rgb}, which model spatial or semantic dependencies between nearby object predictions. Given a set of initial object detections $\{\mathbf{s}_k\}_{k=1}^{N_t}$, the regularized label assignment $\mathbf{L}_t$ can be inferred by minimizing an energy function:
\begin{equation}
    \mathbf{L}_t^* = \arg\min_{\mathbf{L}_t} \sum_k \psi_u(l_k \mid \mathbf{Z}_k) + \sum_{k,k'} \psi_p(l_k, l_{k'} \mid \mathbf{Z}_k, \mathbf{Z}_{k'}),
\end{equation}
where $\psi_u(l_k)$ is the unary potential derived from object detector confidence (\textit{e.g.}, softmax output), and $\psi_p(l_k, l_{k'})$ is the pairwise potential encouraging label smoothness or enforcing contextual compatibility. The pairwise term often captures semantic relationships, such as co-occurrence frequencies (\textit{e.g.}, a keyboard is likely near a monitor) or spatial configurations (\textit{e.g.}, a monitor is typically above a keyboard). In SLAM pipelines, these constraints can be learned from datasets or designed using geometric rules.

Beyond CRFs, recent works incorporate graph neural networks (GNNs) to propagate semantic messages across object nodes. Each object proposal is treated as a node $v_i$ with initial features $f_i$, and a GNN refines these via learned aggregation functions:
\begin{equation}
    f_i' = \gamma\left(f_i, \bigoplus_{j \in \mathcal{N}(i)} \phi(f_i, f_j, e_{ij})\right),
\end{equation}
where $\phi$ is the message function, $e_{ij}$ encodes the edge (\textit{e.g.}, spatial proximity), $\mathcal{N}(i)$ is the neighborhood, and $\gamma$ is the update function.

For example, TopoNet~\cite{pronobis2017learning} and Graph-SLAM~\cite{zhang2024active} use topological graphs over semantic landmarks, improving generalization to novel environments by enforcing object-place co-regularity. Other approaches regularize landmark geometry using learned shape priors (\textit{e.g.}, cubes, quadrics), helping constrain ill-posed object pose estimation problems~\cite{yang2019cubeslam, nicholson2018quadricslam}.

Regularization also improves robustness to dynamic or cluttered environments. For instance, DRV-SLAM~\cite{ji2024drv} applies spatial attention between objects and background regions to refine ambiguous detections. In summary, semantic context regularization bridges low-level noisy detection with high-level scene priors, promoting map consistency and aiding downstream tasks like data association and loop closure.

\section{Stage II: Semantic Localization}\label{sec:localization}

Once semantic information has been extracted from sensor data, the next critical stage in Semantic SLAM systems is Semantic Localization. This process leverages the extracted high-level semantic cues such as object identities, class labels, and scene categories to achieve more robust and accurate camera pose estimation than is possible with purely geometric features. Traditional visual localization relies on matching low-level features like points or lines, which can be unreliable in texture-less environments, under significant viewpoint or illumination changes, or in the presence of perceptual aliasing, where different places look visually similar. Semantic localization addresses these shortcomings by grounding the estimation process in more stable, meaningful, and context-aware landmarks and information.

Formally, given semantic observations $\mathbf{s}_t$ and geometric measurements $\mathbf{Z}_t$, the goal of semantic localization is to estimate the posterior distribution of the robot’s pose $\mathbf{X}_t$:
\begin{equation}
    p(\mathbf{X}_t \mid \mathbf{Z}_{1:t}, \mathbf{S}_{1:t}, \mathbf{U}_{1:t}) \propto p(\mathbf{Z}_t \mid \mathbf{X}_t, \mathbf{s}_t) \cdot p(\mathbf{X}_t \mid \mathbf{X}_{t-1}, \mathbf{U}_t),
\end{equation}
where $\mathbf{u}_t$ represents control inputs. This approach leverages both geometric consistency and semantic coherence to improve localization robustness.

\subsection{Semantic Feature and Landmark-Based Localization} 

One of the first approaches is semantic feature and landmark-based localization. This approach moves beyond low-level points by using semantically meaningful entities as the primary features for tracking and pose estimation. Recognized objects with their rich appearance and geometric properties serve as persistent landmarks. VPS-SLAM~\cite{bavle2020vps} uses a semantic segmentation network to classify detected planes as `floor', `wall', or `ceiling' and cues these large, stable structures as landmarks for robust localization, especially for aerial robots. Similarly, AVP-SLAM~\cite{qin2020avp} is tailored for autonomous valet parking by using robust environmental semantic features like parking lines, guide signs, and speed bumps as landmarks, which are more stable and reliable than point features in texture-poor parking lot environments. Miller \textit{et al.}~\cite{miller2022robust} also demonstrated a system where semantic object detection was used for localization on a free-flying robot, showing improved robustness over feature matching. DRG-SLAM~\cite{wang2022drg} combines point, line, and plane features, using semantics to handle dynamic scenes. Its pose optimization minimizes a weighted sum of error from these different geometric features:
\begin{equation}
    E(T) = \sum E_{point} + \lambda_L \sum E_{line} + \lambda_P \sum E_{plane}.
\end{equation}

In addition, semantics can also guide the matching of traditional low-level features to establish more reliable and longer-term correspondences. Lianos \textit{et al.}~\cite{lianos2018vso} use semantic consistency as a criterion to match points between frames that are temporally distant. This is achieved by projecting a map point into the current frames and searching for matches in a region constrained by the projection and its uncertainty. The semantic label of the map point is used to prune search candidates in the image, allowing for the creation of ``medium-term constraints'' that improve tracking robustness.

\subsection{Segmentation-Guided Localization in Dynamic Environments}

The presence of dynamic objects (\textit{e.g.} pedestrians, vehicles) violates the static world assumption that underpins most traditional visual SLAM algorithms, leading to erroneous feature matching, incorrect pose estimation, and corrupted maps. The most common application is to use segmentation to identify and exclude pixels belonging to dynamic objects from the localization process. Leveraging semantic segmentation has become a cornerstone strategy to address these challenges, enabling robots to distinguish between static and dynamic elements of a scene and thereby improve localization accuracy and robustness. Liu \textit{et al.}~\cite{liu2019visual} propose a semantic supervision method where features are ranked by significance. A feature point $p_i$ belonging to a semantic class $c_i$ is assigned a significance score $R(p_i)$ based on a predefined class weight $W_s(c_i)$:
\begin{equation}
    R(p_i) = W_s(c_i).
\end{equation}
This score, potentially refined by an attention mechanism, helps the system focus on features from more stable classes for more accurate pose estimation. These methods broadly fall into two categories: those that rely purely on semantic priors and those that fuse semantic information with geometric constraints for more reliable motion detection. 

The most direct approach to handling dynamic scenes is to use a pre-trained semantic or instance segmentation network to identify and exclude objects from classes that are 
\textit{a priori} known to be dynamic. In this method, feature points detected on pixels labeled as `person', `car', `bus', \textit{etc.} are simply masked out and not used in the tracking or mapping threads. This method is computationally efficient and straightforward to implement. It can effectively remove a large number of dynamic features with a single forward pass of a segmentation network. However, this method is often too aggressive and can degrade performance by discarding useful information, which are valid static elements that could aid localization. Furthermore, it depends entirely on the segmentation network's accuracy and cannot handle unknown or misclassified moving objects. To overcome the limitations of relying solely on semantic priors, most state-of-the-art systems combine segmentation with geometric motion cues to verify true dynamism. This fusion allows the system to differentiate between a parked car (static) and a moving car (dynamic), leading to more precise and reliable localization. This technique combines semantic segmentation with optical flow. First, a segmentation network identifies regions corresponding to potentially dynamic objects. SOF-SLAM~\cite{cui2019sof} and PSPNet-SLAM~\cite{long2020pspnet} are notable examples that utilize this combined semantic-optical-flow approach to build robust SLAM systems. 

In addition, some researchers employ semantic information to exclusive dynamic objects to enhance localization accuracy in dynamic environments. One of the first attempts~\cite {an2017semantic} proposed a technique utilizing semantic segmentation to identify moving vehicles within autonomous driving contexts, considering them outliers for removal. Based on this, SaD-SLAM~\cite{yuan2020sad}, DGS-SLAM~\cite{yan2022dgs}, Blitz-SLAM~\cite{fan2022blitz}, Dynamic-DSO~\cite{sheng2020dynamic}, DyStSLAM~\cite{li2022dystslam}, and WF-SLAM~\cite{zhong2022wf} also leverage instance segmentation by combining epipolar geometry constraints to eliminate dynamic objects. Excluding features linked to moving objects helps the SLAM system decrease localization inaccuracies caused by these dynamic components. Yet, this strategy might inadvertently eliminate crucial features that could be advantageous for localization in specific contexts. To address this challenge, PLD-SLAM~\cite{zhang2021pld}, YPD-SLAM~\cite{wang2022ypd}, DRG-SLAM~\cite{wang2022drg}, and PLDS-SLAM~\cite{yuan2023plds} combine some novel features, such as point, line, and plane, to enhance the robustness of feature tracking using clustering methods and epipolar geometry. On the other hand, to achieve efficiency in computational complexity, several studies~\cite{lu2020dm, liu2021rds, liu2021rdmo, ji2021towards} extract keyframes and conduct semantic extraction only in these selected keyframes. These papers cluster dynamic points using K-means or random sample consensus (RANSAC)-based or using Bayesian filtering to exclude points with high dynamic probability. Another approach is to reduce the computational cost of deep learning by utilizing a light-weight semantic model. DRE-SLAM~\cite{wang2024survey} and Crowd-SLAM~\cite{soares2021crowd} use YOLOv3/Tiny with clustering or domain-specific training to detect dynamic features and exclude them from tracking. YOLO-SLAM~\cite{wu2022yolo} and CFP-SLAM~\cite{hu2022cfp} leverage depth and epipolar geometry to filter out high-motion regions after YOLOv3/v5 detection. Other systems like DO-SLAM~\cite{wei2023slam}, OVD-SLAM~\cite{he2023ovd}, and MVS-SLAM~\cite{islam2024mvs} extend ORB-SLAM2/3 with YOLOv5/v7 and geometric outlier rejection to improve localization robustness in highly dynamic scenes. 

Furthermore, multi-view geometry provides powerful constraints for identifying dynamic points. For a feature point $p_f$ observation in two frames, its 3D position can be triangulated. Its reprojection error $e_k$ in a keyframe $k$ can be calculated as:
\begin{equation}
    e_k = u_k - \pi_k(T_{kw}, {p_f}_w),
\end{equation}
where $u_k$ is a observed feature location, ${p_k}_w$ is the triangulated 3D point in world coordinates, $T_{kw}$ is the camera pose, and $\pi_k$ is the projection function. For points on dynamic objects, this reprojection error will be large. DynaSLAM~\cite{bescos2018dynaslam} integrates instance segmentation with multi-view geometric verification techniques, then inpainting point clouds in dynamic scenes, whereas MaskFusion~\cite{runz2018maskfusion} proposes a multi-object SLAM based on object reconstruction and multiple moving objects tracking. Then, the optical flow vectors of features within these semantic masks are computed. These vectors are compared against the ego-motion of the camera, which is typically estimated from the motion of the static background. If a feature's motion is inconsistent with the camera's ego-motion, it is confirmed as dynamic and removed. Detect-SLAM~\cite{zhong2018detect} similarly combines an SSD object detector with motion consistency checks. DynaNav-SVO~\cite{contreras2024dynanav} extends this by constructing a region-of-interest (ROI) from \textit{a priori} static urban elements, ensuring reliable feature extraction. DS-SLAM~\cite{yu2018ds} utilizes SegNet for semantic filtering, whereas DynaSLAM. Blitz-SLAM~\cite{fan2022blitz} implements a dual-phase strategy, initially interpreting scenes via deep learning and subsequently confirming them geometrically. CFP-SLAM~\cite{hu2022cfp} uses a hierarchical method grounded in object detection and motion classification, while SG-SLAM~\cite{cheng2022sg} combines semantic interpretation with geometric constraints within a graph-based framework. Zhang~\textit{et al.}~\cite{zhang2025adaptive} evaluate scene quality based on semantic extraction to optimize camera pose estimation. Dynamic elements $d_t$ can corrupt pose and map estimation. We define a binary mask $M_t$ indicating static regions:
\begin{equation}
    d_t' = M_t \odot d_t,
\end{equation}
where $\odot$ is the element-wise product. However, once dynamic objects are masked, they leave ``\textit{holes}'' in the image. To address this, frameworks like DynaSLAM~\cite{bescos2018dynaslam}, IDV-SLAM~\cite{qian2023visual}, and Empty Cities~\cite{bescos2020empty} use background inpainting techniques, often powered by generative adversarial networks (GANs)~\cite{goodfellow2020generative}, to fill in the regions occluded by dynamic objects. This restores a complete and static view of the scene, providing a more stable input for subsequent visual odometry and mapping tasks. RDS-SLAM~\cite{liu2021rds} addresses the challenge of integrating segmentation into a real-time SLAM pipeline by running the semantic segmentation module in a separate, parallel thread. The main tracking thread continues to operate at high frequency, while the segmentation thread periodically provides masks of dynamic regions. This information is then used in the mapping and optimization threads to filter out dynamic map points, preventing map corruption without sacrificing real-time tracking performance. 

While traditional dynamic SLAM methods often treated moving objects as outliers to be excluded, recent research trends emphasize tracking dynamic objects and estimating their motion jointly with camera poses. DynaSLAM II~\cite{bescos2021dynaslam} and Dynamic SLAM~\cite{wang2024survey} jointly optimize camera and object trajectories using semantic segmentation and motion models, generating both static and dynamic maps. DOT-SLAM~\cite{ballester2021dot} and PointSLOT~\cite{zhou2023pointslot} apply instance segmentation and object-based BA to track rigid object motions. Several systems combine semantic instance segmentation, optical flow, and photometric consistency for multi-object tracking and motion estimation~\cite{zhang2020robust}. For example, MOTSLAM~\cite{zhang2022motslam}, DE-SLAM~\cite{xing2022slam}, and TwistSLAM~\cite{gonzalez2022twistslam} use joint optimization frameworks that fuse semantic, geometric, and motion cues to estimate object and camera poses. SLAMANTIC~\cite{schorghuber2019slamantic} and VDO-SLAM~\cite{zhang2020vdo} use dynamic confidence and motion constraints to enhance map reliability. More recent approaches like ClusterVO~\cite{huang2020clustervo} and OTE-SLAM~\cite{chang2023ote} integrate tracking algorithms ByteTrack~\cite{zhang2022bytetrack} with semantic object detection, while also estimating dynamic object poses through twist parameterization or joint pose graphs. This provides a much richer understanding of the environment, which is essential for applications like collision avoidance and behavior prediction in autonomous driving.

\section{Stage III: Semantic Mapping}\label{sec:mapping}
While semantic extraction provides the raw semantic labels for objects, regions, and places, Semantic Mapping is 
concerned with integrating this information into a persistent, structured, and queryable representation of the environment. As discussed in the review by Mascaro and Chi~\cite{mascaro2024scene}, building a rich and actionable model of the world is a crucial capability for autonomous systems, moving beyond simple geometric point clouds to denser, semantically-aware representations. This section reviews how semantic maps are represented (\ref{sec:semanticmaprepre}), how semantic information is integrated and fused into these maps (\ref{sec:semanticfusion}), and how to evaluate semantic uncertainty and formulate cost functions (\ref{sec:semanticcost}).

\subsection{Semantic Map Representations} \label{sec:semanticmaprepre}
The choice of semantic map representations is crucial as it dictates how information is stored, updated, accessed, and utilized by the robot. Different representations offer trade-offs in terms of expressiveness, scalability, memory efficiency, and suitability for particular tasks. Unlike geometric maps (\textit{e.g.}, point clouds, occupancy grids), semantic maps encode high-level scene information. A semantic map can be formally represented as a tuple:
\begin{equation}
    \mathcal{M}_i = \{ (\mathbf{X}_i, \mathcal{F}_i, \mathbb{s}_i) \}_{i=1}^N,
\end{equation}
where $X_i \in SE(3)$ denotes the pose of an element, $\mathbf{s}_i$ is the semantic information, and $\mathcal{F}_i$ is set of observed features.

\subsubsection{Object-Centric Maps} explicitly model the environment as a collection of distinct object instances. Each object is typically stored with its semantic class label, 3D pose (position and orientation), dimensions or shape parameters, and potentially other attributes like appearance features or functional affordances. In many SLAM systems, distinct and recognizable objects serve as semantic landmarks. These landmarks are associated with their 3D positions and class labels in the map. Civera \textit{et al.}~\cite{civera2011towards} proposed an early Semantic SLAM that merged traditional point features with known 3D object models, where objects, once recognized, were inserted and tracked in the EKF SLAM map. SLAM++~\cite{salas2013slam++} used a 3D CAD model to identify and render objects, building a map composed of full 3D object geometries. However, these systems are constrained by their reliance on pre-defined object templates. Several studies~\cite{zhang2018semantic, sharma2021compositional} estimate object poses and incorporate them into the map, effectively creating an object-level representation. However, they often incur a high computational cost and pose a challenge to achieving the real-time performance of the system. These frameworks focus on object SLAM by modeling objects as dual quadric formulation~\cite{nicholson2018quadricslam, hosseinzadeh2019real}, rectangular bounding volumes~\cite{yang2019cubeslam}, specific objects~\cite{chen2022accurate}, and estimating fine geometric models of objects using a CNN trained~\cite {hosseinzadeh2019real}. Dynamic object-level SLAM has been considered in~\cite{kochanov2016scene, cheng2022sg}. Object-oriented semantic mapping, for example by Sünderhauf \textit{et al.}~\cite{sunderhauf2017meaningful} and Canh \textit{et al.}~\cite{canh2023object} aims to create a meaningful map that contains instances of known objects. These methods provide object information for robotics tasks such as navigation or obstacle avoidance. To improve the performance of these tasks, Mascaro \textit{et al.}~\cite{mascaro2022volumetric} presents an individual object-level semantic mapping pipeline by integrating 3D instance segments and diffusion scheme refinement. Hu~\cite{hu2023multi} also introduces a multi-level map, such as inaccurate object modeling and limited object representation based on geometric map, plane map, and object map, particularly in constructing long-term consistent maps. Li \textit{et al.}~\cite{li2020textslam} model scene text as planar semantic features, effectively creating a map of text objects with geometric and semantic properties. On the other hand, Fusion++ uses 2D instance mask predictions and fuses them into the TSDF reconstruction within globally consistent loop-closed object SLAM maps. Voxblox++ also combined geometric segmentation, instance-aware segmentation refinement, data association, and map integration to create an object-centric map. Following this line of research, TSDF++~\cite{grinvald2021tsdf++} introduces a multi-object mapping approach with separate reconstruction volumes for each object based on the TSDF formulation. The studies conducted by Asgharivaskasi and Atanasov~\cite{asgharivaskasi2021active, asgharivaskasi2023semantic} stand out as they introduce a multi-class (semantic) OctoMap. These works employ a closed-form lower bound on the Shannon mutual information between the map and range-category observations to determine robot trajectories that are rich in information.

\subsubsection{Volumetric Semantic Maps} discretize 3D space into elementary units (point, surfel, or voxels) and store semantic information within each unit.

\paragraph{Voxel-Based Semantic Maps} These extend traditional occupancy grids like OctoMap by associating each occupied voxel (or OctoMap leaf node) not just with an occupancy probability, but also with a semantic label or a probability distribution over multiple semantic classes. Stückler \textit{et al.}~\cite{stuckler2012semantic} proposed a Bayesian framework to fuse probabilistic object-class segmentations from multiple RGB-D views into such a voxel-based 3D semantic map. For a voxel $v$ and class $s^c$, the posterior probability $P(s^c \mid v, Z_{1:t}$ given measurements $Z_{1:t}$ is updated recursively:
\begin{equation}
    P(s^c \mid v, \mathbf{Z}_{1:t}) \propto P(s^c \mid v, \mathbf{Z}_t)\frac{P(s^c \mid v, \mathbf{Z}_{1 \cdots t-1}}{P(s^c \mid v)}.
\end{equation}
Kochanov \textit{et al.}~\cite{kochanov2016scene} present a probabilistic mapping approach, which uses recursive Bayesian filtering to update voxel occupancy and semantic labeling based on observation and scene flow~\cite{shi2021rgb, sun2022volumetric}. Other methods~\cite{zhang2018semantic, ran2021rs} also create an ``improved Octomap'' where voxels can store semantic object information. Dyns-SLAM~\cite{barsan2018robust} uses semantic and sparse flow cues to identify and classify dynamic objects and uses voxel block hashing for large-scale reconstruction. CNN-SLAM~\cite{tateno2017cnn} predicts depth map using a CNN network and creates a dense semantic map based on the retraining of the network for semantic segmentation using soft-max later and cross entropy function. Kimera~\cite{rosinol2020kimera} introduced the adaptation of bundled raycasting for building a global 3D mesh using Voxblox and a TSDF model and semantically annotating it using 2D semantic labels, label propagation, and Bayesian updates. Li \textit{et al.}~\cite{li2018dense} build a dense map upon ORB-SLAM and a semi-global stereo matching algorithm for disparity map generation. Shi \textit{et al.}~\cite{shi2021rgb} introduced a ``label-oriented voxelgrid filter'' that ensures intra-frame spatial continuity and inter-frame spatiotemporal consistency when fusing 2D semantic labels into a 3D voxelized map. Liu \textit{et al.}~\cite{liu2024fm} integrate vision-language foundation models to improve the construction of a generalizable instance-aware semantic map. 
Qian \textit{et al.}~\label{qian2022pocd} introduces a novel probabilistic object state representation that models both stationarity and magnitude of geometric change for each object and a Bayesian update rule that incorporates geometric and semantic information for consistent map maintenance.

\paragraph{Surfel-Based Semantic Maps} Surfel representations model surfaces using a collection of 3D discs or ellipses. Semantic information can be attached to each surfel. Stuckler \textit{et al.}~\cite{stuckler2015dense} integrate object-class segmentation, SLAM, and semantic 3D fusion into a real-time operating semantic mapping system based on multi-resolution surfel maps (MRSMap). Canh \textit{et al.}~\cite{canh2024s3m} in S3M, build a semantic sparse map for UAVs using a semantic surfel cloud, where each surfel $sf_k$ stores geometric attributes (position $p_k$, normal $n_k$, radius $r_k$) and a semantic label distribution $L_k = \{ P(s^c_1 \mid sf_k), \cdots, P(s^c_N \mid sf_k) \}$. The semantic label is updated based on projected 2D segmentations based on the current frame's segmentation $L_k^{obs}$:
\begin{equation}
    L_k^{new} = \eta \cdot L_k^{obs} + (1-\eta)\cdot L_k^{old}.
\end{equation}
Reddy~\cite{reddy2015dynamic} combines classical semantic segmentation and motion constraints to separate dynamic objects from static scenes, which starts with the computation of low-level features like SIFT descriptors, optical flow, and stereo disparity, and reconstructing the environment. Building on a similar idea, MaskFusion~\cite{runz2018maskfusion} tracks multiple moving objects even when they move independently from the camera and reconstructs them as surfel clouds. SemanticFusion~\cite{mccormac2017semanticfusion} and Co-Fusion~\cite{runz2017co} utilize CNNs and state-of-the-art dense SLAM - ElasticFusion to predict semantics from multiple viewpoints and fuse them into a surfel map using a Bayesian update scheme. Morreale \textit{et al.}~\cite{morreale2019dense} present dense visual mapping, characterized by two main trends: high accuracy and dense mapping from sparse data. This approach begins with the highest free-space vote and iteratively collects free-space tetrahedra, ensuring manifold properties through four semantic simplification approaches. Seichter \textit{et al.}~\cite{seichter2022efficient} propose incorporating semantic information into NDT (normal distributions transform) maps. Each cell in an NDT grid models the local surface as a Gaussian distribution, 
representing the probability of finding a point at a certain location. In a semantic NDT map, each cell stores a separate NDT for each semantic class, enabling more accurate and semantically consistent mapping.

\paragraph{Panoptic Semantic Maps} To address the limitation of lack the ability to distinguish individual instances belonging to the same category in higher-level semantic understanding, panoptic representations aim to integrate class labels for background areas (stuff), while simultaneously segmenting and identifying distinct foreground objects (things) one by one. Nakajima \textit{et al.}~\cite{nakajima2018fast} describes the four components of the method: dense approach of InfiniTAMv3~\cite{prisacariu2017infinitam} for SLAM, 2D semantic segmentation with a specifically designed CNN, incrementally building a geometric 3D map, and updating class probabilities assigned to each segment of the geometric 3D map. PanopticFusion~\cite{narita2019panopticfusion} introduces a novel online volumetric semantic mapping system that densely predicts class labels of stuff and things based on 2D panoptic label prediction, panoptic label tracking, thing label probability integration, and online map regularization. Pham \textit{et al.}~\cite{pham2019real} designs an inference-optimal segmentation from predictions of a DNNs, then progresses super-voxel clustering to achieve a real-time dense reconstruction of 3D indoor scenes. Yang and Liu~\cite{yang2021tupper} implemented geometric segmentation to discover novel scene elements and refine them by panoptic segmentation results using UPSNet~\cite{xiong2019upsnet}, then associate temporal data based on VPSNet~\cite{kim2020video}. Panoptic Multi-TSDFs~\cite{schmid2022panoptic} introduced a hierarchical map representation and temporal hierarchy based on panoptic entities structured as submaps, then integrated measurements into active submaps using projective updates and TSDF weighting. PanopticNDT~\cite{seichter2023panopticndt} presented an efficient and robust panoptic mapping approach based on occupancy NDT mapping by combining a subsequent mapping stage and a panoptic segmentation stage using EMSANet~\cite{seichter2022efficient}, which includes semantic and instance prediction, fusion, and confidence score generation. PanoRecon~\cite{wu2024panorecon} addressed a challenge of realized online geometry reconstruction along with dense semantic and instance labeling by incrementally performing 3D geometric reconstruction and 3D panoptic segmentation in a view-independent 3D feature volume. EPRecon~\cite{zhou2024eprecon} achieved real-time performance by integrating a lightweight module for depth prior estimation in 3D volume and a novel panoptic features extraction strategy that combines voxel features and image features. Panoptic-SLAM~\cite{abati2024panoptic} and PS-SLAM~\cite{li2025ps} build upon ORB-SLAM3 and use panoptic segmentation, including the categorization of objects as ``things'' and ``stuff'', and handling of unknown objects to filter dynamic objects in dynamic environments.

\subsubsection{Topological-Semantic Maps} represent environments as graphs where nodes correspond to places or regions (\textit{e.g.}, rooms, corridors) and edges signify connectivity or traversability. This representation is particularly useful for high-level path planning and human-robot interaction. Bernuy and Solar~\cite{bernuy2015semantic} define the topological semantic map (TSM) as a graph structure describing the road and its environment, including type, spatial position, objects list, traversability index, curvature index for road nodes, and odometry based on the semantic description of images. Some studies~\cite{lin2018topological, bernuy2018topological, tian2022vision} also address the challenge of autonomous driving applications by leveraging semantic observation of the environment, a TSM for storing elected semantic observations, and a topological localization algorithm which uses a Particle Filter for obtaining the vehicle's pose in the TSM. Zhao \textit{et al.}~\cite{zhao2018indoor} utilized DNNs to identify indoor scenes without training in a specific environment and estimate the semantic region. Topomap can improve the performance of navigation tasks by simplifying to reduce computation-intensive path planning algorithms and providing sufficient knowledge about traversable spaces~\cite{blochliger2018topomap, chen2021topological, kim2023topological}. 3D scene representation or 3D scene understanding is one of the fundamental and important problems in robotics. Several studies ~\cite{armeni20193d,kim20193,wu2021scenegraphfusion,mehan2024questmaps, gu2024conceptgraphs} focus on this challenge by creating a 3D scene graph to construct entire buildings, rooms, and include semantics on objects (\textit{e.g.}, shape, class, material). Armeni \textit{et al.}~\cite{armeni20193d} represent semantic information in a 3D four-layer structured scene graph based on 3D mesh models, registered RGB panoramas, and camera parameters. Another 3D scene graph for intelligent agents is proposed by Kim \textit{et al.}~\cite{kim20193}, which integrates a data processing module: adaptive blurry image rejection (ABIR) based on the variance of Laplacian and adaptive thresholding, keyframe group extraction (KGE) for classifying image frames into keyframes, anchor frames and garbage frames to reduce redundancy, and spurious detection rejection (SDR) using 3-D position information, word2vec semantics, and rule-based and local/global scene graph construction as a directed graph with nodes representing objects and edges representing relations. Wu \textit{et al.}~\cite{wu2021scenegraphfusion} create a globally consistent scene graph by fusing node feature extraction, edge feature computation, graph neural network (GNN) feature propagation, and class prediction based on a feature-wise attention (FAT) mechanism, which re-weights individual latent features at each target node to handle missing neighboring points. Mehan \textit{et al.}~\cite{mehan2024questmaps} present indoor environment as a multi-channel occupancy representation based on room mask prediction and transition region detection, formulating it as a 3-class instance segmentation task using Mask R-CNN architecture. Gu \textit{et al.}~\cite{gu2024conceptgraphs} create object-based 3D mapping by class-agnostic 2D segmentation and estimating spatial relationships between objects by calculating the 3D bounding box IoU to obtain a similarity matrix. Soucsa and Bassani~\cite{sousa2022topological} consolidate deep visual features extracted by GoogLeNet~\cite{szegedy2015going} and shallow Multilayer Perceptron (MLP) to create representations of regions and use a moving average and visual habituation mechanism. Yang \textit{et al.}~\cite{yang2022automated} parse and vectorize indoor architecture from floor-plan raster maps, which leverage a learning-based hierarchical approach to identify a set of geometric primitives with semantics and use of mixed integer programming (MIP) to fuse primitives and their relationship information into vector graphics while enforcing high-level structural constraints. Upon this idea, Cao \textit{et al.}~\cite{cao2024context} presents a Context-Enhanced Full-Resolution Network (CEFRN) for floor-plan segmentation to improve the accuracy of topological semantic maps for the Partially Sighted or Visually Impaired (PSVI). In addition, some studies~\cite{lin2021topology, cao2024semantictopoloop} use the topology of the object-level map to achieve the effectiveness and robustness of the loop closure process. Fredriksson \textit{et al.}~\cite{fredriksson2023semantic} identify and classify intersections for semantic and topological mapping, then find paths between intersections, dead ends, and pathways to unexplored areas to generate the robot paths. SENT-Map~\cite{kathirvel2025sent} also provides exciting potential for robots operating in complex human environments. PRISM-TopoMap~\cite{muravyev2025prism} is a recent example of an online topological mapping method that uses learnable multimodal place recognition to build and maintain the graph structure without relying on global metric coordinates.

\subsubsection{Hybrid and Hierarchical Maps} employ hierarchical maps that combine the strengths of different representations across multiple levels of abstraction, like metric, semantic, and topological, to bridge the gap between detailed and strategic planning. For example, executing high-level directives such as \textit{``go to the living room and bring the coffee mug I left on the table''} necessitates developing a functional world model that comprehends the associations among semantic entities and integrates real-world details across varying levels of abstraction. The first attempt by Tomatis \textit{et al.}~\cite{tomatis2003hybrid} integrates metric and topological paradigms in a hybrid system for both localization and map building using two levels of abstraction in combination with the different types of environmental structures. A similar approach by Kuipers \textit{et al.}~\cite{kuipers2004local} defines four distinct representations of spatial knowledge in the Spatial Semantic Hierarchy (SSH) for building the tree of topological maps on the basis of SSH. Galindo \textit{et al.}~\cite{galindo2005multi} introduce a multi-hierarchical approach including the spatial hierarchy, which contains spatial and metric information from the robot environment, and the conceptual hierarchy, which models semantic knowledge. The correspondence between symbols and sensor data of these hierarchies is created and maintained by anchoring that refers to the same physical object. Drouilly \textit{et al.}~\cite{drouilly2015hybrid} develop a new scheme for updating maps of a large-scale dynamic environment using stable semantic representation, which updates semantic layers of the map directly and models changes in terms of static class occlusions caused by dynamic objects. Yang \textit{et al.}~\cite{yang2017semantic} introduce the use of CRFs for joint optimization of grid labels and present the formulation of a hierarchical CRF model incorporating unary, pairwise, and high-order potential. The unary potential is defined as the negative logarithm of prior probability computed based on 2D semantic segmentation from a dilated CNN. The pairwise potential is modeled as a combination of Gaussian kernels based on color similarity and 3D grid position. This high-order term is formulated using a robust $P^n$ Potts model~\cite{kohli2007p3} to enforce label consistency within superpixel-defined cliques. This high-order term is transformed into a hierarchical pairwise graph with auxiliary clique variables. Malleson \textit{et al.}~\cite{malleson2018hybrid} introduce a hybrid 4D model approach combining prior surfel graph modeling and higher resolution of volumetric fusion based on pairwise data and smoothness costs, label cost, intra-part fusion, and inter-part fusion. Luo and Chiou~\cite{luo2018hierarchical} combine a 2D occupancy grid map with an overlaid topological graph, which is organized by a hierarchical semantic organization structure, linking abstract concepts with tangible objects, while Wen \textit{et al.}~\cite{wen2020hybrid} introduce 3D semi-dense based on Mask R-CNN for navigation and semantic information storage. Yue \textit{et al.}~\cite{yue2020hierarchical} propose a hierarchical framework for collaborative semantic 3D mapping based on multimodal semantic information fusion and collaborative semantic map fusion using the EM algorithm. Building upon this idea, Deng \textit{et al.}~\cite{deng2022hd} involve constructing a graph model and apply a label diffusion method to create an accurate hierarchical and dense semantic occupancy map. 
Rosinol \textit{et al.}~\cite{rosinol2021kimera} combine 5 layers and hierarchical representation, including metric-semantic, objects and agents, place and structures, rooms and buildings into a single model, a 3D scene graph. Zhang \textit{et al.}~\cite{zhang2021building} introduce a human-inspired object search strategy that builds a metric-topological map and selects the optimal search node by considering object co-occurrence relationships and robot location. Recent papers by Hughes \textit{et al.}~\cite{hughes2022hydra, hughes2024foundations} present a real-time spatial perception system for incremental construction of scene graph layers, including local mesh and object Euclidean Signed Distance Field (ESDF) generation, topological place subgraph extraction using a Generalized Voronoi Diagram, and room segmentation using a community-detection approach. H2-Mapping~\cite{jiang2023h} and its improved version H3-Mapping~\cite{jiang2024h3} leveraged a hierarchical hybrid representation with a quasi-heterogeneous feature grid for real-time performance and high-quality construction, reducing redundant feature grid allocation and improving the training efficiency of texture with limited sampling and training time. Wald \textit{et al.}~\cite{wald2020learning} presented a Scene Graph Prediction Network (SGPN) for generating a graph describing objects (nodes) including a class hierarchy and a set of attributes describing visual and physical appearance and their relationships (edges) with different categories of relationships including spatial/proximity relationships, support relationships, and comparative relationships. Khronos~\cite{Schmid2024Khronos} formally solved the Spatio-Temporal Metric-Semantic SLAM (SMS) problem, which aims to construct a dense metric-semantic model of the world at all times. It enables long-term robot autonomy by local estimation via an active window, as well as global optimization using Truncated Least Squares (TLS) loss with Graduated Non-Convexity (GNC)~\cite{yang2020graduated} in GTSAM~\cite{gtsam} for estimating the complete state of the scene over time. Following this approach, Clio ~\cite{maggio2024clio} defined the task-driven 3D scene understanding problem, where the robot builds a compressed map representation and dynamically selects the level of detail and objects for its maps based on the task. This method utilized FastSAM~\cite{zhao2023fast} and CLIP~\cite{radford2021learning} for segmentation, temporal association, and 3D reconstruction, and built upon Hydra~\cite{hughes2022hydra} to construct 3D place primitives and describe how CLIP embeddings are associated with places.

\subsection{Semantic Fusion through Multiple Observations} \label{sec:semanticfusion}
A core challenge in semantic mapping is to distill noisy, often conflicting, per-frame semantic predictions into a single, coherent, and globally consistent 3D semantic map. Semantic fusion is the process that addresses this challenge by aggregating information from multiple viewpoints and observations over time. This process is fundamentally reliant on multi-view consistency, where a robust underlying SLAM system provides accurate camera poses and establishes correspondences between frames. These correspondences allow semantic labels for the same 3D points or regions, observed from different perspectives, to be associated (in the data association step) and then fused, thereby reducing the uncertainty and correcting errors from individual predictions.

The dominant approach for fusing semantic labels in dense representation (\textit{e.g.,} voxels or surfels) is probabilistic. This method treats the labels of each map element as a random variable and uses new observations to update its class probability distribution iteratively. A foundational technique involves recursive Bayesian update~\cite{stuckler2015dense}. For each map element $v$, a probability distribution $P(s^{cf}_v)$ over the set of semantic classes $s^c$ is maintained. When a new observation $\mathbf{Z}_t$ is registered, the semantic predictions from a CNN at the corresponding pixel location are used as the likelihood $P(Z_t \mid s^{cf}_v = s^c)$ to update the prior probability $P(s^{cf}_v = s^c)$. In log-odd form, this update for each class $s^c$ is:
\begin{equation}
    s^{cf}_{t, v, c} = s^{cf}_{t-1, v, c} + \log \frac{P(s^{cf}_v = s^c \mid \mathbf{Z}_t)}{1-P(s^{cf}_v = s^c \mid \mathbf{Z}_t)}.
\end{equation}

This fusion over multiple independent observations effectively averages out noise from the segmentation network and leads to a more accurate and stable semantic map. This principle is widely used in systems like SemanticFusion~\cite{mccormac2017semanticfusion}, as well as in other voxel-based and segment-based maps. Fusing panoptic information is more complex as it must handle both ``stuff'' (amorphous regions like roads, walls) and ``thing'' (countable object instances) categories. PanopticFusion~\cite{narita2019panopticfusion} achieves this by maintaining separate probability distributions for semantic class (stuff) and instance IDs (things) for each voxel. For a voxel $v$, the semantic distribution $P_{\mathbf{s}}(s^c \mid v)$ is updated with a standard Bayesian update. The instance distribution $P_I(i \mid v)$ for instance $i$ is updated differently to prevent instances from blending into each other. Specifically, it uses a non-normalized update and then assigns the voxel to the single instance with the highest probability. The update, for instance, $i$ at voxel $v$ given observation $Z_t$ is:
\begin{equation}
    P_t(i \mid v) = P(\mathbf{Z}_t \mid i)P_{i-1}(i \mid v).
\end{equation}
In addition, several articles~\cite{barsan2018robust, runz2018maskfusion, mccormac2018fusion++, hoang2020object, grinvald2021tsdf++, kabalar2023towards, Schmid2024Khronos} use the IoU tracker~\cite{bochinski2017high}, and define the overlap threshold to match predicted objects from the previous frame to the current frame. To enhance the ability of this method, some works~\cite{grinvald2019volumetric,mascaro2022volumetric} introduce a 3D IOU score, which calculates a pairwise of instance segments $s^c_i$ with the number of points $|s_i|$ and object instance $\mathbf{s}_o$:
\begin{equation}
    IOU(s^c_i, s^o) = \frac{\Pi(s^c_i, \mathbf{s}_o)}{|s_i|+\Pi(s^c_o, \mathbf{s}_o) + \sum_{i \neq i'}\Pi(s_{i'}^c, \mathbf{s}_o)}.
\end{equation}
where $\Pi(s^c_i, s^o)$ is the total number of points belonging to object $\mathcal{s}_o$. Besides that, Mascaro \textit{et al.}~\cite{mascaro2022volumetric} propose a technique that first performs probabilistic fusion and then refines the result via label diffusion on a voxel graph. They construct a k-Nearest Neighbor (k-NN) graph of the voxels and formulate an energy minimization problem to find a consistent labeling $s^{cf}$. The energy function balances fidelity to the observed data with spatial smoothness:
\begin{equation}
    Q(s^{cf}) = \sum_{i\in \mathcal{C}}(p^r_i - p^f_i)^2 + \mu \sum_{i, j\in N_k} w_{ij} \Big (\frac{p^r_i}{\sqrt{d_i}} - \frac{p^r_j}{\sqrt{d_j}}\Big )^2,
\end{equation}
where $p^f_i$ is the initial fused probability for an observed voxel $i$, the set $\mathcal{C}$, $p^r_i$ is the refined distribution, and the second term regularizes the solution by encouraging neighboring voxels to have similar labels, normalized by their degree $d$. This diffusion process propagates strong semantic predictions from confident regions to less certain or unobserved regions, resulting in a more complete and consistent semantic map. Iro \textit{et al.}~\cite{armeni20193d} use a voting scheme for multi-view consistency, which defines the weight for each frame $f_i$ based on robot position $\mathbf{x}_i$ as $w_{i,j} = \frac{\sum_{i,j} ||\mathbf{x}_i - f_{cj}||}{||\mathbf{x}_i - f_{cj}||}$. However, this method does not consider the uncertainty of objects, especially in a complex environment. Popular methods~\cite{yang2017semantic, antonello2018multi, nakajima_fast_2018, ran2021rs, liu2024fm, canh2024s3m} include Bayesian updates or other probabilistic aggregation that recursively combine new observations $Z_t$ with prior belief:
\begin{equation}
    P(\mathbf{s} \mid v, \mathbf{Z}_{1:t}) \propto P(\mathbf{s} \mid v, \mathbf{Z}_t) \cdot \frac{P(\mathbf{s} \mid v, \mathbf{Z}_{1:t-1})}{P(\mathbf{s} \mid v)},
\end{equation}
where $P(\mathbf{s} \mid v, \mathbf{Z}_t)$ is the semantic classifier output. This fusion reduces uncertainty and corrects individual frame errors, yielding coherent semantic maps. These studies~\cite{kim20193, qian2022pocd, gu2024conceptgraphs} construct a pair of semantic and geometric similarity with respect to all objects in the map by calculating position similarity score $R_{c^p}$ and color similarity score $R_{c^c}$ and cut-off by a predefined threshold $\theta_{cutoff}$ and find the final fusion by an optimal strategy like Hungarian algorithm or greedy assignment. Alternative fusion methods include learned fusion networks~\cite{tateno2017cnn}, splat rendering for projective fusion, and an update scheme similar to previous surfel-based approaches~\cite{runz2017co}, self-supervised feature fusion using deep vision transformers (DINO, CLIP), and generating open-set instance clusters and refining label certainty with repeated observation~\cite{singh2024loss, jatavallabhula2023conceptfusion}. These methods also enforce spatial and temporal label consistency.

\subsection{Semantic Uncertainty and Cost Function} \label{sec:semanticcost}

Beyond guiding feature selection, semantic information can be integrated directly into the mathematical core of the localization and mapping process. This is achieved by either (1) formulating probabilistic observation models that explicitly account for semantic uncertainty or (2) designing novel cost functions for optimization-based SLAM that penalize semantic and geometric inconsistencies. These two concepts are deeply intertwined, as the uncertainty model often informs the weighting of terms in the cost function.

\subsubsection{Probabilistic Observation Models and Semantic Uncertainty} 

In probabilistic localization frameworks (\textit{e.g.}, particle filters, Kalman filters), beliefs about the robot's state are updated using an observation likelihood model, $P(\mathcal{Z} \mid\mathcal{X}, \mathcal{M})$, which quantifies the probability of making a measurement $\mathbf{Z}$ for a pose $\mathbf{X}$ and map $\mathbf{M}$. Semantic localization enriches this by incorporating semantic measurements $\mathcal{S}$. However, semantic classifiers are imperfect; they produce probabilistic outputs and can make errors. Therefore, robustly handling the uncertainty of these semantic predictions is crucial. A naive approach might only use the highest probability class prediction, discarding valuable uncertainty information~\cite{kochanov2016scene}. A more principled approach is to model the uncertainty of the entire class probability vector. Akai \textit{et al.}~\cite{akai2020semantic} propose a method that achieves this using Dirichlet distributions, which are conjugate priors to the categorical distribution used for classification. This allows the localization framework to naturally down-weight semantic measurements with high uncertainty (\textit{e.g.}, when an object is ambiguous or partially occluded), leading to more robust pose estimation. The Dirichlet distribution $\textbf{Dir}(\theta \mid \boldsymbol{\alpha})$ over a simplex of class probabilities $\theta$  and $\boldsymbol{\alpha}$ is the concentration parameters of the Dirichlet. In their framework, when an object is observed, the classifier outputs a probability vector $\pi_k = (\pi_{k,1}, \cdots,\pi_{k,C})$ over $C$ classes. The core idea is to calculate the likelihood of observing this specific vector $\pi_k$, given the predicted probabilities from the map at the robot's hypothetical pose. This likelihood, $P(\pi \mid \mathcal{M}, \mathbf{X}_t)$, is used to update the weight of the particle filter. It is calculated by marginalizing over all possible tree class distributions $\theta_k$:
\begin{equation}
P(\pi_k \mid \mathcal{M}, \mathbf{X}_t) = \int p(\pi_k \mid\theta_k) p (\theta_k \mid \mathcal{M}, \mathbf{X}_t) d\theta_k.
\label{eq:dirichlet}
\end{equation}

Assuming the likelihood $p(\theta_k \mid \mathcal{M}, \mathbf{X}_t)$ follows a Dirichlet distribution with parameters $\alpha_k$, this integral has a closed-form solution:
\begin{equation}
    P(\pi_k \mid \mathcal{M}, \mathbf{X}_t) = \frac{\Gamma\left(\sum_{c=1}^C \alpha_{k,c}\right)}{\prod_{c=1}^C \Gamma(\alpha_{k,c})} \prod_{c=1}^C \pi_{k,c}^{\alpha_{k,c} - 1}
\end{equation}
where $\Gamma(\cdot)$ is the Gamma function, and the Dirichlet parameters $\alpha_{k,c}$ are set based on the map's prediction $p_m(c)$ for that location and a concentration parameter $\lambda$ that reflects confidence: $a_{k,c} = \lambda p_m(c) +1$.
In addition, active mapping aims to plan a robot's path to reduce the map's uncertainty most effectively. Lu \textit{et al.}~\cite{lu2024semantics} quantify the semantic uncertainty of mapped semantic object instance $\mathbf{s}_i$ using the entropy of its fused class probability distribution $p(s^c|\mathbf{s}_i)$. The total semantic uncertainty of the map $\mathcal{M}$ is the sum of entropies over all objects:
\begin{equation}
    \mathbb{H}(\mathcal{M}) = -\sum_{\mathbf{s}_i \in \mathcal{M}} \sum_{s^c\in C} p(s^c \mid \mathbf{s}_i) \log p(s^c \mid \mathbf{s}_i).
\end{equation}

\subsubsection{Semantic Cost Functions}

In optimization-based SLAM, camera poses are typically found by minimizing a cost function, often a sum of geometric reprojection errors. A semantic cost function augments this objective with terms that penalize semantic inconsistencies between the current observation and the map. VSO~\cite{lianos2018vso} introduces a semantic consistency term into the optimization. In addition to the standard geometric reprojection error, it adds a semantic error term that penalizes assigning a 3D map point to a semantic class that is inconsistent with the segmentation of the image. The total cost function over a set of map points:
\begin{equation}
    E = E_{\mathrm{geo}} + \lambda E_{s},
\end{equation}
and $E_s$ captures reprojection error in the semantic label space:
\begin{equation}
\begin{aligned}
    E_s = &\sum_{k, i} e_s(k, i), \\ 
    &e_s(k, i) = \sum_{c \in C} w_i(c) \cdot log (p(\mathbf{s}_k \mid \mathbf{X}_k, \mathbf{M}_i, s_k^c=c)), 
\end{aligned}
\end{equation}
where $w_i(c)$ the class weight for landmark $i$. Direct semantic alignment approaches optimize over semantic probability maps, which can be more robust to lighting changes than photometric error alone~\cite{lianos2018vso}. Instead of merely aligning pixel intensities, methods like SDVO \cite{bao2022semantic} are also based on the direct semantic alignment approach, which proposes aligning semantic probability maps generated by segmentation networks, which are less susceptible to illumination changes. The semantic alignment error $E_{sem}$ can be formulated as:
\begin{equation}
    E_{s} = \sum_{i \in \mathcal{P}} w_{s^c} ||\mathbf{P}_{s}(i) - \mathbf{P}'_{s}(i')||_{\gamma},
\label{eq:semantic_alignment}
\end{equation}
where $\mathbf{P}_{s}(i)$ is the semantic probability vector at pixel $i$ in the current frame, $\mathbf{P}'_{s}(i')$ is the projected semantic probability vector from the map at the corresponding pixel $i'$, $w_{s^c}$ is the heuristic weighting factor for the semantic $s^c$, and $||\cdot||_{\gamma}$ is the Huber norm.  CNN-SLAM~\cite{tateno2017cnn} enhances a direct SLAM method - LSD-SLAM~\cite{engel2014lsd} by integrating a depth map predicted by a CNN. The system's energy function includes a term that penalizes deviations from the CNN's depth prediction, but is weighted by the uncertainty of that prediction. This allows the system to trust the geometric photometric alignment in high-gradient regions and fall back on the learned depth prior in textureless areas. Malleson \textit{et al.}~\cite{malleson2018hybrid} formulated a semantic cost function based on pair-wise and MDL (minimum description length) formulation:
\begin{equation}
    E_s = \sum_{i \in \mathcal{P}} c_{i, s^c_i} + \sum_{ij \in \mathcal{N}} V_{ij}(s^c_i, s^c_j) + \sum_{m \in \mathcal{M}} \text{MDL}_m,
\end{equation}
where $\sum_{i \in \mathcal{P}} c_{i, s^c_i}$ is data cost, which combines the mean of Euclidean distance between input point tracks and modelled point tracks with an incompleteness penalty, $\sum_{ij \in \mathcal{N}} V_{ij}(s^c_i, s^c_j)$ is the smoothness cost, which calculates a cost for each edge $ij$ in the total $\mathcal{N}$ edges in sufel graphs, and $\sum_{m \in \mathcal{M}} \text{MDL}_m$ is label cost, which adds a fixed cost for each label for each models in the set $\mathcal{M}$ with at least one point assigned to it.

On the other hand, when objects are represented by geometric primitives like quadrics, planes, or text, the cost function can be formulated to directly optimize the parameters of these objects alongside the camera poses. Hosseinzadeh \textit{et al.}~\cite{hosseinzadeh2019real}  propose an object-aware BA where the cost function minimizes the reprojection error of both 3D points $Z_j$ and points on the surface of object quadrics $Q_k$. The error for a point on a quadric is the distance between the observed 2D point $\mathbf{x}_{ik}$  and the projection of the 3D quadric:
\begin{equation}
    E = \sum_{i,j} (||\pi(\mathbf{X}_i, \mathbf{Z}_i)||^2_{\sum_{ij}}) + \sum_{i,k} (||\pi(\mathbf{X}_i, Q_k)||^2_{\sum_{ik}}),
\end{equation}
where $\pi(\mathbf{X}_i, \mathbf{Z}_i)$ is the projection of the 3D quadric $Q_k$ into the camera frame with pose $\mathbf{X}_i$. TextSLAM \cite{li2020textslam, li2023textslam} uses a cost function that combines standard point reprojection error $E_{point}$ with a photometric error computed directly on planar text features $E_{text}$, anchoring pose estimates to these stable semantic landmarks. Yang \textit{et al.}~\cite{yang2022automated} also integrates the pixel-wise corner likelihood $E^c_{d}$ of corner information and pixel-wise edge likelihood $E^\varepsilon_{d}$ of edge information into a semantic cost function.

By explicitly modeling semantic uncertainty and incorporating semantic terms into optimization cost functions, these methods create a powerful synergy between geometric estimation and high-level scene understanding, leading to significant gains in robustness, accuracy, and overall capability.

\section{Stage IV: Semantic Data Association}\label{sec:association}

Data association is the crucial, and often challenging, process of establishing correct correspondences between current sensor observations and the elements already stored in the map. It is the fundamental problem of answering the question: ``Is this object I see now the same physical object I have seen before?'' As highlighted in recent surveys, a correct association enables the system to fuse new information into an existing map landmark, while an incorrect association can lead to catastrophic failures, such as creating duplicate representations of the same object, or worse, corrupting the camera's pose estimate and the entire map structure. While traditional data association relies on geometric and appearance-based cues, semantics provide a powerful additional layer of information that can make this process significantly more robust and efficient.

Mathematically, the data association problem can be framed as finding the optimal assignment hypothesis, $\mathcal{C}^*$. The goal is to find the most likely set of pairings between measurements and landmarks:
\begin{equation}
    \mathcal{C}^* = \argmax_{\mathcal{C}} P(\mathcal{C} \mid \mathcal{Z}_t, \mathcal{M}_{t-1}).
\end{equation}

However, this formulation hides the immense combinatorial complexity of the problem. For $m$ measurements and $n$ landmarks, the number of possible association hypotheses is enormous, making a brute-force search computationally infeasible for any non-trivial scenario. This inherent complexity is compounded by several significant real-world challenges that create ambiguity, such as perceptual aliasing, viewpoint and appearance variation, sensor noise and measurement uncertainty, and dynamic environments.

\subsection{Geometric and Appearance-Based Approach}

Most modern SLAM systems use deterministic or score-based data association methods that are computationally efficient. These approaches typically combine multiple sources of information—semantic, geometric, and appearance—into a unified scoring function to determine the single best match for an observation. This multi-cue approach mitigates the risk of relying on a single, potentially unreliable source of information. The fundamental idea is to first use the semantic class as a hard constraint to drastically prune the search space. A new observation is only compared against existing landmarks in the map. After this filtering, various cues are combined to score the remaining potential matches.

Sünderhauf \textit{et al.}~\cite{sunderhauf2017meaningful} and Zhang \textit{et al.}~\cite{zhang2018semantic} match 3D points in the landmark and temporary semantic object based on nearest neighbor search and k-d tree. This method, however, faces challenges within environments with noise or complexity, where obstacles and the swift movement of objects create uncertainty. Similarly, some studies~\cite{hosseinzadeh2019real, li2019semantic} also match the projected landmarks with bounding boxes by using the Hungarian/Munkres~\cite{kuhn1955hungarian} algorithm to find the optimal solution by minimizing the cost matrix.  CubeSLAM~\cite{yang2019cubeslam} is a classic example that performs data association for both static and dynamic objects by finding the candidate object in the map that has the highest 2D bounding box IoU with the current detection. This is further refined using geometric distance checks and an optional appearance similarity score, demonstrating a simple yet effective multi-cue approach. A sophisticated approach is to explicitly combine multiple cues in a weighted score. Mu \textit{et al.}~\cite{mu2016slam} introduce the concept of a nonparametric pose graph to address the challenge of data association and SLAM with an unknown number of objects and ambiguous data association. This method uses 
Dirichlet Process prior to model data associations $D_{\mathbf{s}_i}^{\mathbf{Z}_k}$, allowing for the inference of objects $\mathbf{s}_i$ and their association with observation $\mathbf{Z}_k$:
\begin{equation}
    p (D_{\mathbf{s}_i}^{\mathbf{Z}_k}= i) = \text{DP}_i = \begin{cases}
\frac{N_{\mathbf{Z}_i}}{\sum_i N_{\mathbf{Z}_i} + \alpha} & 1 \leq i \leq N_{\mathbf{s}_i}, \\
\frac{\alpha}{\sum_i N_{\mathbf{Z}_i} + \alpha} & i = N_{\mathbf{s}_i}+1,
\end{cases}
\end{equation}
where $N_{\mathbf{Z}_i}$ is the number of observation of 
the $i$th-object in the total of $N_{\mathbf{s}_i}$ objects, and $\alpha$ is the concentration parameter of DP prior. Based on this paper, EAO-SLAM~\cite{wu2020eao} proposes an ``ensemble data association'' score, which applies the Wilcoxon Rank-sum test to point cloud data and the Single-sample T-test to object centroids. The observation is associated with the map landmark that determines the null hypothesis by these scores. However, the performance of this method is highly dependent on the careful tuning of the pre-defined threshold, which may not be optimal across different environments or object types. Chen \textit{et al.}~\cite{chen2022accurate} are also inspired by 
the ideas of~\cite{mu2016slam}, which proposes a two-stage association strategy for single keyframes and consecutive keyframes. First, a coarse association is performed using the FairMOT~\cite{zhang2021fairmot} to measure geometric consistency between the observed object and candidate map objects. In the second stage, a fine-grained semantic similarity is modeled as a Gaussian mixture model (GMM)~\cite{reynolds2015gaussian} for multiple object observation. This method emphasizes the improved accuracy and robustness in handling close and similar objects and in estimating object poses, however, this multi-stage approach can prematurely discard the correct match in the coarse filtering stage if the initial state estimates are poor. With the advent of deep learning, some systems use learned shape priors. DSP-SLAM~\cite{wang2021dsp} uses a deep shape decoder to predict a low-dimensional shape code for each object. Data association is then performed by finding the mapped object with the most similar shape code, making the association more robust to textureless objects. The framework proposed by Wu $et al.$~\cite{wu2023object} provides a comprehensive scoring function that combines 2D bounding box overlap, 3D object distance, and appearance similarity (from ORB features) to determine the best association. The object with the highest combined score is selected as the match. However, like other score-based methods, it relies on a hard assignment, which can be brittle in ambiguous situations where multiple good candidates exist.

Geometric and appearance-based approaches typically achieve efficient computational performance for semantic data association in semantic vSLAM. However, the primary weakness is that these methods make a single, ``hard'' decision for each association based on a score threshold. In ambiguous situations where multiple good candidates exist (\textit{e.g.} two identical chairs nearby), this can lead to an incorrect match. Unlike probabilistic methods that can maintain multiple hypotheses, a single incorrect hard assignment can introduce a permanent error into the system. The performance of score-based methods is highly dependent on the careful tuning of weights and thresholds. These parameters are often hand-tuned for a specific environment or sensor setup and may not generalize well to new conditions, requiring tedious recalibration. These methods are fundamentally limited by the performance of the upstream object detector. A missed detection means that no association can be made, and an incorrect class label will prevent a valid association by incorrectly pruning the correct landmark from the search space at the very first step.
\subsection{Probabilistic Approach}
To overcome the brittleness of making ``hard'' decisions about the single best match, probabilistic frameworks compute a probability for every potential association. This approach is more robust to the inherent ambiguity in real-world scenes, as it does not have to commit to a single, possibly incorrect, correspondence. These methods are particularly well-suited for integration into factor graph-based SLAM systems, where they can be represented as soft, probabilistic constraints.

A highly effective method, pioneered by Bowman \textit{et al.}~\cite{bowman2017probabilistic}, is to treat data association and landmark class labels as latent variables, utilizing the EM algorithm. The EM algorithm offers an alternative, iterative approach to probabilistic association between two steps: (1) E-Step: given the current best estimates for the states of the mapped objects, this step computes the probability, or ``responsibility'', $\gamma_ji$ that measurement $j$ originated from object $k$. This step associates observed semantic entities $\mathbf{s}_t$ with existing map entries $\mathcal{M}$ and landmark position $L_i$. The association likelihood can be written as:
\begin{equation}
    p(s^c_t \mid \mathbf{s}_t, \mathcal{M}) \propto \sum_i \delta(s^c_t = s^c_i) \cdot \mathbb{I}[\mathbf{s}_t \sim (\mathbf{l}_i, s^c_i)],
\end{equation}
where $\delta$ is the Kronecker delta and $\mathbb{I}$ is an indicator function measuring similarity \cite{bowman2017probabilistic}. (2) M-Step: using these responsibilities as weights, this step updates the state of each object (\textit{e.g.} pose, shape, velocity) by computing a weighted average over all measurements. This step is formulated as a nonlinear least squares problem by combining multiple factors, including semantic factors $f^s_{kj}(\mathbf{X}_i, L_j)$ for each landmark $j$, geometric factors $f_j^y(\mathcal{X})$, and inertial factors $f_i^{\mathcal{I}}(\mathcal{X})$:
\begin{equation}
\begin{aligned}
    \hat{\mathbf{X}}_{1:T}, \hat{L}_{1:M} = \argmin_{\mathcal{X}, L_{1:M}} &\sum_{k=1}^K \sum_{j=1}^M f^s_{kj} (\mathcal{X}, L_{1:M})  \\
    + &\sum_{i=1}^{N_y} f_i^y (\mathcal{X}) + \sum_{t=1}^{T-1} f_t^{\mathcal{I}} (\mathcal{X}).
\end{aligned}
\end{equation}

EM-Fusion~\cite{strecke2019fusion} uses this EM framework for robustly tracking multiple dynamic objects, where the responsibilities help to handle occlusions and complex interactions. Parkison \textit{et al.}~\cite{parkison2018semantic} propose a Semantic ICP algorithm, where EM is used to find the optimal alignment between two point clouds. The semantic labels are incorporated into the E-step to influence the correspondence probabilities, making the alignment process more robust to poor initializations. The joint likelihood based on residual information $\mathcal{R}$ is formulated as:
\begin{equation}
    p(\mathcal{R}, \mathcal{S}, \mathcal{D} \mid \mathcal{X}) \propto p(\mathcal{R}, \mathcal{S} \mid \mathcal{D}, \mathcal{X}) p(\mathcal{S} \mid \mathcal{D}, \mathcal{X}) p(\mathcal{D \mid \mathcal{X}}).
\end{equation}

Doherty \textit{et al.}~\cite{doherty2019multimodal} propose the use of multimodality for posterior inference over poses and landmarks, given a factor graph to accommodate non-Gaussian variables using nonparametric belief propagation, which approximates the belief over continuous state variables using Gibbs sampling and kernel density:
\begin{equation}
    p(\mathcal{X}, \mathcal{L} \mid \mathcal{X}) \propto \prod_\varphi(\mathcal{X},\mathcal{L}, \mathcal{Z}) \prod_\psi \psi(\mathcal{X}, \mathcal{L}).
\end{equation}
where $\varphi$ and $\psi$ are a measurement factor and a prior factor, respectively. Another approach by Doherty \textit{et al.}~\cite{doherty2020probabilistic} uses the ``max-marginal'' instead of the ``sum-marginal'' to mitigate the effects of data association errors while preserving the Gaussian nature of the problem. 
\begin{equation}
    p(\mathcal{X}, \mathcal{L} \mid \mathcal{X}) \propto \prod_i f_i(V_i), V_i \in \{\mathcal{X}, \mathcal{L}\}.
\end{equation}
In a factor graph, this creates a single, non-Gaussian factor for each measurement that connects to the robot pose and all potential landmark matches. The negative log-likelihood of this mixture model is minimized as the factor's cost. This makes the system extremely robust to outliers and ambiguity. If a measurement is geometrically and semantically consistent with two different landmarks, it will contribute a partial constraint to both, avoiding a risky, hard assignment. This framework has been extended to multimodal SLAM, where semantic labels from different sensors are fused within the probabilistic association.  However, this method requires evaluating each measurement against every plausible landmark in the map. In large-scale environments with thousands of landmarks, this can become a significant bottleneck. Michael \textit{et al.}~\cite{michael2022probabilistic} directly address this challenge by proposing several approximations to make probabilistic data association scalable. These include only considering a local subset of landmarks for association (\textit{e.g.}, within a radius of the robot), and using multi-resolution semantic maps to perform coarse-to-fine association. These optimizations significantly reduce the computational burden without sacrificing too much of the robustness that makes the probabilistic approach 
intensively powerful. Sünderhauf \textit{et al.}~\cite{sunderhauf2017dual, nicholson2018quadricslam} factor the conditional probability distribution as:
\begin{equation}
\begin{aligned}
    P(\mathcal{X}, \mathcal{Q} \mid \mathcal{Z}, \mathcal{L}) \propto &\prod_i P(\mathbf{X}_{i+1} \mid \mathbf{X}_i, \mathbf{Z}_i) \\ &\prod_{ijk} P(Q_j \mid\mathbf{X}_i, L_{ijk}),
\end{aligned}
\end{equation}
where $\mathcal{Q}$ and $\mathcal{L}$ are the set of quadric and line, respectively. The optimal $X^*$, $Q^*$ are sought as optimal values based on maximum a posteriori (MAP).
Zhang \textit{et al.}~\cite{zhang2019hierarchical} propose using a hierarchical topic model where an object is analogous to a document, semantic classes are analogous to topics, and visual features (\textit{e.g.}, SIFT words) are analogous to words. Data association is then performed by calculating the probability that a new observation (a ``document'' of ``words'') belongs to an existing object ``document'' based on their shared ``topic'' and ``word'' distributions. This allows the system to reason about object identity at a higher level of abstraction. 

In multi-robot SLAM, data association must also occur between the maps of different robots. Yue \textit{et al.}~\cite{yue2020hierarchical} and Deng \textit{et al.}~\cite{deng2022hd} propose hierarchical frameworks where this association is based on both geometric consistency and semantic similarity. The semantic similarity between two matched objects or voxels, with class distributions $p_i(s^c)$ and $p_j(s^c)$, can be quantified using the Kullback-Leibler (KL) divergence:
 \begin{equation}
     D_{KL} (p_i \mid \mid p_j) = \sum_c p_i(s^c) \log \frac{p_i(s^c)}{p_j(s^c)}.
 \end{equation}
A low KL divergence indicates high semantic similarity, increasing the probability of a correct association. In these collaborative mapping frameworks, a server agent can aggregate submaps from multiple robots, solve this large-scale data association problem to find the correct global alignment of all maps, and then fuse them into a unified global semantic map.

Despite their robustness, probabilistic data association frameworks are not a panacea and face several significant limitations and open challenges that are active areas of research. The performance of these frameworks is critically dependent on the quality of the upstream perception system (\textit{e.g.}, the object detector or segmentation network). If the detector fails to see an object, provides a noisy or inaccurate bounding box, or—most critically—assigns an incorrect semantic label, the probabilistic association can be severely misled. For instance, in the mixture model framework, if an object is misclassified as a table instead of a desk, its association weight $w_k$ with the correct desk landmark in the map will be zero, effectively preventing a correct association regardless of how geometrically close it is. In addition, the core strength of methods like mixture models—considering all possible associations—is also their primary weakness. The computational cost grows rapidly with the number of measurements and the number of landmarks in the map. In a large-scale, object-rich environment, evaluating every new detection against every landmark in the map becomes computationally infeasible for real-time operation. While scalability solutions like local sub-map association and multi-resolution queries have been proposed, these are approximations that trade some of the method's theoretical robustness for speed, and they may fail in scenarios with significant localization drift or large, ambiguous structures. Iterative methods like Expectation-Maximization are sensitive to initialization. A poor initial guess for object poses can lead the algorithm to converge to a local minimum, resulting in an incorrect set of associations. In environments populated with many identical or visually similar objects, such as a lecture hall with rows of identical chairs or a warehouse with uniform pallets, the data association problem remains extremely challenging. While semantics can narrow the search space to the correct class, geometric and appearance cues may be insufficient to distinguish between different instances. Finally, probabilistic methods can represent this ambiguity by assigning partial weights to multiple hypotheses, but without a truly distinguishing feature, the system may struggle to build a clean, instance-level map. When multiple objects are moving and occluding each other simultaneously, the number of plausible association hypotheses can explode, and motion models may be insufficient to accurately predict object states, leading to track-switching or loss of tracked objects.

\subsection{Advanced Data Association Contexts}

Beyond associating static objects in a scene, data association methods must handle more complex scenarios, including dynamic objects, articulated models, and long-term changes. This requires more advanced reasoning and contextual understanding. In dynamic environments, data association is synonymous with tracking—maintaining a consistent label $s^c_i$ for each object over time, regardless of its motion. DynaSLAM II~\cite{bescos2021dynaslam} is a seminal work that tightly couples the tracking of multiple dynamic objects with the camera's ego-motion estimation. Data association involves matching new object segmentation from Mask R-CNN to existing object tracks. This is guided by a motion model (\textit{e.g.}, constant velocity), which predicts the object's bounding box in the current frame. The IoU between the predicted and detected boxes is then used to establish the correspondence. These methods enhance the capabilities of accurate tracking and real-time performance, but their reliance on a simple motion model may fail for objects with abrupt or non-linear motion, leading to lost tracks. MID-Fusion~\cite{xu2019mid} formulates a cost matrix between newly observed object segments and existing dynamic object models in the map. The cost $C(\mathbf{s}_i, \mathbf{M}_k)$ for associating segment $i$ with model $k$ combines a geometric term and a semantic consistency term. While effective in robust camera tracking, continuous object information estimation, and real-world applications, the computational complexity of solving the assignment problem grows with the number of objects and detections, which can be a bottleneck in very cluttered scenes. Instead of relying on a top-down object detector, Iqbal and Gans~\cite{iqbal2018localization} use a nonparametric clustering algorithm like DBSCAN on low-level feature points to group them into spatial clusters. A data association step then matches these geometric clusters to existing object models in the map. While this method can discover novel objects, it cannot assign them a semantic label without an additional classification step. Compositional Object SLAM~\cite{sharma2021compositional} performs data association at the part level. This allows for robust tracking even when the object's overall configuration changes. However, this approach requires pre-defined part-based models for each object class, which can be difficult and time-consuming to create. PS-SLAM~\cite{li2025ps} performs data association on panoptic segments (``things''). For a detected instance, it first checks for overlap with the reprojection of previously mapped dynamic objects. If there is a significant overlap, the instance is associated with the existing track. Otherwise, it is treated as a new dynamic object instance, and a new track is initialized. Since the accuracy is highly dependent on the quality of the panoptic segmentation masks, a fragmented or incorrect mask can easily lead to a failed association.
 
In summary, advanced data association contexts offer robust tracking and efficiency in handling the complex environments. However, tracking methods for dynamic scenes often rely on simplified motion models. These models are insufficient for capturing complex, non-linear, or abrupt motions, which can lead to lost tracks or incorrect associations. In addition, the challenge of the requirement for \textit{a priori} models and the lack of semantics in unsupervised methods leads to the usefulness of the resulting map for higher-level semantic tasks unless an additional classification stage is incorporated. The effectiveness of these specialized methods can be limited to the specific contexts they were designed for. A part-based model for a chair will not help with a rigid object, and a dynamic object tracker adds unnecessary computational overhead in a static scene. Furthermore, the complexity of managing and associating data for a large number of dynamic, articulated, or collaboratively mapped objects remains a significant scalability challenge.

\section{Stage V: Loop Closure Optimization for Semantic SLAM}\label{sec:optimization}

While sequential tracking and mapping build a locally consistent view of the environment, the incremental nature of this process inevitably leads to the accumulation of small errors. Over long trajectories, this results in significant drift, where the robot's estimated pose and the map become increasingly inaccurate and inconsistent. Loop closure is the process of recognizing a previously visited place, calculating the accumulated drift, and correcting the entire map and trajectory to enforce global consistency. It is the critical step that transforms a locally accurate trajectory into a globally correct map. Traditionally, loop closure has been achieved using appearance-based place recognition methods like Bag-of-Words (BoW)~\cite{shekhar2012word}, which are susceptible to perceptual aliasing and viewpoint changes. Semantics provides a higher-level, more robust source of information for both detecting loop closures and optimizing the map afterward. This section details how semantics enhance loop closure detection (Sec.~\ref{sec:loop}) and the subsequent pose graph optimization (Sec.~\ref{sec:posegraph}).

\subsection{Semantic Loop Closing} \label{sec:loop}

Detecting that a loop has occurred is the first and most critical step in achieving global consistency. Traditional place recognition methods, which rely on matching low-level visual features, are often brittle and prone to failure in the face of significant viewpoint changes or perceptual aliasing (where different places look visually similar). Semantic information provides a higher level of abstraction, enabling the system to recognize high-level scene understanding based on its meaning and structure, which is inherently more robust.

A common and effective strategy is to combine traditional appearance-based methods with a semantic consistency check. This approach leverages the speed of appearance-based matching while using semantics as a powerful verification step to reject false positives. First, an appearance-based method generates a set of loop closure candidates. Second, each candidate is scored for semantic consistency. CALC2.0~\cite{merrill2019calc2} computes a semantic score based on the similarity score of two corresponding observations to determine a loop closure. Arshad and Kim~\cite{arshad2021semantics} propose a weighted fusion of similarity scores to improve robustness. The similarity score $R_{s}(\mathbf{Z}_i, \mathbf{Z}_j)$ for semantic loop closure can be defined as:
    \begin{equation}
R_{s}(\mathbf{Z}_i, \mathbf{Z}_j) = \sum_{k=1}^N w_k \cdot \text{R}(\mathbf{s}_{i,k}, \mathbf{s}_{j,k}),
\label{eq:semantic_loop_closure_similarity}
\end{equation}
where $w_k$ is a weighting factor, and $\text{R}(\cdot, \cdot)$ is a similarity metric like the Jaccard index and cosine similarity of semantic embeddings. Hu \textit{et al.}~\cite{hu2019loop} and SV-Loop~\cite{yuan2021sv} propose a similar fusion, where the final similarity is a weighted sum of the BoW score and a semantic score derived from the consistency of detected object labels. This strategy is also effective in specialized environments, such as a forest, where semantic cues like ``trunk'' and ``canopy'' are used to verify loop closures proposed by an appearance-based method. This hybrid approach maintains the efficiency of traditional methods while significantly improving precision by rejecting geometrically plausible but semantically inconsistent matches. It is still fundamentally reliant on the initial appearance-based candidate generation. If the BoW model fails to propose the correct loop closure candidate due to extreme viewpoint or illumination changes, the semantic check cannot recover from this failure.

To address this challenge, some studies integrate semantic information directly into the BoW model itself to create more distinctive visual words instead of using semantics as a separate verification step. A ``semantic visual vocabulary'' is created where each visual word is associated not only with its appearance but also with a semantic label~\cite{tsintotas2018seqslam}. However, this approach is limited by the limitations of the pre-trained visual vocabulary, which restricts its application in real-world environments. By clustering the description vector along with the robot's navigation part to enhance the generation of visual words, system accuracy is improved 
at the expense of computational cost. Papapetros \textit{et al.}~\cite{papapetros2023semantic} achieves this by assigning the semantic class of the region where a feature was extracted to the visual word it belongs to. Tightly integrating semantics into the vocabulary makes the BoW descriptor itself more discriminative, helping to resolve perceptual aliasing from the start. However, this approach can discard potentially useful information. By forcing features to belong to a semantic class, it may struggle with features on the border between two regions or in areas where the segmentation is ambiguous. In addition, a 3D semantic graph shifts from matching holistic scene appearance to matching the spatial arrangement of semantic objects, which is often more invariant to viewpoint changes. In this graph, nodes represent object instances and edges represent their spatial and topological relationships, in which a place is represented by a ``constellation'' of objects. A loop closure scene is detected by finding an isomorphism (a structural match) between the graph of the current scene and a previously constructed graph in the map. SemanticTopoLoop~\cite{cao2024context} builds a topological graph from object landmarks (quadrics) and calculates a similarity score between a query graph $G_q$ and a candidate graph $G_c$:
\begin{equation}
    R(G_q, G_c) = R_{c} + \rho R_s,
\end{equation}
where $R_c$ is the spatial semantic similarity and $R_s$ is the topology structure similarity based on the consistency of topological relationships. SemanticLoop~\cite{yu2023semanticloop} and the work by Lin \textit{et al.}~\cite{lin2021topology} also use similar graph-matching approaches, which calculate matching reward based on vertices similarity and edges similarity. This approach achieves an efficient performance in drift correction in drastic scene and viewpoint changes, but the accuracy and robustness can be affected by the partially reconstructed areas and the challenge of the outdoor environment. Quian~\textit{et al.}~\cite{qian2022towards} propose using a 3D semantic covisibility graph, where edges also encode the fact that two objects were seen from the same keyframe, adding another layer of topological constraint. This structural approach is highly robust to changes in viewpoint and illumination, as it relies on the fundamental structure of the scene rather than its appearance. It can also handle minor changes in the environment, like a moved chair, by finding a partial graph match. However, it may fail in sparse environments. Furthermore, as graph matching is an NP-hard (Nondeterministic Polynomial-time) problem, albeit heuristics are used, it can be computationally expensive for very large graphs. Xiao \textit{et al.}~\cite{xiao2023semantic} apply a similar two-stage approach using 360$^\circ$ panoramic images. After candidate retrieval, they project the semantic segments of the query image into the candidate image's frame. The final score is a combination of the visual similarity and the ratio of reprojection-consistent semantic pixels, effectively measuring how well the semantic layouts align. This method makes the place descriptor more viewpoint-invariant from the start, but requires specialized panoramic camera hardware and does not apply to standard monocular or stereo SLAM systems. Kim and Kim~\cite{kim2025semantic} propose a method where, after an initial set of loop candidates is generated, a semantic consistency check is performed. They compute a semantic score based on the uniqueness score of semantic object labels between the query and candidate frames. Singh \textit{et al.}~\cite{singh2021hierarchical} propose a two-level descriptor for this study. The top-level ``Semantic-Geometric Descriptor'' (SGD) is a histogram that encodes the types and relative poses of objects in the scene. The lower-level descriptor captures local geometric details. A coarse search is performed efficiently using the low-dimensional SGD, and only the top candidates are then compared using the more expensive fine-grained matching. This approach is highly efficient and scalable, as it avoids expensive geometric comparisons for the vast majority of non-matching keyframes. However, the performance is contingent on the distinctiveness of the high-level semantic constellations. In environments with many repetitive but semantically similar areas (\textit{e.g.}, a street with identical bus stops), the top-level filter may not be very effective. Chen \textit{et al.}~\cite{chen2021semantic} focus on this problem in industrial settings. Their method first performs a coarse loop closure detection using object detection to identify all objects in both the query and candidate frames. By comparing the object IDs, it explicitly identifies objects that have moved or been removed between the two views. These inconsistent instances are masked out, and the final geometric verification for the loop closure is performed only on the stable, static parts of the scene. In addition, to handle extreme viewpoint variations, PRIOR-SLAM~\cite{wang2025prior} uses a novel view synthesis approach. It generates a canonical view of the scene from the current keyframe's observation. Loop closure is then performed by matching these canonical views, which are much more consistent across different robot trajectories than the original egocentric views. The similarity score is based on a learned descriptor of this synthesized canonical image.

In summary, the integration of semantics into loop closure detection marks a significant evolution from simple appearance matching to a more robust form of scene understanding. The various strategies, from using semantic labels to verify appearance-based candidates, building hierarchical descriptors for efficient search, to matching object constellations and scene graphs for viewpoint invariance, all contribute to a common goal: improving the accuracy and robustness of semantic loop closure detection. By providing a higher level of abstraction, semantics help overcome the key limitations of traditional methods, especially in challenging real-world scenarios involving perceptual aliasing, dynamic elements, and extreme viewpoint or illumination changes. 

\subsection{Semantically-Informed Pose Graph Optimization} \label{sec:posegraph}

Once a loop closure has been detected between the current robot pose $\mathbf{x}_i$ and a previous pose $\mathbf{x}_j$, a loop closure constraint is added to the pose graph. This constraint represents the relative transformation $\Delta T_{ij}$ between two poses. The final step is to optimize the entire pose graph to find the configuration of all robot poses that best satisfy both the sequential odometry constraints and the new loop closure constraints, thereby distributing the correction across the entire trajectory. Semantic enhances this back-end optimization by providing more robust constraints and a richer multi-layer optimization problem. A typical SLAM back-end is formulated as a factor graph, a graphical model where nodes represent the robot poses and landmark positions, and factors represent constraints between them. The optimization problem to find the set of poses $\mathbf{X}^*$ and landmark positions $\mathbf{L}^*$ that minimizes a non-linear least squares cost function is represented as:
\begin{equation}
\begin{aligned}
\mathbf{X}^*, \mathbf{L}^* = \argmin_{\mathbf{X}, \mathbf{L}} &\underset{\text{odometry constraints}}{\underbrace{\sum_{(i,j) \in \mathcal{E}_G} ||e(\mathbf{X}_i, \mathbf{X}_j), \Delta T_{ij}||^2_{\Omega_{ij}}}} + \\
&\underset{\text{loop closure constraints}}{\underbrace{\sum_{(k,l) \in \mathcal{E}_L} ||e(\mathbf{X}_k, \mathbf{X}_l), \Delta T_{kl}||^2_{\Omega_{kl}}}} + \\
&\underset{\text{landmark measurement constraints}}{\underbrace{\sum_{(i,m) \in \mathcal{E}_M} ||e(\mathbf{X}_i, \mathbf{X}_m), \mathbf{z}_{im}||^2_{\Omega_{im}}}},
\end{aligned}
\label{eq:pose_graph_semantic_cost}
\end{equation}
where $e(\cdot)$ is an error function, and $\Omega$ denotes the information matrix (inverse covariance) representing the confidence in the constraint. When the objects are used as landmarks, they provide strong geometric constraints. If a loop closure is established by matching several objects, the relative transformation can be computed by finding the rigid body transform that best aligns these matched objects. Furthermore, objects with known shapes (\textit{e.g.}, from a CAD model or a learned prior) can be used to add object pose constraints to the graph. The error term for an object observation is the difference between object properties and the projection of the 3D object model into the camera frame. VSO~\cite{lianos2018vso} includes a semantic error term in its optimization that penalizes assigning a 3D map point to a semantic class that is inconsistent with the image segmentation, effectively adding a soft semantic constraint to the graph. SemanticLoop~\cite{yu2023semanticloop} uses alignment from its graph matching to generate a highly reliable pose constraint for the loop closure. Iqbal and Gans~\cite{iqbal2018localization} use nonparametric clustering to form object-level landmarks from low-level features. These object landmarks are then used as nodes in the pose graph optimization. This approach allows for the creation of object-level constraints without relying on a pre-trained object detector, making it adaptable to novel environments. VPS-SLAM~\cite{bavle2020vps} is designed for aerial robots and uses semantic segmentation to identify these planes. It adds planar constraints to the pose graph, where the error term is the point-to-plane distance between observed 3D points and the corresponding plane model in the map. This provides strong geometric stability, especially for correcting drift in height and orientation.

In addition, the richer semantic representation, like 3D scene graphs, enables a move from a single-layer pose graph to a more complex, hierarchical optimization problem.  Hydra~\cite{hughes2022hydra} builds a hierarchical system with a three-layered graph: a low-level graph of agent poses and features, a mid-level graph of object and place nodes, and a high-level graph of building structures. The optimization problem then involves minimizing errors across all these layers simultaneously. This includes not only the standard geometric errors but also inter-layer consistency errors, such as ensuring that an object node is contained within the correct room node. Agrawal \textit{et al.}~\cite{agrawal2022slam} propose a ``hybrid'' optimization that combines the continuous optimization of the pose graph with a discrete optimization over the semantic graph. After the standard pose graph optimization, a second stage optimizes the semantic labels and relationships in the scene graph to ensure that they are consistent with the newly optimized geometry. This iterative process allows geometric and semantic information to mutually refine each other. However, the complexity of the optimization problem is significantly higher. Solving multi-layered, hybrid continuous/discrete optimization problems in real-time is a major computational challenge and an active area of research.

In summary, semantically-informed pose graph optimization enriches the traditional SLAM back-end by adding a wealth of new constraints derived from a high-level understanding of the scene. By incorporating factors from objects, planes, and topological relationships, these methods can achieve a level of global accuracy and semantic consistency that is unattainable with purely geometric approaches.

\section{Deep Learning-based and Foundation Models Approaches}\label{sec:learning}
While the previous sections have discussed the integration of semantic modules into traditional SLAM architectures, recent years have witnessed a significant evolution toward learning-based approaches. These methods harness the representational power of deep learning (DL) and, more recently, foundation models (FMs) to improve or even replace key SLAM components, offering more flexible and generalizable solutions. The approaches range from supervised learning on specific tasks to open-world reasoning enabled by vision-language models.
\subsection{Deep Learning (DL)}
Deep learning techniques, particularly convolutional and recurrent neural networks, are now widely adopted as perception front-ends in SLAM systems. These models are trained on large-scale annotated datasets to predict semantics, depth, optical flow, and ego-motion. In a generalized form, learning-based SLAM can be expressed as:
\begin{equation}
    (\hat{\mathcal{X}}_t, \hat{\mathcal{M}}) = f_\theta(\mathcal{Z}_{1:t}, \mathcal{U}_{1:t}),
\end{equation}
where $f_\theta$ is a deep neural network, $\mathcal{Z}$ denotes sensor observations, and $\mathcal{U}$ control inputs~\cite{yang2020d3vo}. In monocular settings, estimating the scale of the world and handling dynamic objects are major challenges. CNN-SLAM~\cite{tateno2017cnn} was a pioneering work that integrated a CNN-based depth prediction network directly into a direct LSD-SLAM~\cite{engel2014lsd} framework. The system's energy function combines a standard photometric error with a depth consistency error weighted by the uncertainty of the CNN's prediction. This allows the system to trust the geometric photometric alignment in high-gradient regions and fall back on the learned depth prior in textureless areas. MonoRec~\cite{wimbauer2021monorec} proposes a self-supervised approach that jointly estimates camera motion, depth, and dynamic object motion, effectively disentangling ego-motion from scene motion. This demonstrates a holistic learned approach to disentangling camera motion from object motion. VLocNet++~\cite{radwan2018vlocnet++} is a deep multitask learning architecture that jointly learns semantic segmentation, global pose regression, and visual odometry. The network uses a shared encoder to extract common features and then branches into separate decoders for each specific task. The total loss is a weighted combination of the individual task losses, for example, the global pose loss $\mathcal{L}_p$ and the semantic segmentation loss $\mathcal{L}_s$:
\begin{equation}
    L_{total} = \alpha_1 \mathcal{L}_p + \alpha_2 \mathcal{L}_s.
\end{equation}
Several studies~\cite{toft2018semantic, xiao2019dynamic, wu2022learning} utilize DNNs to predict robot poses directly from semantic-rich inputs, allowing the system to implicitly learn robust semantic-geometric relationships from data. In addition, a recent trend is to move from modular pipelines to end-to-end learned frameworks that jointly estimate geometry and semantics. SimVODIS~\cite{kim2020simvodis} and SimVODIS++~\cite{kim2022simvodis++} propose a single neural architecture that simultaneously performs visual odometry, object detection, and instance segmentation. This multi-task learning allows the network to leverage shared representations. The pose and depth branches are trained in a self-supervised manner using photometric consistency, while the semantic branches are supervised. SimVODIS++ further introduces an attentive pose estimation module that learns to focus on salient and likely static regions, effectively ignoring dynamic objects without explicit masking. BEV-Locator~\cite{zhang2025bev} proposes an end-to-end network that transforms multi-view images into a Bird's-Eye-View (BEV) representation. This visual BEV feature map is then directly matched against a pre-existing semantic BEV map to regress the camera pose. Although promising, these systems are often limited to the training distribution.

In summary, these methods are fundamentally limited by the supervised learning paradigm. Their performance is constrained by the scale and diversity of the annotated training data. A model trained on an indoor dataset may perform poorly in an outdoor driving scenario. This dependency on labeled data is a significant bottleneck for creating truly general-purpose SLAM systems and motivates the recent shift towards foundation models, which learn from much broader, web-scale data.

\subsection{Foundation Model}

The most recent and disruptive trend in AI is the rise of large-scale, pre-trained Foundation Models, particularly Vision-Language Models (VLMs) like CLIP, which are trained on web-scale datasets of image-text pairs. Their emergence is a paradigm shift for Semantic SLAM, moving the field from a ``closed-world'' assumption (with a small, predefined set of object classes) to an ``open-world'' or ``open-vocabulary'' setting.  Instead of relying on a detector trained on a fixed set of classes, a VLM can be prompted with arbitrary natural language text to detect virtually any object. This is achieved by comparing the embedding of an image region with the text embedding of the query. The similarity $R$ between an image patch $I$ and a text query $T$ is typically calculated as the cosine similarity of their respective embeddings, $E_I$ and $E_T$, produced by the foundation model:
\begin{equation}
    R(I,T) = \frac{E_I \cdot E_T}{||E_I|| \cdot ||E_T|| + \epsilon}.
\end{equation}

The primary impact of foundation models is the ability to build dense, queryable 3D maps that are not limited to a predefined set of classes. ConceptFusion~\cite{jatavallabhula2023conceptfusion} was a pioneering work that demonstrated how to fuse open-set CLIP features from multiple views into a globally consistent 3D map, creating a rich, multi-modal feature field that can be queried with text or images. FindAnything\cite{laina2025findanything} and LOSS-SLAM~\cite{singh2024loss} build on this by creating lightweight, open-vocabulary, and object-centric mapping frameworks. They use a Segment Anything Model (SAM) to generate class-agnostic segments and then fuse the associated VLM features of these segments into volumetric submaps, creating maps that are both dense and queryable. These methods typically fuse high-dimensional pixel-aligned features with 3D point clouds or meshes using volumetric fusion or learned neural fields:
\begin{equation}
\mathbf{f}p = \frac{1}{N} \sum_{i=1}^N \mathcal{P}(g_\phi(I_i))[u_i],
\end{equation}
where $\mathcal{P}$ projects image features to the 3D map, $g_\phi$ is the feature encoder (\textit{e.g.}, CLIP), and $u_i$ is the pixel coordinate corresponding to 3D point 
$p$ in image $I_i$. However,  CLIP and SAM require significant GPU memory and latency, limiting onboard deployment. This approach has also been applied to real-time panoptic reconstruction, as shown by PanoRecon~\cite{wu2024panorecon}, which demonstrates the fusion of such open-set semantic features into a panoptic 3D map.

Foundation models enable the construction of more abstract and structured representations like 3D Scene Graphs with open-ended semantic nodes. While earlier works like Wald \textit{et al.}~\cite{wald2020learning} and SceneGraphFusion~\cite{wu2021scenegraphfusion} laid the groundwork for incrementally building 3D scene graphs from RGB-D data with a fixed set of classes, foundation models have enabled a shift to open-vocabulary graphs. ConceptGraphs~\cite{gu2024conceptgraphs} builds a 3D scene graph where object nodes are associated not with a fixed class label, but with open-vocabulary ``concepts'' derived from VLM features. QueSTMaps~\cite{mehan2024questmaps} focuses on creating queryable semantic topological maps where each node in the topological map is associated with a local scene graph. While these methods achieve efficient performance in 3D representation, high-dimensional features lack clear spatial anchoring, complicating drift correction and map updates. This hierarchical structure allows for efficient, language-based queries about both the high-level structure of the environment and the detailed object relationships within a specific place. Clio~\cite{maggio2024clio} proposes a task-driven approach, where the robot uses a VLM to decide what semantic information is relevant to include in its scene graph based on a list of tasks provided in natural language, addressing the key question of what level of semantic granularity is actually needed. In addition, Hier-SLAM\cite{li2025hier} and Hier-SLAM++\cite{li2025hier++} use an LLM to generate a hierarchical tree of semantic concepts (\textit{e.g.}, chair $\rightarrow$ seating furniture $\rightarrow$ furnishings) to support coarse-to-fine reasoning and memory-efficient learning. This symbolic knowledge graph is then used to produce compact, hierarchical semantic embeddings for each 3D Gaussian primitive. A novel semantic loss function is used during optimization to ensure that the learned semantic features are consistent both within and across the levels of this hierarchy, creating a highly structured and efficient semantic map. The semantic class can be expressed hierarchically as:
\begin{equation}
s^c_i = \{ v_i^l, e_i^m | \quad l = 0, \dots, L; m = 0, \dots, L-1 \},
\end{equation}
where $v_i^l$ is the node at level $l$ and $e_i^m$ represents hierarchical semantic links. Li \textit{et al.}~\cite{li2024resolving} used VLMs to resolve the perceptual aliasing that plagues loop closure in repetitive environments. By querying the model with images of two visually similar locations, the VLM can provide fine-grained textual descriptions that highlight subtle differences (\textit{e.g.}, ``the desk with a blue pen''), allowing the system to correctly reject a false loop closure. However, assigning per-point features leads to large memory footprints, limiting scalability to large scenes. SENT-Map\cite{kathirvel2025sent} exploits LLMs to create topological or task-driven maps stored in structured formats and used for symbolic task planning. These maps encode not only geometry and semantics but also ownership, affordances, and scene-level context. LEXIS~\cite{kassab2024language}  leverages VLMs to allow for loop closures to be initiated via language queries. For example, a user could command the robot to ``return to the room with the red chair''. The system then uses a Vision-Language Model (VLM) to ground this query, identifying past keyframes that match the description and using them as high-confidence loop closure candidates. However, it is not a fully autonomous loop closure system and relies on a human operator to provide the language-based cues for initiating the search.

In summary, foundation models are revolutionizing the perception and reasoning capabilities of SLAM systems, pushing them from simple geometric mappers with a fixed set of classes towards genuine open-world, language-grounded scene understanding systems. While open-set in capability, many models still fail in ambiguous, repetitive, or occluded environments without further fine-tuning. To bridge language and geometry effectively, future directions include neuro-symbolic fusion, scene graphs, and compact embeddings. Combining CLIP features with learned scene embeddings~\cite{wu2021scenegraphfusion} or using memory-efficient retrieval architectures~\cite{liu2024fm} offers promising directions.

\section{Continuous Representation}
\label{sec:continuous}
Traditional SLAM systems have relied on discrete map representations like point clouds, voxel grids, or meshes. While effective, these representations can be memory-intensive and may struggle to represent complex geometry and appearance in a continuous and photorealistic manner. As detailed in the recent survey by Tosi \textit{et al.}~\cite{tosi2024nerfs}, a paradigm shift has occurred towards continuous representations, where the scene is modeled not as a discrete set of elements, but as a continuous function. These learned functions, often parameterized by neural networks or other primitives, can store geometry and appearance at arbitrary resolutions, enabling high-fidelity rendering and novel view synthesis. This section reviews the main categories of continuous representations emerging in the SLAM field: neural implicit representation, 3D Gaussian splatting, and hybrid approaches.

\subsection{Neural Implicit Representation}

Neural implicit representations model a scene as a continuous function $f$ that maps a 3D coordinate (and potentially a viewing direction) to a physical property, such as occupancy, signed distance, or color and density based on  Neural Radiance Fields (NeRFs)~\cite{mildenhall2021nerf}. A NeRF learns a mapping $f_\theta: \mathbb{R}^3 \times \mathbb{R}^d \rightarrow \mathbb{R}^C \times \mathbb{R}$, with $\mathbf{d}$ the viewing direction, and $\sigma$ the volume density:
\begin{equation}
    (\mathbf{X}, \mathbf{d}) \rightarrow  (\sigma, \mathbf{s}^c).
\end{equation}
This mapping is parameterized by the weight $\theta$ of a Multi-Layer Perceptron (MLP). To render a pixel, volumetric rendering is used to integrate the semantic color and density predictions along a camera ray $\mathbf{r}(t) = \mathbf{o} + t\mathbf{d}$ passing through that pixel:
\begin{equation}
\begin{aligned}
    C(\mathbf{r}) &= \int_{t_n}^{t_f} T(t)\sigma(\mathbf{r}(t))\mathbf{s}^c(\mathbf{r}(t), \mathbf{d})dt, \\
             T(t) &= \exp \Big ( -\int_{t_n}^{t_f} \sigma(\mathbf{r}(s))ds  \Big ).
\end{aligned}
\end{equation}
By optimizing the network weights $\theta$ to minimize the photometric error between the rendered and ground truth images, the MLP learns a complete, continuous representation of the scene. An earlier, related concept was presented by Czarnowski \textit{et al.}~\cite{czarnowski2020deepfactors}, which introduced the idea of using a learned, compact code vector to represent the geometry of a local keyframe. This ``deep factor'' was integrated into a factor graph for probabilistic dense monocular SLAM, showcasing the potential of learned compact codes. However, this method did not produce a continuous map of the entire scene, instead representing geometry on a pre-keyframe basis. iMAP~\cite{sucar2021imap} was a landmark achievement, demonstrating the first real-time SLAM system using a pure implicit representation. Instead of a large MLP, it uses a single, small MLP and a set of active keyframes to represent the scene. The use of low-capacity MLP meant that it could only map very small-scale scenes and was prone to catastrophic forgetting. Several works~\cite{kong2023vmap, han2023ro} also integrate NeRF-based objects with dedicated MLP models into a lightweight SLAM framework, achieving real-time object-level mapping from monocular RGB input. In addition, to overcome the scalability and efficiency limitations of the initial real-time methods, MeSLAM~\cite{kruzhkov2022meslam} addresses them by using multiple lightweight MLPs, each small MLP representing a local region of the scene, allowing the total memory footprint to scale more gracefully with the environment size. However, managing the consistency and overlap between multiple MLPs can be complex. Co-SLAM~\cite{wang2023co} introduced a hybrid implicit representation that combines a sparse set of feature grids with a coordinate-based MLP. ESLAM~\cite{johari2023eslam} and NISB-Map~\cite{xiang2023nisb} propose submap-based approaches to enable large-scale mapping, which divide the environment into a series of local submaps and represent them by their own independent neural implicit model. While it enables theoretically unbounded scalability, as the memory and computational load are dependent on the size of a submap, stitching submaps together seamlessly and handling loop closures between them is a significant challenge. Point-SLAM~\cite{sandstrom2023point} proposes anchoring the neural radiance field to an explicit point cloud to regularize the neural field, leading to sharper reconstructions. NeRF-SLAM~\cite{rosinol2023nerf} was one of the first works to tackle this for dense monocular SLAM, but it was limited to small workspaces and not real-time. To address the speed issue, NICE-SLAM~\cite{zhu2022nice}, IMODE~\cite{matsuki2023imode}, and NICER-SLAM~\cite{zhu2024nicer} extend NeRF-based SLAM with hierarchical neural fields, supporting efficient optimization and scene decomposition. SNI-SLAM~\cite{zhu2024sni}  employs a neural implicit representation, hierarchical semantic encoding for multilevel scene comprehension, and cross-attention mechanisms for collaborative integration of appearance, geometry, and semantic features. Loopy-SLAM~\cite{liso2024loopy} addressed a major limitation of the lack of loop closure by introducing a pose graph optimization back-end to an implicit SLAM system. When a loop is detected, a pose graph is optimized, and the resulting correction is used to deform the neural implicit map, ensuring global consistency. DynaMoN~\cite{schischka2024dynamon} incorporates semantic segmentation to differentiate static and dynamic regions, allowing for a motion-aware localization and reconstruction process. By focusing the NeRF training on the static background while accounting for dynamic elements, it achieves robust camera tracking and high-quality novel-view synthesis even in the presence of motion.

\subsection{3D Gaussian Splatting}
In contrast to implicit methods that encode the entire scene into the weights of a coordinate-based neural network, another powerful approach is to use an explicit continuous representation. In this paradigm, the scene is still modeled as a continuous function, but it is composed of a discrete set of primitives, each with its own explicit and optimizable properties. This avoids the need for expensive, repetitive neural network queries during rendering, which is the primary bottleneck for pure implicit methods. The state-of-the-art explicit representation that has revolutionized the field is 3D Gaussian Splatting. A scene is represented as a collection of 3D Gaussians, where each Gaussian $G$ is a continuous volumetric function defined by its mean $\mu$, covariance matrix $\bm \sum$, semantic color $\mathbf{s}^c$, and opacity $\alpha$:
\begin{equation}
    G(X) = \alpha \cdot \exp \Big ( -\frac{1}{2} (\mathbf{X} - \mu)^T \bm \sum^{-1} (\mathbf{X}-\mu)\Big).
\end{equation}
Rendering is achieved by ``splatting'' these 3D Gaussians onto the 2D image plane and blending them in depth order. Because this process is fully differentiable and relies on fast GPU rasterization, the parameters of all Gaussians can be optimized directly by minimizing the error between the rendered and ground truth images at very high speeds. The first wave of Gaussian Splatting SLAM systems demonstrated its viability for real-time, dense mapping. SplaTam~\cite{keetha2024splatam} was a pioneering work that used RGB-D data to incrementally build a map of 3D Gaussians while simultaneously tracking the camera pose against the map. It established a core framework of tracking by rendering the map and minimizing photometric and geometric errors, followed by a mapping step that adds and optimizes Gaussians from new keyframes. Similarly, GS-SLAM~\cite{yan2024gs} and Gaussian Splatting SLAM~\cite{matsuki2024gaussian} demonstrated comparable capabilities, solidifying Gaussian Splatting as a leading representation for dense visual SLAM. However, the number of Gaussians required to represent a scene grows linearly with the size and complexity of the environment, making it difficult to scale to very large areas without submapping strategies, which introduce their own complexities like inter-map loop closures. SGS-SLAM~\cite{li2024sgs}, SemGauss-SLAM~\cite{zhu2025semgauss}, NEDS-SLAM~\cite{ji2024neds}, and GS$^3$LAM~\cite{li2024gs3lam} all augment each 3D Gaussian with a learned semantic feature vector. This allows the system to render not only color and depth but also dense, pixel-aligned semantic maps from novel viewpoints. The optimization objective in these systems is typically extended to include a semantic loss (\textit{e.g.}, cross-entropy) between the rendered semantic labels and the ground truth segmentation from a 2D perception network. Hier-SLAM~\cite{li2025hier} extends this concept to Semantic SLAM. It augments each 3D Gaussian with a learned semantic feature vector. This allows the system to render not only color and depth but also semantic maps. Hier-SLAM++~\cite{li2025hier++} further uses an LLM to create a hierarchical semantic structure, which is then encoded into the Gaussians, enabling a highly efficient and structured semantic map. SDD-SLAM~\cite{liu2025sdd} assigns semantic labels to distinguish static background Gaussians from dynamic object Gaussians, allowing for independent tracking and rendering. To address scalability, Xin \textit{et al.}~\cite{xin2025large} propose a submap-based approach, where the global map is composed of multiple local Gaussian Splatting submaps that are optimized and stitched together, addressing the memory challenges of a single monolithic representation. Yugay \textit{et al.}~\cite{yugay2025gaussian} tackles the problem of long-term dynamics by introducing a keyframe management system that can discard outdated observations, allowing the Gaussian map to adapt to changes in the environment over time.

\subsection{Hybrid Representation}
Hybrid representation aims to combine the advantages of both implicit and explicit models to achieve both high quality and real-time performance. It typically uses an explicit, sparse data structure (like a feature grid, an octree, or a point cloud) to partition the scene and provide a geometric scaffold. This structure is then coupled with a lightweight, implicit neural network that learns to represent fine details and appearance. NICE-SLAM~\cite{zhu2022nice} was a foundation work that introduced a hierarchical, multi-resolution feature grid, enabling high-quality surface reconstruction in real-time. It demonstrated that this hybrid approach could overcome the speed limitation of pure NeRF-based SLAM. Subsequent works like Co-SLAM~\cite{wang2023co}, H$^2$-Mapping~\cite{jiang2023h}, and H$^3$-Mapping~\cite{jiang2024h3} further refined this by using explicit geometric priors to guide the training of the implicit feature grids, which accelerates convergence and improves reconstruction quality. Point-SLAM~\cite{sandstrom2023point} anchors the neural radiance field to an explicit point cloud generated by a standard SLAM front-end. This provides a strong geometric prior that regularizes the implicit field. Similarly, Neural Surfel Reconstruction~\cite{cui2024neural} uses explicit surfels to represent the scene's geometry, where each surfel is associated with a learned neural descriptor that captures local appearance. A higher-level hybrid approach is taken by Neural Topological SLAM~\cite{chaplot2020neural}, which combines an explicit topological graph of the environment with local implicit models at each node capable of reconstructing the appearance of a specific place. While hybrid methods are powerful, they represent a compromise. The quality of the final reconstruction is often constrained by the resolution of the underlying explicit grid structure. While they are much faster than pure implicit methods, they can be more complex to implement and may not achieve the same rendering speed as pure explicit methods like Gaussian Splatting.

In summary, the move towards continuous representations marks a significant evolution in SLAM, enabling levels of photorealism and detail previously unattainable with discrete maps. A clear trade-off exists between the different approaches. Implicit representations like NeRFs offer state-of-the-art qualities for novel view synthesis but have historically been too slow for real-time SLAM. Explicit representations, dominated by 3D Gaussian Splatting, offer a compelling alternative with extremely fast rendering speeds, making them highly suitable for real-time SLAM applications, though they can be memory-intensive. Hybrid methods have emerged as a powerful middle ground, combining the strengths of both to achieve real-time performance with high-quality reconstruction. The current trajectory of the field shows immense momentum behind explicit Gaussian Splatting and sophisticated hybrid models as the most promising avenues for building the next generation of dense, photorealistic, and semantically-aware SLAM systems.

\section{Multi-robot Semantic SLAM}\label{multirobot}
Deploying a team of robots to map an environment offers significant advantages over a single-robot system, including increased speed, efficiency, and robustness to individual robot failures. Multi-robot SLAM, however, introduces a new layer of complexity: in addition to solving the standard SLAM problem for itself, each robot must also be able to share information, understand the perspective of its teammates, and fuse its local map into a globally consistent world model. When semantics are involved, these challenges extend to the semantic domain, requiring robots to build a shared, coherent understanding of the environment's meaning. This section reviews the key challenges and approaches in multi-robot Semantic SLAM, covering system architectures, map fusion, and collaborative active mapping.

\subsection{System Architectures: Centralized vs. Decentralized}

Multi-robot SLAM can be formulated as a joint optimization problem. Given a team of $N$ robots, the goal is to estimate the set of all trajectories $\mathcal{X}^N = \{\mathcal{X}, \cdots, \mathcal{X}_N \}$ and the global semantic map $\mathcal{M}$ by maximizing the posterior probability given all inter- and intra-robot measurements $\mathcal{Z}$:
\begin{equation}
    (\mathcal{\hat{X}}^N, \hat{\mathcal{M}}) = \argmax_{\mathcal{X}^N, \mathcal{M}} P(\mathcal{X}^N, \mathcal{M} \mid \mathcal{Z}).
\end{equation}

The architecture of the system—centralized or decentralized—determines how this joint problem is solved. In a centralized architecture, all robots send their raw or partially processed data (\textit{e.g.}, keyframes, local submaps, detected objects) to a single, powerful base station or server. This central server is responsible for performing the most computationally intensive tasks: jointly optimizing all robot trajectories, detecting inter-robot loop closures, and fusing all data into a single, global map.
Systems like HD-CCSOM~\cite{deng2022hd} and the framework by Yue \textit{et al.}~\cite{yue2020hierarchical} use this approach. A central server receives semantic occupancy submaps or object lists from multiple robots. It then solves the large-scale data association problem to find the correct relative poses between all the submaps and fuses them into a unified global representation. Hammer~\cite{yu2025hammer} demonstrates this for modern representations, where multiple robots, even heterogeneous ones, send their individually constructed Gaussian Splatting maps to a central server for alignment and fusion. However, this approach requires high bandwidth and reliable communication with the central server. It has a single point of failure; if the server goes down, the entire collaborative system fails. In a decentralized architecture, there is no central server. Robots communicate directly with each other (peer-to-peer) to share information and build a consistent map. Each robot is responsible for maintaining its own estimate of the global map. Kimera-Multi~\cite{tian2022kimera} is a state-of-the-art decentralized system that builds a distributed, dense metric-semantic mesh map. When two robots are within communication range, they exchange their local trajectory and map data, perform a relative pose estimation to ``dock'' their maps, and then communicate updates to maintain consistency. SlideSLAM~\cite{liu2024slideslam} and the work by Tchuiev and Indelman~\cite{tchuiev2020distributed} also propose robust decentralized frameworks that allow for consistent, distributed semantic mapping and localization without a central server.

\subsection{Map Fusion and Inter-Robot Data Association}
The core technical challenge in multi-robot SLAM is fusing the maps from different robots. This requires finding the rigid body transformation $_WT_{R_j \rightarrow R_i}$ that aligns the coordinate frame of robot $j$ with that of robot $i$ in the world frame $W$. The most common method is to perform place recognition between robots. When robot $i$ observes a place that robot $j$ has seen before, an inter-robot loop closure constraint is created. The optimization problem is to find the transformation that best aligns the matched landmarks from the two maps. Given a set of matched landmark pairs $(l_{i,k}, l_{j,k})$, the transformation is found by minimizing a geometric error:
\begin{equation}
    _WT_{R_j \rightarrow R_i}^* = \argmin_T{\sum_K ||m_{i,k} = T \cdot m_{j,k}||^2}.
\end{equation}
Semantics play a crucial role in making this inter-robot association more robust. The association can be based on matching object constellations or semantic graph structures between the two maps. The hierarchical framework by Yue \textit{et al.}~\cite{yue2020hierarchical} determines inter-robot object association using both geometric consistency and semantic similarity, measured by the KL divergence between the objects' semantic label distributions. A low KL divergence indicates high semantic similarity, increasing the confidence in the match. CoSAR~\cite{hu2023cosar} explicitly addresses the limitation of network bandwidth in realistic scenarios by designing a system that can adapt to varying communication quality. It prioritizes the transmission of compact semantic information (like object detections) over dense geometric data when the bandwidth is low, ensuring that the most critical information for alignment can still be shared.

\subsection{Collaborative Active Mapping and Exploration}
Given a team of robots, the goal of active mapping is to coordinate their movements to explore and map an unknown environment as efficiently as possible. The problem is often framed as a decentralized decision-making process where each robot $k$ chooses its next action $a_k$ to contribute to a global utility function $f_U$. A common formulation is to maximize the information gain about the map:
\begin{equation}
    \mathcal{A}^* = \argmax_{\mathcal{A}} f_U(\mathcal{A}) = I(\mathcal{M, \mathcal{Z}} \mid \mathcal{A}),
\end{equation}
where $\mathcal{A} = \{a_1, \cdots, a_N \}$ is the set of joint actions for all robots, and $I(\cdot)$ is the mutual information between the map $\mathcal{M}$ and future expected measurements $\mathcal{Z}$ given those actions. The problem is often framed as finding a set of next-best-view waypoints for the team that maximizes a global utility function. This utility function can be designed to encourage rapid exploration (\textit{e.g.}, maximizing frontier coverage) or to maximize the quality of the semantic map. Liu \textit{et al.}~\cite{liu2022active} define the utility as the expected reduction in the uncertainty of the metric-semantic map. This directs robots towards regions where the semantic classification of objects is still uncertain, thereby improving the quality of the final semantic map. Asgharivaskasi \textit{et al.}~\cite{asgharivaskasi_riemannian_2025} use Riemannian optimization to plan paths for robot teams that maximize map quality. Aguilar \textit{et al.}~\cite{aguilar2025multi} demonstrate a system with a heterogeneous team of ground robots and a drone, where the drone is used to provide a high-level overview to guide the exploration of the ground robots in building a detailed semantic map.

In summary, multi-robot Semantic SLAM extends the single-robot problem with significant challenges in communication, data fusion, and coordination. The choice between centralized and decentralized architectures presents a fundamental trade-off between optimality and robustness. Semantics provides a crucial common language for robots to understand each other's maps and to coordinate their actions, enabling teams of robots to build rich, globally consistent world models more efficiently than any single robot could alone.

\section{Open Research Questions}\label{sec:question} 
Despite the significant progress in Semantic SLAM over the past decade, several fundamental challenges remain. The transition from controlled laboratory settings to complex, dynamic, and long-term real-world deployments requires further innovation. This section highlights key open research questions and promising future directions that will shape the next generation of Semantic SLAM systems, drawing upon the insights from several recent surveys and forward-looking research papers.

\subsection{Semantic Benchmarks and Evaluation Methods}   

The progress of any data-driven field is intrinsically linked to the quality of its benchmarks and the rigor of its evaluation metrics. For Semantic SLAM, evaluation must go beyond geometric accuracy to assess the quality of the semantic understanding itself. This requires both suitable datasets with rich ground truth and standardized metrics that can capture semantic correctness, as consistently highlighted in major surveys of the field.

Evaluating semantic mapping requires datasets that provide not only sensor data and ground truth trajectories but also dense, per-pixel or per-point semantic and instance-level annotations.  For indoor scenes, datasets like ScanNet~\cite{dai2017scannet}, Matterport3D~\cite{Matterport3D}, and the Active Vision Dataset~\cite{Phil2018active} have become standards, offering dense semantic and instance-level labels, with some providing higher-level annotations for rooms and object relationships, crucial for evaluating place categorization as demonstrated by Sünderhauf \textit{et al.}
~\cite{sunderhauf2016place}. For outdoor and autonomous driving contexts, KITTI Vision~\cite{geiger2012cvpr, Menze2015CVPR}, KITTI-360~\cite{Liao2022PAMI}, and nuScenes~\cite{caesar2020nuscenes} are the primary benchmarks, providing dense per-point labels, 3D object bounding boxes, and 3D semantic segmentation in complex urban environments. Most existing benchmarks, while useful, were not designed for the primary purpose of evaluating semantic, long-term, or dynamic mapping. The majority of popular indoor datasets
are captured in static environments over very short trajectories. This makes them unsuitable for developing or fairly evaluating systems designed to handle dynamic objects or environmental changes. There is a critical need for large-scale, long-term datasets that capture the same environment over weeks, months, or seasons. These datasets would be essential for developing and evaluating lifelong SLAM systems capable of handling appearance changes, evolving object layouts, and seasonal variations. While datasets like ScanNet provide instance-level semantic labels, the annotations are typically ``flat''. They lack the hierarchical structure, functional affordances, and inter-object relationships that are necessary to train and evaluate advanced scene understanding systems that build representations like 3D scene graphs. 

While the geometric accuracy of a system is still evaluated using standard metrics like Absolute Trajectory Error (ATE) and Relative Pose Error (RPE), the semantic quality of the map is assessed through a different set of criteria. For dense semantic maps, the most common metric is the mean Intersection-over-Union (mIoU), calculated as the average IoU across all classes. For panoptic systems that distinguish object instances, the Panoptic Quality (PQ) is used, which combines Segmentation Quality (SQ) and Recognition Quality (RQ) in the formula $PQ = SQ \times RQ$. Beyond pixel-level accuracy, a crucial aspect is evaluating the system's utility for downstream tasks. This is often done by measuring performance on high-level tasks such as place categorization, using standard classification metrics like precision, recall, and F1-score. For the critical sub-task of loop closure detection, a precision-recall curve is the most informative metric, as it captures the essential trade-off between correctly identifying re-observed locations and rejecting false positives, a key focus in the work by Garg \textit{et al.}~\cite{garg2017improving}. However, a significant open challenge is the lack of quantitative semantic evaluation for the most abstract and open-ended representations. The evaluation of the most abstract representations remains an open challenge, often relying on computationally expensive metrics like graph edit distance. For systems producing complex 3D scene graphs or leveraging open-vocabulary foundation models, objective metrics are difficult to define. Metrics like graph edit distance are computationally expensive and not widely adopted. As a result, most recent studies in these advanced areas focus on qualitative evaluation, showcasing compelling examples of reconstructed scenes, correct query responses, or visually plausible maps, rather than reporting standardized, quantitative scores. This gap highlights a critical need for the community to develop new, rigorous metrics to benchmark the true semantic and reasoning capabilities of these state-of-the-art systems.

\subsection{From Semantic SLAM to Active Semantic SLAM}

Most current SLAM systems are passive; they build a map based on a pre-defined or human-controlled trajectory. This is analogous to a passenger in a car who builds a mental map of a city but has no control over the route. The next frontier, and a significant open research challenge, is Active SLAM, where the robot intelligently and autonomously decides where to move next in order to build the most useful map as efficiently as possible. This requires a tight coupling between perception, mapping, and decision-making, transforming the robot from a passive observer into an active agent. This problem is typically formulated as a Partially Observable Markov Decision Process (POMDP), where the robot chooses an action $a^{*}$ from a set of possible actions $\mathcal{A}$ that maximizes a long-term utility function $U(\cdot)$ representing the expected information gain about the map. While traditional active SLAM focuses on geometric exploration by directing the robot towards frontiers between known free space and unknown territory, Active Semantic SLAM reframes this objective in semantic terms, making the process significantly more intelligent.

In this paradigm, the utility function is designed to maximize semantic understanding. For instance, the core objective can be to reduce the uncertainty of object classifications in the map, a principle explored in the work of Liu \textit{et al.}~\cite{liu2023active} and Tao \textit{et al.}~\cite{tao20243d}. The objective can also be driven by a high-level task, such as finding a specific object category (\textit{e.g.}, ``find a chair''), where the system learns to map the environment in a way that is explicitly useful for achieving this semantic goal, as demonstrated by Georgakis \textit{et al.}~\cite{georgakis2021learning}. To develop these intelligent exploration policies, recent approaches have turned to advanced decision-making techniques. Zhang \textit{et al.}~\cite{zhang2024active} propose a method based on spectral analysis of the semantic pose graph to identify the most informative exploration candidates, while Tian \textit{et al.}~\cite{tian2024rasls}  use deep reinforcement learning to train an agent that can leverage high-level layout semantics to navigate more efficiently than methods based on geometric frontiers alone. Ravichandran \textit{et al.}~\cite{ravichandran2022hierarchical} demonstrate how a robot can learn an effective navigation policy by using a Graph Neural Network (GNN) that operates directly on a 3D scene graph.  A foundational concept in this area, explored in the thesis by Baxter~\cite{baxter2020toward}, is the idea of goal-oriented exploration, where an agent learns to build a map specifically to help it find a target object, demonstrating the tight coupling between perception and action. Despite its promise, active Semantic SLAM remains a challenging research problem. The computational cost of evaluating the potential information gain from many possible future viewpoints can be prohibitive for real-time operation. Furthermore, designing utility functions that perfectly capture complex, task-driven semantic goals is difficult and an area of ongoing research. Finally, long-horizon planning in large, unknown environments remains a significant challenge for any autonomous system.

\subsection{Lifelong Semantic vSLAM}
Real-world environments are not static; they are in a constant state of flux. Furniture is rearranged, seasons alter the appearance of outdoor scenes, new objects are introduced, and old ones are removed. A truly autonomous robot must be able to operate for weeks, months, or years within such evolving spaces, continuously updating its map to reflect these changes. This is the grand challenge of Lifelong SLAM. It requires a fundamental shift from the traditional goal of building a single, static, and complete map to the much more difficult task of maintaining a dynamic, ever-evolving world model that can learn and adapt over time without human intervention. This long-term adaptation is intrinsically a continual learning problem. The robot is constantly presented with a non-stationary stream of data from its environment, and it must learn from this new information without catastrophically forgetting the stable, foundational knowledge it has learned in the past. As framed by Vödisch \textit{et al.}~\cite{vodisch2022continual}, a lifelong system must achieve a delicate balance between plasticity (the ability to rapidly learn new information and changes) and stability (the ability to retain and preserve old, still-valid knowledge). Li \textit{et al.}~\cite{li2024learn} formalize this as the need for the system to both ``learn to memorize and to forget''. The system must robustly memorize the permanent or semi-permanent structure of the environment while actively identifying and forgetting transient elements (like a delivery box that is only present for a day) or outdated information (like the position of a desk that has been moved) to prevent map corruption and unbounded growth. Early works like CoVIO~\cite{vodisch2023covio} are beginning to develop online continual learning frameworks for core SLAM components like visual-inertial odometry, which is a foundational step towards building fully adaptive systems.

A key technical innovation for enabling lifelong operation is the move towards spatio-temporal mapping, where the map explicitly represents the dimension of time. A standard 3D map is atemporal; it cannot distinguish between a chair that was present yesterday and one that is present today. Systems like Khronos~\cite{Schmid2024Khronos} pioneer a solution by proposing a unified spatio-temporal Metric-Semantic SLAM system. It builds a 4D (3D space + time) graph representation of the world. In this graph, object observations are associated over time to create a ``world-tube'' that represents the object's lifecycle—when it appeared, if it moved, and when it disappeared. This allows the system to explicitly reason about object permanence and dynamics, a crucial capability for understanding a changing world. Since manually annotating data for every new environment a robot might encounter over its lifetime is impossible, lifelong systems must also adapt in a self-supervised manner. BYE (Build Your Encoder) by Huang \textit{et al.}~\cite{huang2025bye} presents a novel framework for this, allowing a robot to build its own customized perception encoder from a single unlabeled exploration sequence in a new environment. By leveraging temporal consistency and multiview geometry as a powerful self-supervisory signal, the robot can fine-tune its perception model to the specific objects and appearance of its current surroundings, enabling robust long-term understanding without any human annotation. Ultimately, the purpose of a lifelong map is to serve as a persistent, queryable knowledge base—a long-term memory for the robot. This map should be able to answer questions that involve both space and time, such as ``Where did I last see my keys yesterday?''. Systems like QueSTMaps~\cite{mehan2024questmaps}, which focus on creating queryable semantic topological maps, represent a significant step in this direction. A true lifelong system would extend this queryable memory with a temporal dimension, allowing the robot to reason about the history of its environment. Despite this progress, Lifelong SLAM remains one of the most significant open challenges, primarily due to the immense computational and memory costs of storing and reasoning about a 4D world model and the difficulty of developing robust methods for change detection, forgetting, and true continual learning without failure.

\subsection{Generalization \& Robustness}

A major open challenge is achieving true generalization and robustness, two deeply intertwined concepts that are paramount for real-world deployment and are consistently highlighted as critical research frontiers in recent surveys. Generalization refers to the ability of a system to perform well in new, unseen environments, a task made difficult by the heavy reliance of modern Semantic SLAM on deep learning models. This creates a ``domain gap,'' where a perception model trained on one type of data (\textit{e.g.}, indoor offices) fails to generalize to another (\textit{e.g.}, a residential home), representing a primary bottleneck. The shift towards foundation models that learn from web-scale data and the development of online adaptation methods are promising avenues for mitigating this issue. Robustness, on the other hand, is the ability of a SLAM system to maintain consistent and bounded error performance even when faced with challenging conditions that violate its core assumptions, such as aggressive motion, sensor noise, or visual degradation. As formalized in studies on measuring SLAM robustness, this can be quantified by observing a system's failure rate and accuracy degradation under controlled ``perturbations''. Many recent systems are explicitly designed with this in mind; for example, RS-SLAM~\cite{ran2021rs} is designed to be a ``Robust Semantic SLAM'' system for dynamic environments, while Kimera-Multi~\cite{tian2022kimera} is engineered to be a ``Robust, Distributed'' system for multi-robot teams that can handle intermittent communication. Ultimately, a truly robust system must also be capable of graceful failure. This requires the system to quantify its own uncertainty and, upon detecting a high likelihood of failure, enter a safe mode (\textit{e.g.}, stopping or asking for help) rather than producing a confident but catastrophically incorrect result. This ability to reason about its own limitations is a hallmark of a truly mature and deployable autonomous system.

\subsection{Reproducible Research}
The increasing complexity of modern Semantic SLAM systems, which often integrate multiple deep learning models, complex data fusion pipelines, and sophisticated back-end optimizers, makes reproducibility a significant and persistent challenge. This has long been an issue in the robotics community, where real-life experiments are often difficult to replicate across different research groups due to variations in hardware, software environments, and physical conditions. While static datasets are commonly used for benchmarking the perception and reconstruction components of an SLAM system, they cannot fully capture the nuances of a complete, end-to-end system's performance, which is highly sensitive to implementation details and parameter tuning. As a result, even when using the same dataset, it is often difficult to make a fair, one-to-one comparison between different published methods, a challenge consistently highlighted in recent surveys. To address this, there is a dire need for a continued and expanded commitment within the community to open-sourcing code and models. Open-source frameworks, such as the libraries provided for Kimera~\cite{rosinol2020kimera} and Hydra~\cite{hughes2022hydra}, are invaluable assets; they not only allow other researchers to verify reported results but also provide a common baseline upon which new algorithms can be built and fairly compared. By allowing others to modify specific modules (\textit{e.g.}, only the data association or loop closure component), these frameworks make it much more straightforward to isolate and evaluate the impact of a novel contribution. In summary, for the field of Semantic SLAM to mature and achieve real-world impact, improving the reproducibility of its research is essential, requiring a concerted effort towards open-sourcing systems, establishing more comprehensive benchmarks, and agreeing upon a unified set of evaluation metrics that capture both geometric accuracy and semantic quality.

\subsection{Practical Applications}
Ultimately, the goal of Semantic SLAM research is to enable autonomous systems to perform useful tasks in the real world. For the field to have a broad impact, research must be increasingly driven by the demands of practical applications, moving beyond trajectory accuracy on benchmarks to task-based performance in complex, human-centric environments. The rich, structured world models produced by Semantic SLAM are a crucial enabling technology for a wide range of applications. In service robotics and assisted living, a semantic map allows a robot to understand and execute natural language commands like ``bring me the cup from the kitchen table,'' a capability that systems like QueSTMaps~\cite{mehan2024questmaps} and LEXIS~\cite{kassab2024language} are designed to support. For autonomous driving, vehicles require rich, semantic maps of road networks that include lane markings, traffic signs, and dynamic agents, a challenge addressed by robust systems like DynaSLAM II~\cite{bescos2021dynaslam} and RS-SLAM~\cite{ran2021rs}. In augmented and virtual reality, the high-fidelity, photorealistic maps created by modern continuous representation SLAM (\textit{e.g.}, Gaussian Splatting SLAM~\cite{matsuki2024gaussian}) are essential for seamlessly blending virtual content with the real world, allowing virtual objects to interact realistically with physical surfaces. Furthermore, in aerial robotics and inspection, semantic maps enable drones to understand and perform complex tasks, such as inspecting specific windows on a building or monitoring particular crops, a use case targeted by systems like VPS-SLAM~\cite{bavle2020vps}. Finally, in domains like search and rescue, a semantic map provides invaluable situational awareness, allowing first responders to quickly understand the layout of an unknown building and locate people or objects of interest. Addressing the specific challenges of these practical applications will continue to be a primary driver of innovation in the Semantic SLAM field.
\section{Conclusion}\label{sec:conclusion}

This survey has explored the rapidly advancing field of semantic visual SLAM, a critical evolution that enriches traditional geometric mapping with a high-level, human-like understanding of the environment. We began by analyzing the limitations of conventional SLAM systems in complex, dynamic scenes, establishing the need for a paradigm shift from purely geometric representations to context-aware world models. Our review systematically deconstructed the modern Semantic SLAM pipeline, analyzing its core components from semantic extraction and localization to the intricacies of semantic mapping, data association, fusion, and globally consistent loop closure optimization. Beyond summarizing the current state of these modular components, this survey has highlighted the transformative impact of two recent technological shifts. First, the move towards continuous representations—from implicit Neural Radiance Fields to the explicit, real-time capabilities of 3D Gaussian Splatting—is fundamentally reshaping the fidelity and realism of the mapping process. Second, the advent of Foundation Models is revolutionizing the perception and reasoning layers, moving the field from a ``closed-world'' assumption with a fixed vocabulary to a truly ``open-world'' paradigm, enabling zero-shot object recognition and language-grounded scene understanding. Despite this remarkable progress, we have also identified key open research questions that must be addressed for the field to mature. The need for more comprehensive benchmarks that capture long-term dynamics, the transition from passive mapping to active SLAM for task-driven exploration, and the grand challenge of achieving true lifelong operation in ever-changing environments remain significant frontiers. Furthermore, issues of generalization, robustness, and reproducible research are essential to transitioning these advanced algorithms from the laboratory to real-world applications. 

In conclusion, this survey consolidates existing knowledge, provides a structured overview of the state-of-the-art, and identifies the pivotal challenges and future directions that will inspire the next wave of innovation. By advancing Semantic SLAM, we aim not only to build more accurate maps but to drive the development of truly intelligent systems capable of perceiving, understanding, and meaningfully interacting with the complexity of the physical world.

\bibliographystyle{IEEEtran}
\bibliography{ref}

\begin{thebibliography}{100}
\providecommand{\url}[1]{#1}
\csname url@samestyle\endcsname
\providecommand{\newblock}{\relax}
\providecommand{\bibinfo}[2]{#2}
\providecommand{\BIBentrySTDinterwordspacing}{\spaceskip=0pt\relax}
\providecommand{\BIBentryALTinterwordstretchfactor}{4}
\providecommand{\BIBentryALTinterwordspacing}{\spaceskip=\fontdimen2\font plus
\BIBentryALTinterwordstretchfactor\fontdimen3\font minus \fontdimen4\font\relax}
\providecommand{\BIBforeignlanguage}[2]{{%
\expandafter\ifx\csname l@#1\endcsname\relax
\typeout{** WARNING: IEEEtran.bst: No hyphenation pattern has been}%
\typeout{** loaded for the language `#1'. Using the pattern for}%
\typeout{** the default language instead.}%
\else
\language=\csname l@#1\endcsname
\fi
#2}}
\providecommand{\BIBdecl}{\relax}
\BIBdecl

\bibitem{smith1986representation}
R.~C. Smith and P.~Cheeseman, ``On the representation and estimation of spatial uncertainty,'' \emph{The international journal of Robotics Research}, vol.~5, no.~4, pp. 56--68, 1986.

\bibitem{placed2023survey}
J.~A. Placed, J.~Strader, H.~Carrillo, N.~Atanasov, V.~Indelman, L.~Carlone, and J.~A. Castellanos, ``A survey on active simultaneous localization and mapping: State of the art and new frontiers,'' \emph{IEEE Transactions on Robotics}, 2023.

\bibitem{zhang2022visual}
S.~Zhang, S.~Zhao, D.~An, J.~Liu, H.~Wang, Y.~Feng, D.~Li, and R.~Zhao, ``Visual slam for underwater vehicles: A survey,'' \emph{Computer Science Review}, vol.~46, p. 100510, 2022.

\bibitem{canh2024s3m}
T.~N. Canh, V.-T. Nguyen, X.~HoangVan, A.~Elibol, and N.~Y. Chong, ``S3m: Semantic segmentation sparse mapping for uavs with rgb-d camera,'' in \emph{2024 IEEE/SICE International Symposium on System Integration (SII)}.\hskip 1em plus 0.5em minus 0.4em\relax IEEE, 2024, pp. 899--905.

\bibitem{cheng2022review}
J.~Cheng, L.~Zhang, Q.~Chen, X.~Hu, and J.~Cai, ``A review of visual slam methods for autonomous driving vehicles,'' \emph{Engineering Applications of Artificial Intelligence}, vol. 114, p. 104992, 2022.

\bibitem{lee2018monocular}
T.-j. Lee, C.-h. Kim, and D.-i.~D. Cho, ``A monocular vision sensor-based efficient slam method for indoor service robots,'' \emph{IEEE Transactions on Industrial Electronics}, vol.~66, no.~1, pp. 318--328, 2018.

\bibitem{munoz2018augmented}
F.~Munoz-Montoya, M.-C. Juan, M.~Mendez-Lopez, and C.~Fidalgo, ``Augmented reality based on slam to assess spatial short-term memory,'' \emph{IEEE Access}, vol.~7, pp. 2453--2466, 2018.

\bibitem{jiang2023slam}
X.~Jiang, L.~Zhu, J.~Liu, and A.~Song, ``A slam-based 6dof controller with smooth auto-calibration for virtual reality,'' \emph{The Visual Computer}, vol.~39, no.~9, pp. 3873--3886, 2023.

\bibitem{hess2016real}
W.~Hess, D.~Kohler, H.~Rapp, and D.~Andor, ``Real-time loop closure in 2d lidar slam,'' in \emph{2016 IEEE international conference on robotics and automation (ICRA)}.\hskip 1em plus 0.5em minus 0.4em\relax IEEE, 2016, pp. 1271--1278.

\bibitem{zou2021comparative}
Q.~Zou, Q.~Sun, L.~Chen, B.~Nie, and Q.~Li, ``A comparative analysis of lidar slam-based indoor navigation for autonomous vehicles,'' \emph{IEEE Transactions on Intelligent Transportation Systems}, vol.~23, no.~7, pp. 6907--6921, 2021.

\bibitem{kazerouni2022survey}
I.~A. Kazerouni, L.~Fitzgerald, G.~Dooly, and D.~Toal, ``A survey of state-of-the-art on visual slam,'' \emph{Expert Systems with Applications}, vol. 205, p. 117734, 2022.

\bibitem{campos2021orb}
C.~Campos, R.~Elvira, J.~J.~G. Rodr{\'\i}guez, J.~M. Montiel, and J.~D. Tard{\'o}s, ``Orb-slam3: An accurate open-source library for visual, visual--inertial, and multimap slam,'' \emph{IEEE Transactions on Robotics}, vol.~37, no.~6, pp. 1874--1890, 2021.

\bibitem{lu1997globally}
F.~Lu and E.~Milios, ``Globally consistent range scan alignment for environment mapping,'' \emph{Autonomous robots}, vol.~4, pp. 333--349, 1997.

\bibitem{ma2017multi}
L.~Ma, J.~St{\"u}ckler, C.~Kerl, and D.~Cremers, ``Multi-view deep learning for consistent semantic mapping with rgb-d cameras,'' in \emph{2017 IEEE/RSJ International Conference on Intelligent Robots and Systems (IROS)}.\hskip 1em plus 0.5em minus 0.4em\relax IEEE, 2017, pp. 598--605.

\bibitem{mccormac2017semanticfusion}
J.~McCormac, A.~Handa, A.~Davison, and S.~Leutenegger, ``Semanticfusion: Dense 3d semantic mapping with convolutional neural networks,'' in \emph{2017 IEEE International Conference on Robotics and automation (ICRA)}.\hskip 1em plus 0.5em minus 0.4em\relax IEEE, 2017, pp. 4628--4635.

\bibitem{sunderhauf2017meaningful}
N.~S{\"u}nderhauf, T.~T. Pham, Y.~Latif, M.~Milford, and I.~Reid, ``Meaningful maps with object-oriented semantic mapping,'' in \emph{2017 IEEE/RSJ International Conference on Intelligent Robots and Systems (IROS)}.\hskip 1em plus 0.5em minus 0.4em\relax IEEE, 2017, pp. 5079--5085.

\bibitem{chen2025semantic}
K.~Chen, J.~Xiao, J.~Liu, Q.~Tong, H.~Zhang, R.~Liu, J.~Zhang, A.~Ajoudani, and S.~Chen, ``Semantic visual simultaneous localization and mapping: A survey,'' \emph{IEEE Transactions on Intelligent Transportation Systems}, 2025.

\bibitem{canh2022multisensor}
T.~N. Canh, T.~S. Nguyen, C.~H. Quach, X.~HoangVan, and M.~D. Phung, ``Multisensor data fusion for reliable obstacle avoidance,'' in \emph{2022 11th International Conference on Control, Automation and Information Sciences (ICCAIS)}.\hskip 1em plus 0.5em minus 0.4em\relax IEEE, 2022, pp. 385--390.

\bibitem{dellaert2004semantic}
F.~Dellaert, D.~J. Bruemmer, and A.~C.~C. Workspace, ``Semantic slam for collaborative cognitive workspaces.'' in \emph{AAAI Technical Report (5)}, 2004, pp. 85--86.

\bibitem{zhao2019lidar}
Z.~Zhao, W.~Zhang, J.~Gu, J.~Yang, and K.~Huang, ``Lidar mapping optimization based on lightweight semantic segmentation,'' \emph{IEEE Transactions on Intelligent Vehicles}, vol.~4, no.~3, pp. 353--362, 2019.

\bibitem{jin2021semantic}
C.~Jin, A.~Elibol, P.~Zhu, and N.~Y. Chong, ``Semantic mapping based on image feature fusion in indoor environments,'' in \emph{2021 21st International Conference on Control, Automation and Systems (ICCAS)}.\hskip 1em plus 0.5em minus 0.4em\relax IEEE, 2021, pp. 693--698.

\bibitem{canh2023object}
T.~N. Canh, A.~Elibol, N.~Y. Chong, and X.~HoangVan, ``Object-oriented semantic mapping for reliable uavs navigation,'' in \emph{2023 12th International Conference on Control, Automation and Information Sciences (ICCAIS)}.\hskip 1em plus 0.5em minus 0.4em\relax IEEE, 2023, pp. 139--144.

\bibitem{mccormac2018fusion++}
J.~McCormac, R.~Clark, M.~Bloesch, A.~Davison, and S.~Leutenegger, ``Fusion++: Volumetric object-level slam,'' in \emph{2018 international conference on 3D vision (3DV)}.\hskip 1em plus 0.5em minus 0.4em\relax IEEE, 2018, pp. 32--41.

\bibitem{hempel2022online}
T.~Hempel and A.~Al-Hamadi, ``An online semantic mapping system for extending and enhancing visual slam,'' \emph{Engineering Applications of Artificial Intelligence}, vol. 111, p. 104830, 2022.

\bibitem{ruan2022semantic}
X.~Ruan, P.~Guo, and J.~Huang, ``A semantic octomap mapping method based on cbam-pspnet,'' \emph{Journal of web engineering}, vol.~21, no.~3, pp. 879--910, 2022.

\bibitem{liu2023data}
Z.~Liu, P.~van Oosterom, J.~Balado, A.~Swart, and B.~Beers, ``Data frame aware optimized octomap-based dynamic object detection and removal in mobile laser scanning data,'' \emph{Alexandria Engineering Journal}, vol.~74, pp. 327--344, 2023.

\bibitem{li2022vox}
H.~Li, X.~Yang, H.~Zhai, Y.~Liu, H.~Bao, and G.~Zhang, ``Vox-surf: Voxel-based implicit surface representation,'' \emph{IEEE Transactions on Visualization and Computer Graphics}, vol.~30, no.~3, pp. 1743--1755, 2022.

\bibitem{matez2024voxeland}
J.-L. Matez-Bandera, P.~Ojeda, J.~Monroy, J.~Gonzalez-Jimenez, and J.-R. Ruiz-Sarmiento, ``Voxeland: Probabilistic instance-aware semantic mapping with evidence-based uncertainty quantification,'' \emph{arXiv preprint arXiv:2411.08727}, 2024.

\bibitem{zaganidis2018integrating}
A.~Zaganidis, L.~Sun, T.~Duckett, and G.~Cielniak, ``Integrating deep semantic segmentation into 3-d point cloud registration,'' \emph{IEEE Robotics and automation letters}, vol.~3, no.~4, pp. 2942--2949, 2018.

\bibitem{chen2019suma++}
X.~Chen, A.~Milioto, E.~Palazzolo, P.~Giguere, J.~Behley, and C.~Stachniss, ``Suma++: Efficient lidar-based semantic slam,'' in \emph{2019 IEEE/RSJ International Conference on Intelligent Robots and Systems (IROS)}.\hskip 1em plus 0.5em minus 0.4em\relax IEEE, 2019, pp. 4530--4537.

\bibitem{yi2009active}
C.~Yi, I.~H. Suh, G.~H. Lim, and B.-U. Choi, ``Active-semantic localization with a single consumer-grade camera,'' in \emph{2009 IEEE International Conference on Systems, Man and Cybernetics}.\hskip 1em plus 0.5em minus 0.4em\relax IEEE, 2009, pp. 2161--2166.

\bibitem{atanasov2016localization}
N.~Atanasov, M.~Zhu, K.~Daniilidis, and G.~J. Pappas, ``Localization from semantic observations via the matrix permanent,'' \emph{The International Journal of Robotics Research}, vol.~35, no. 1-3, pp. 73--99, 2016.

\bibitem{akai2020semantic}
N.~Akai, T.~Hirayama, and H.~Murase, ``Semantic localization considering uncertainty of object recognition,'' \emph{IEEE Robotics and Automation Letters}, vol.~5, no.~3, pp. 4384--4391, 2020.

\bibitem{akai2022mobile}
N.~Akai, ``Mobile robot localization considering uncertainty of depth regression from camera images,'' \emph{IEEE Robotics and Automation Letters}, vol.~7, no.~2, pp. 1431--1438, 2022.

\bibitem{toft2018semantic}
C.~Toft, E.~Stenborg, L.~Hammarstrand, L.~Brynte, M.~Pollefeys, T.~Sattler, and F.~Kahl, ``Semantic match consistency for long-term visual localization,'' in \emph{Proceedings of the European Conference on Computer Vision (ECCV)}, 2018, pp. 383--399.

\bibitem{schonberger2018semantic}
J.~L. Sch{\"o}nberger, M.~Pollefeys, A.~Geiger, and T.~Sattler, ``Semantic visual localization,'' in \emph{Proceedings of the IEEE conference on computer vision and pattern recognition}, 2018, pp. 6896--6906.

\bibitem{xiao2019dynamic}
L.~Xiao, J.~Wang, X.~Qiu, Z.~Rong, and X.~Zou, ``Dynamic-slam: Semantic monocular visual localization and mapping based on deep learning in dynamic environment,'' \emph{Robotics and Autonomous Systems}, vol. 117, pp. 1--16, 2019.

\bibitem{wu2022learning}
J.~Wu, Q.~Shi, Q.~Lu, X.~Liu, X.~Zhu, and Z.~Lin, ``Learning invariant semantic representation for long-term robust visual localization,'' \emph{Engineering Applications of Artificial Intelligence}, vol. 111, p. 104793, 2022.

\bibitem{zhang2025bev}
Z.~Zhang, M.~Xu, W.~Zhou, T.~Peng, L.~Li, and S.~Poslad, ``Bev-locator: An end-to-end visual semantic localization network using multi-view images,'' \emph{Science China Information Sciences}, vol.~68, no.~2, p. 122106, 2025.

\bibitem{yang2019cubeslam}
S.~Yang and S.~Scherer, ``Cubeslam: Monocular 3-d object slam,'' \emph{IEEE Transactions on Robotics}, vol.~35, no.~4, pp. 925--938, 2019.

\bibitem{bowman2017probabilistic}
S.~L. Bowman, N.~Atanasov, K.~Daniilidis, and G.~J. Pappas, ``Probabilistic data association for semantic slam,'' in \emph{2017 IEEE international conference on robotics and automation (ICRA)}.\hskip 1em plus 0.5em minus 0.4em\relax IEEE, 2017, pp. 1722--1729.

\bibitem{asgharivaskasi2021active}
A.~Asgharivaskasi and N.~Atanasov, ``Active bayesian multi-class mapping from range and semantic segmentation observations,'' in \emph{2021 IEEE international conference on robotics and automation (ICRA)}.\hskip 1em plus 0.5em minus 0.4em\relax IEEE, 2021, pp. 1--7.

\bibitem{chen2020sloam}
S.~W. Chen, G.~V. Nardari, E.~S. Lee, C.~Qu, X.~Liu, R.~A.~F. Romero, and V.~Kumar, ``Sloam: Semantic lidar odometry and mapping for forest inventory,'' \emph{IEEE Robotics and Automation Letters}, vol.~5, no.~2, pp. 612--619, 2020.

\bibitem{miller2022stronger}
I.~D. Miller, F.~Cladera, T.~Smith, C.~J. Taylor, and V.~Kumar, ``Stronger together: Air-ground robotic collaboration using semantics,'' \emph{IEEE Robotics and Automation Letters}, vol.~7, no.~4, pp. 9643--9650, 2022.

\bibitem{galindo2005multi}
C.~Galindo, A.~Saffiotti, S.~Coradeschi, P.~Buschka, J.-A. Fernandez-Madrigal, and J.~Gonz{\'a}lez, ``Multi-hierarchical semantic maps for mobile robotics,'' in \emph{2005 IEEE/RSJ international conference on intelligent robots and systems}.\hskip 1em plus 0.5em minus 0.4em\relax IEEE, 2005, pp. 2278--2283.

\bibitem{nicholson2018quadricslam}
L.~Nicholson, M.~Milford, and N.~S{\"u}nderhauf, ``Quadricslam: Dual quadrics from object detections as landmarks in object-oriented slam,'' \emph{IEEE Robotics and Automation Letters}, vol.~4, no.~1, pp. 1--8, 2018.

\bibitem{hosseinzadeh2019structure}
M.~Hosseinzadeh, Y.~Latif, T.~Pham, N.~Suenderhauf, and I.~Reid, ``Structure aware slam using quadrics and planes,'' in \emph{Computer Vision--ACCV 2018: 14th Asian Conference on Computer Vision, Perth, Australia, December 2--6, 2018, Revised Selected Papers, Part III 14}.\hskip 1em plus 0.5em minus 0.4em\relax Springer, 2019, pp. 410--426.

\bibitem{wang2022drg}
Y.~Wang, K.~Xu, Y.~Tian, and X.~Ding, ``Drg-slam: a semantic rgb-d slam using geometric features for indoor dynamic scene,'' in \emph{2022 IEEE/RSJ International Conference on Intelligent Robots and Systems (IROS)}.\hskip 1em plus 0.5em minus 0.4em\relax IEEE, 2022, pp. 1352--1359.

\bibitem{doherty2019multimodal}
K.~Doherty, D.~Fourie, and J.~Leonard, ``Multimodal semantic slam with probabilistic data association,'' in \emph{2019 international conference on robotics and automation (ICRA)}.\hskip 1em plus 0.5em minus 0.4em\relax IEEE, 2019, pp. 2419--2425.

\bibitem{doherty2020probabilistic}
K.~J. Doherty, D.~P. Baxter, E.~Schneeweiss, and J.~J. Leonard, ``Probabilistic data association via mixture models for robust semantic slam,'' in \emph{2020 IEEE International Conference on Robotics and Automation (ICRA)}.\hskip 1em plus 0.5em minus 0.4em\relax IEEE, 2020, pp. 1098--1104.

\bibitem{michael2022probabilistic}
E.~Michael, T.~Summers, T.~A. Wood, C.~Manzie, and I.~Shames, ``Probabilistic data association for semantic slam at scale,'' in \emph{2022 IEEE/RSJ International Conference on Intelligent Robots and Systems (IROS)}.\hskip 1em plus 0.5em minus 0.4em\relax IEEE, 2022, pp. 4359--4364.

\bibitem{hornung2013octomap}
A.~Hornung, K.~M. Wurm, M.~Bennewitz, C.~Stachniss, and W.~Burgard, ``Octomap: An efficient probabilistic 3d mapping framework based on octrees,'' \emph{Autonomous robots}, vol.~34, pp. 189--206, 2013.

\bibitem{pagliari2014kinect}
D.~Pagliari, F.~Menna, R.~Roncella, F.~Remondino, and L.~Pinto, ``Kinect fusion improvement using depth camera calibration,'' \emph{The International Archives of the Photogrammetry, Remote Sensing and Spatial Information Sciences}, vol.~40, pp. 479--485, 2014.

\bibitem{zhu2022nice}
Z.~Zhu, S.~Peng, V.~Larsson, W.~Xu, H.~Bao, Z.~Cui, M.~R. Oswald, and M.~Pollefeys, ``Nice-slam: Neural implicit scalable encoding for slam,'' in \emph{Proceedings of the IEEE/CVF conference on computer vision and pattern recognition}, 2022, pp. 12\,786--12\,796.

\bibitem{yang2022vox}
X.~Yang, H.~Li, H.~Zhai, Y.~Ming, Y.~Liu, and G.~Zhang, ``Vox-fusion: Dense tracking and mapping with voxel-based neural implicit representation,'' in \emph{2022 IEEE International Symposium on Mixed and Augmented Reality (ISMAR)}.\hskip 1em plus 0.5em minus 0.4em\relax IEEE, 2022, pp. 499--507.

\bibitem{matsuki2024gaussian}
H.~Matsuki, R.~Murai, P.~H. Kelly, and A.~J. Davison, ``Gaussian splatting slam,'' in \emph{Proceedings of the IEEE/CVF Conference on Computer Vision and Pattern Recognition}, 2024, pp. 18\,039--18\,048.

\bibitem{yan2024gs}
C.~Yan, D.~Qu, D.~Xu, B.~Zhao, Z.~Wang, D.~Wang, and X.~Li, ``Gs-slam: Dense visual slam with 3d gaussian splatting,'' in \emph{Proceedings of the IEEE/CVF Conference on Computer Vision and Pattern Recognition}, 2024, pp. 19\,595--19\,604.

\bibitem{li2024sgs}
M.~Li, S.~Liu, H.~Zhou, G.~Zhu, N.~Cheng, T.~Deng, and H.~Wang, ``Sgs-slam: Semantic gaussian splatting for neural dense slam,'' in \emph{European Conference on Computer Vision}.\hskip 1em plus 0.5em minus 0.4em\relax Springer, 2024, pp. 163--179.

\bibitem{li2025hier}
B.~Li, Z.~Cai, Y.-F. Li, I.~Reid, and H.~Rezatofighi, ``Hier-slam: Scaling-up semantics in slam with a hierarchically categorical gaussian splatting,'' \emph{arXiv preprint arXiv:2409.12518}, 2024.

\bibitem{kostavelis2015semantic}
I.~Kostavelis and A.~Gasteratos, ``Semantic mapping for mobile robotics tasks: A survey,'' \emph{Robotics and Autonomous Systems}, vol.~66, pp. 86--103, 2015.

\bibitem{sualeh2019simultaneous}
M.~Sualeh and G.-W. Kim, ``Simultaneous localization and mapping in the epoch of semantics: a survey,'' \emph{International Journal of Control, Automation and Systems}, vol.~17, no.~3, pp. 729--742, 2019.

\bibitem{garg2020semantics}
S.~Garg, N.~S{\"u}nderhauf, F.~Dayoub, D.~Morrison, A.~Cosgun, G.~Carneiro, Q.~Wu, T.-J. Chin, I.~Reid, S.~Gould \emph{et~al.}, ``Semantics for robotic mapping, perception and interaction: A survey,'' \emph{Foundations and Trends{\textregistered} in Robotics}, vol.~8, no. 1--2, pp. 1--224, 2020.

\bibitem{wen2021semantic}
S.~Wen, P.~Li, Y.~Zhao, H.~Zhang, F.~Sun, and Z.~Wang, ``Semantic visual slam in dynamic environment,'' \emph{Autonomous Robots}, vol.~45, no.~4, pp. 493--504, 2021.

\bibitem{lai2022review}
T.~Lai, ``A review on visual-slam: Advancements from geometric modelling to learning-based semantic scene understanding using multi-modal sensor fusion,'' \emph{Sensors}, vol.~22, no.~19, p. 7265, 2022.

\bibitem{chen2022overview}
W.~Chen, G.~Shang, A.~Ji, C.~Zhou, X.~Wang, C.~Xu, Z.~Li, and K.~Hu, ``An overview on visual slam: From tradition to semantic,'' \emph{Remote Sensing}, vol.~14, no.~13, p. 3010, 2022.

\bibitem{pu2023visual}
H.~Pu, J.~Luo, G.~Wang, T.~Huang, and H.~Liu, ``Visual slam integration with semantic segmentation and deep learning: A review,'' \emph{IEEE Sensors Journal}, vol.~23, no.~19, pp. 22\,119--22\,138, 2023.

\bibitem{bavle2023slam}
H.~Bavle, J.~L. Sanchez-Lopez, C.~Cimarelli, A.~Tourani, and H.~Voos, ``From slam to situational awareness: Challenges and survey,'' \emph{Sensors}, vol.~23, no.~10, p. 4849, 2023.

\bibitem{wang2024survey}
Y.~Wang, Y.~Tian, J.~Chen, K.~Xu, and X.~Ding, ``A survey of visual slam in dynamic environment: The evolution from geometric to semantic approaches,'' \emph{IEEE Transactions on Instrumentation and Measurement}, 2024.

\bibitem{georgevichsemantic}
B.~Georgevich~Ferreira, A.~J. Miranda~de Sousa, and L.~Reis, ``Semantic mapping for robotics: Survey, trends and challenges,'' \emph{Trends and Challenges}, 2025.

\bibitem{song2025semantic}
X.~Song, X.~Liang, and Z.~Huaidong, ``Semantic mapping techniques for indoor mobile robots: Review and prospect,'' \emph{Measurement and Control}, vol.~58, no.~3, pp. 377--393, 2025.

\bibitem{georgevich_ferreira_semantic_2025}
\BIBentryALTinterwordspacing
B.~Georgevich~Ferreira, A.~J. Miranda De~Sousa, and L.~Reis, ``\BIBforeignlanguage{en}{Semantic {Mapping} for {Robotics}: {Survey}, {Trends} and {Challenges}},'' 2025. [Online]. Available: \url{https://www.ssrn.com/abstract=5085045}
\BIBentrySTDinterwordspacing

\bibitem{tateno2017cnn}
K.~Tateno, F.~Tombari, I.~Laina, and N.~Navab, ``Cnn-slam: Real-time dense monocular slam with learned depth prediction,'' in \emph{Proceedings of the IEEE conference on computer vision and pattern recognition}, 2017, pp. 6243--6252.

\bibitem{runz2017co}
M.~R{\"u}nz and L.~Agapito, ``Co-fusion: Real-time segmentation, tracking and fusion of multiple objects,'' in \emph{2017 IEEE International Conference on Robotics and Automation (ICRA)}.\hskip 1em plus 0.5em minus 0.4em\relax IEEE, 2017, pp. 4471--4478.

\bibitem{lianos2018vso}
K.-N. Lianos, J.~L. Schonberger, M.~Pollefeys, and T.~Sattler, ``Vso: Visual semantic odometry,'' in \emph{Proceedings of the European conference on computer vision (ECCV)}, 2018, pp. 234--250.

\bibitem{yu2018ds}
C.~Yu, Z.~Liu, X.-J. Liu, F.~Xie, Y.~Yang, Q.~Wei, and Q.~Fei, ``Ds-slam: A semantic visual slam towards dynamic environments,'' in \emph{2018 IEEE/RSJ international conference on intelligent robots and systems (IROS)}.\hskip 1em plus 0.5em minus 0.4em\relax IEEE, 2018, pp. 1168--1174.

\bibitem{runz2018maskfusion}
M.~Runz, M.~Buffier, and L.~Agapito, ``Maskfusion: Real-time recognition, tracking and reconstruction of multiple moving objects,'' in \emph{2018 IEEE international symposium on mixed and augmented reality (ISMAR)}.\hskip 1em plus 0.5em minus 0.4em\relax IEEE, 2018, pp. 10--20.

\bibitem{wu2020eao}
Y.~Wu, Y.~Zhang, D.~Zhu, Y.~Feng, S.~Coleman, and D.~Kerr, ``Eao-slam: Monocular semi-dense object slam based on ensemble data association,'' in \emph{2020 IEEE/RSJ International Conference on Intelligent Robots and Systems (IROS)}.\hskip 1em plus 0.5em minus 0.4em\relax IEEE, 2020, pp. 4966--4973.

\bibitem{narita2019panopticfusion}
G.~Narita, T.~Seno, T.~Ishikawa, and Y.~Kaji, ``Panopticfusion: Online volumetric semantic mapping at the level of stuff and things,'' in \emph{2019 IEEE/RSJ International Conference on Intelligent Robots and Systems (IROS)}.\hskip 1em plus 0.5em minus 0.4em\relax IEEE, 2019, pp. 4205--4212.

\bibitem{rosinol2020kimera}
A.~Rosinol, M.~Abate, Y.~Chang, and L.~Carlone, ``Kimera: an open-source library for real-time metric-semantic localization and mapping,'' in \emph{2020 IEEE International Conference on Robotics and Automation (ICRA)}.\hskip 1em plus 0.5em minus 0.4em\relax IEEE, 2020, pp. 1689--1696.

\bibitem{hughes2022hydra}
N.~Hughes, Y.~Chang, and L.~Carlone, ``Hydra: A real-time spatial perception system for 3d scene graph construction and optimization,'' \emph{arXiv preprint arXiv:2201.13360}, 2022.

\bibitem{sucar2021imap}
E.~Sucar, S.~Liu, J.~Ortiz, and A.~J. Davison, ``imap: Implicit mapping and positioning in real-time,'' in \emph{Proceedings of the IEEE/CVF international conference on computer vision}, 2021, pp. 6229--6238.

\bibitem{sandstrom2023point}
E.~Sandstr{\"o}m, Y.~Li, L.~Van~Gool, and M.~R. Oswald, ``Point-slam: Dense neural point cloud-based slam,'' in \emph{Proceedings of the IEEE/CVF International Conference on Computer Vision}, 2023, pp. 18\,433--18\,444.

\bibitem{jiang2023h}
C.~Jiang, H.~Zhang, P.~Liu, Z.~Yu, H.~Cheng, B.~Zhou, and S.~Shen, ``H $ \_ $\{$2$\}$ $-mapping: Real-time dense mapping using hierarchical hybrid representation,'' \emph{IEEE Robotics and Automation Letters}, vol.~8, no.~10, pp. 6787--6794, 2023.

\bibitem{zhu2024sni}
S.~Zhu, G.~Wang, H.~Blum, J.~Liu, L.~Song, M.~Pollefeys, and H.~Wang, ``Sni-slam: Semantic neural implicit slam,'' in \emph{Proceedings of the IEEE/CVF Conference on Computer Vision and Pattern Recognition}, 2024, pp. 21\,167--21\,177.

\bibitem{jiang2024h3}
C.~Jiang, Y.~Luo, B.~Zhou, and S.~Shen, ``H3-mapping: Quasi-heterogeneous feature grids for real-time dense mapping using hierarchical hybrid representation,'' \emph{IEEE Robotics and Automation Letters}, 2024.

\bibitem{schischka2024dynamon}
N.~Schischka, H.~Schieber, M.~A. Karaoglu, M.~Gorgulu, F.~Gr{\"o}tzner, A.~Ladikos, N.~Navab, D.~Roth, and B.~Busam, ``Dynamon: Motion-aware fast and robust camera localization for dynamic neural radiance fields,'' \emph{IEEE Robotics and Automation Letters}, 2024.

\bibitem{keetha2024splatam}
N.~Keetha, J.~Karhade, K.~M. Jatavallabhula, G.~Yang, S.~Scherer, D.~Ramanan, and J.~Luiten, ``Splatam: Splat track \& map 3d gaussians for dense rgb-d slam,'' in \emph{Proceedings of the IEEE/CVF Conference on Computer Vision and Pattern Recognition}, 2024, pp. 21\,357--21\,366.

\bibitem{ji2024neds}
Y.~Ji, Y.~Liu, G.~Xie, B.~Ma, Z.~Xie, and H.~Liu, ``Neds-slam: A neural explicit dense semantic slam framework using 3d gaussian splatting,'' \emph{IEEE Robotics and Automation Letters}, 2024.

\bibitem{zhu2025semgauss}
S.~Zhu, R.~Qin, G.~Wang, J.~Liu, and H.~Wang, ``Semgauss-slam: Dense semantic gaussian splatting slam,'' \emph{arXiv preprint arXiv:2503.07494}, 2025.

\bibitem{jatavallabhula2023conceptfusion}
K.~M. Jatavallabhula, A.~Kuwajerwala, Q.~Gu, M.~Omama, T.~Chen, A.~Maalouf, S.~Li, G.~Iyer, S.~Saryazdi, N.~Keetha \emph{et~al.}, ``Conceptfusion: Open-set multimodal 3d mapping,'' \emph{arXiv preprint arXiv:2302.07241}, 2023.

\bibitem{liu2024fm}
C.~Liu, K.~Wang, J.~Shi, Z.~Qiao, and S.~Shen, ``Fm-fusion: Instance-aware semantic mapping boosted by vision-language foundation models,'' \emph{IEEE Robotics and Automation Letters}, vol.~9, no.~3, pp. 2232--2239, 2024.

\bibitem{gu2024conceptgraphs}
Q.~Gu, A.~Kuwajerwala, S.~Morin, K.~M. Jatavallabhula, B.~Sen, A.~Agarwal, C.~Rivera, W.~Paul, K.~Ellis, R.~Chellappa \emph{et~al.}, ``Conceptgraphs: Open-vocabulary 3d scene graphs for perception and planning,'' in \emph{2024 IEEE International Conference on Robotics and Automation (ICRA)}.\hskip 1em plus 0.5em minus 0.4em\relax IEEE, 2024, pp. 5021--5028.

\bibitem{singh2024loss}
K.~Singh, T.~Magoun, and J.~J. Leonard, ``Loss-slam: Lightweight open-set semantic simultaneous localization and mapping,'' \emph{arXiv preprint arXiv:2404.04377}, 2024.

\bibitem{kassab2024language}
C.~Kassab, M.~Mattamala, L.~Zhang, and M.~Fallon, ``Language-extended indoor slam (lexis): A versatile system for real-time visual scene understanding,'' in \emph{2024 IEEE International Conference on Robotics and Automation (ICRA)}.\hskip 1em plus 0.5em minus 0.4em\relax IEEE, 2024, pp. 15\,988--15\,994.

\bibitem{li2025hier++}
B.~Li, V.~C. Hao, P.~J. Stuckey, I.~Reid, and H.~Rezatofighi, ``Hier-slam++: Neuro-symbolic semantic slam with a hierarchically categorical gaussian splatting,'' \emph{arXiv preprint arXiv:2502.14931}, 2025.

\bibitem{tian2022kimera}
Y.~Tian, Y.~Chang, F.~H. Arias, C.~Nieto-Granda, J.~P. How, and L.~Carlone, ``Kimera-multi: Robust, distributed, dense metric-semantic slam for multi-robot systems,'' \emph{IEEE Transactions on Robotics}, vol.~38, no.~4, 2022.

\bibitem{liu2024slideslam}
X.~Liu, J.~Lei, A.~Prabhu, Y.~Tao, I.~Spasojevic, P.~Chaudhari, N.~Atanasov, and V.~Kumar, ``Slideslam: Sparse, lightweight, decentralized metric-semantic slam for multi-robot navigation,'' \emph{arXiv preprint arXiv:2406.17249}, 2024.

\bibitem{yu2025hammer}
J.~Yu, T.~Chen, and M.~Schwager, ``Hammer: Heterogeneous, multi-robot semantic gaussian splatting,'' \emph{IEEE Robotics and Automation Letters}, 2025.

\bibitem{chen2022semantic}
K.~Chen, J.~Zhang, J.~Liu, Q.~Tong, R.~Liu, and S.~Chen, ``Semantic visual simultaneous localization and mapping: A survey,'' \emph{arXiv preprint arXiv:2209.06428}, 2022.

\bibitem{thrun2002probabilistic}
S.~Thrun, ``Probabilistic robotics,'' \emph{Communications of the ACM}, vol.~45, no.~3, pp. 52--57, 2002.

\bibitem{thrun2005probabilistic}
S.~Thrun, W.~Burgard, and D.~Fox, \emph{Probabilistic roboticsv2}.\hskip 1em plus 0.5em minus 0.4em\relax MIT press, 2005.

\bibitem{sigaud2013markov}
O.~Sigaud and O.~Buffet, \emph{Markov decision processes in artificial intelligence}.\hskip 1em plus 0.5em minus 0.4em\relax John Wiley \& Sons, 2013.

\bibitem{long2015fully}
J.~Long, E.~Shelhamer, and T.~Darrell, ``Fully convolutional networks for semantic segmentation,'' \emph{Proceedings of the IEEE conference on computer vision and pattern recognition}, pp. 3431--3440, 2015.

\bibitem{he2017mask}
K.~He, G.~Gkioxari, P.~Doll{'a}r, and R.~Girshick, ``Mask r-cnn,'' \emph{Proceedings of the IEEE international conference on computer vision}, pp. 2961--2969, 2017.

\bibitem{indelman2015planning}
V.~Indelman, L.~Carlone, and F.~Dellaert, ``Planning in the continuous domain: A generalized belief space approach for autonomous navigation in unknown environments,'' \emph{The International Journal of Robotics Research}, vol.~34, no.~7, pp. 849--882, 2015.

\bibitem{sunderhauf2017dual}
N.~S{\"u}nderhauf and M.~Milford, ``Dual quadrics from object detection boundingboxes as landmark representations in slam,'' \emph{arXiv preprint arXiv:1708.00965}, 2017.

\bibitem{salas2013slam++}
R.~F. Salas-Moreno, R.~A. Newcombe, H.~Strasdat, P.~H. Kelly, and A.~J. Davison, ``Slam++: Simultaneous localisation and mapping at the level of objects,'' in \emph{Proceedings of the IEEE conference on computer vision and pattern recognition}, 2013, pp. 1352--1359.

\bibitem{chen2017rethinking}
L.-C. Chen, G.~Papandreou, F.~Schroff, and H.~Adam, ``Rethinking atrous convolution for semantic image segmentation,'' \emph{arXiv preprint arXiv:1706.05587}, 2017.

\bibitem{redmon2016you}
J.~Redmon, S.~Divvala, R.~Girshick, and A.~Farhadi, ``You only look once: Unified, real-time object detection,'' in \emph{Proceedings of the IEEE conference on computer vision and pattern recognition}, 2016, pp. 779--788.

\bibitem{redmon2017yolo9000}
J.~Redmon and A.~Farhadi, ``Yolo9000: better, faster, stronger,'' in \emph{Proceedings of the IEEE conference on computer vision and pattern recognition}, 2017, pp. 7263--7271.

\bibitem{redmon2018yolov3}
------, ``Yolov3: An incremental improvement,'' \emph{arXiv preprint arXiv:1804.02767}, 2018.

\bibitem{bochkovskiy2020yolov4}
A.~Bochkovskiy, C.-Y. Wang, and H.-Y.~M. Liao, ``Yolov4: Optimal speed and accuracy of object detection,'' \emph{arXiv preprint arXiv:2004.10934}, 2020.

\bibitem{yolov5}
\BIBentryALTinterwordspacing
G.~Jocher, ``Ultralytics yolov5,'' 2020. [Online]. Available: \url{https://github.com/ultralytics/yolov5}
\BIBentrySTDinterwordspacing

\bibitem{li2022yolov6}
C.~Li, L.~Li, H.~Jiang, K.~Weng, Y.~Geng, L.~Li, Z.~Ke, Q.~Li, M.~Cheng, W.~Nie \emph{et~al.}, ``Yolov6: A single-stage object detection framework for industrial applications,'' \emph{arXiv preprint arXiv:2209.02976}, 2022.

\bibitem{wang2023yolov7}
C.-Y. Wang, A.~Bochkovskiy, and H.-Y.~M. Liao, ``Yolov7: Trainable bag-of-freebies sets new state-of-the-art for real-time object detectors,'' in \emph{Proceedings of the IEEE/CVF conference on computer vision and pattern recognition}, 2023, pp. 7464--7475.

\bibitem{yolov8}
\BIBentryALTinterwordspacing
G.~Jocher, J.~Qiu, and A.~Chaurasia, ``{Ultralytics YOLO},'' Jan. 2023. [Online]. Available: \url{https://github.com/ultralytics/ultralytics}
\BIBentrySTDinterwordspacing

\bibitem{wang2024yolov9}
C.-Y. Wang, I.-H. Yeh, and H.-Y. Mark~Liao, ``Yolov9: Learning what you want to learn using programmable gradient information,'' in \emph{European conference on computer vision}.\hskip 1em plus 0.5em minus 0.4em\relax Springer, 2024, pp. 1--21.

\bibitem{liu2016ssd}
W.~Liu, D.~Anguelov, D.~Erhan, C.~Szegedy, S.~Reed, C.-Y. Fu, and A.~C. Berg, ``Ssd: Single shot multibox detector,'' in \emph{Computer Vision--ECCV 2016: 14th European Conference, Amsterdam, The Netherlands, October 11--14, 2016, Proceedings, Part I 14}.\hskip 1em plus 0.5em minus 0.4em\relax Springer, 2016, pp. 21--37.

\bibitem{girshick2014rich}
R.~Girshick, J.~Donahue, T.~Darrell, and J.~Malik, ``Rich feature hierarchies for accurate object detection and semantic segmentation,'' in \emph{Proceedings of the IEEE conference on computer vision and pattern recognition}, 2014, pp. 580--587.

\bibitem{girshick2015fast}
R.~Girshick, ``Fast r-cnn,'' in \emph{Proceedings of the IEEE international conference on computer vision}, 2015, pp. 1440--1448.

\bibitem{ren2015faster}
S.~Ren, K.~He, R.~Girshick, and J.~Sun, ``Faster r-cnn: Towards real-time object detection with region proposal networks,'' \emph{Advances in neural information processing systems}, vol.~28, 2015.

\bibitem{carion2020end}
N.~Carion, F.~Massa, G.~Synnaeve, N.~Usunier, A.~Kirillov, and S.~Zagoruyko, ``End-to-end object detection with transformers,'' in \emph{European conference on computer vision}.\hskip 1em plus 0.5em minus 0.4em\relax Springer, 2020, pp. 213--229.

\bibitem{zhang2022dino}
H.~Zhang, F.~Li, S.~Liu, L.~Zhang, H.~Su, J.~Zhu, L.~M. Ni, and H.-Y. Shum, ``Dino: Detr with improved denoising anchor boxes for end-to-end object detection,'' \emph{arXiv preprint arXiv:2203.03605}, 2022.

\bibitem{zhou2018voxelnet}
Y.~Zhou and O.~Tuzel, ``Voxelnet: End-to-end learning for point cloud based 3d object detection,'' in \emph{Proceedings of the IEEE conference on computer vision and pattern recognition}, 2018, pp. 4490--4499.

\bibitem{qi2018frustum}
C.~R. Qi, W.~Liu, C.~Wu, H.~Su, and L.~J. Guibas, ``Frustum pointnets for 3d object detection from rgb-d data,'' in \emph{Proceedings of the IEEE conference on computer vision and pattern recognition}, 2018, pp. 918--927.

\bibitem{wang2019densefusion}
C.~Wang, D.~Xu, Y.~Zhu, R.~Mart{\'\i}n-Mart{\'\i}n, C.~Lu, L.~Fei-Fei, and S.~Savarese, ``Densefusion: 6d object pose estimation by iterative dense fusion,'' in \emph{Proceedings of the IEEE/CVF conference on computer vision and pattern recognition}, 2019, pp. 3343--3352.

\bibitem{soares2021crowd}
J.~C.~V. Soares, M.~Gattass, and M.~A. Meggiolaro, ``Crowd-slam: visual slam towards crowded environments using object detection,'' \emph{Journal of Intelligent \& Robotic Systems}, vol. 102, no.~2, p.~50, 2021.

\bibitem{wu2022yolo}
W.~Wu, L.~Guo, H.~Gao, Z.~You, Y.~Liu, and Z.~Chen, ``Yolo-slam: A semantic slam system towards dynamic environment with geometric constraint,'' \emph{Neural Computing and Applications}, pp. 1--16, 2022.

\bibitem{xia2023yolo}
X.~Xia, P.~Zhang, and J.~Sun, ``Yolo-based semantic segmentation for dynamic removal in visual-inertial slam,'' in \emph{Chinese Intelligent Systems Conference}.\hskip 1em plus 0.5em minus 0.4em\relax Springer, 2023, pp. 377--389.

\bibitem{gong2024real}
C.~Gong, Y.~Sun, C.~Zou, B.~Tao, L.~Huang, Z.~Fang, and D.~Tang, ``Real-time visual slam based yolo-fastest for dynamic scenes,'' \emph{Measurement Science and Technology}, vol.~35, no.~5, p. 056305, 2024.

\bibitem{rui2021multi}
C.~Rui, Y.~Liu, J.~Shen, Z.~Li, and Z.~Xie, ``A multi-sensory blind guidance system based on yolo and orb-slam,'' in \emph{2021 IEEE International Conference on Progress in Informatics and Computing (PIC)}.\hskip 1em plus 0.5em minus 0.4em\relax IEEE, 2021, pp. 409--414.

\bibitem{xie2022multi}
Z.~Xie, Z.~Li, Y.~Zhang, J.~Zhang, F.~Liu, and W.~Chen, ``A multi-sensory guidance system for the visually impaired using yolo and orb-slam,'' \emph{Information}, vol.~13, no.~7, p. 343, 2022.

\bibitem{li2023rgbd}
G.~Li, H.~Fan, G.~Jiang, D.~Jiang, Y.~Liu, B.~Tao, and J.~Yun, ``Rgbd-slam based on object detection with two-stream yolov4-mobilenetv3 in autonomous driving,'' \emph{IEEE Transactions on Intelligent Transportation Systems}, vol.~25, no.~3, pp. 2847--2857, 2023.

\bibitem{shao2021faster}
C.~Shao, L.~Zhang, and W.~Pan, ``Faster r-cnn learning-based semantic filter for geometry estimation and its application in vslam systems,'' \emph{IEEE Transactions on Intelligent Transportation Systems}, vol.~23, no.~6, pp. 5257--5266, 2021.

\bibitem{zhang2022dynamic}
X.~Zhang, X.~Wang, and R.~Zhang, ``Dynamic semantics slam based on improved mask r-cnn,'' \emph{IEEE Access}, vol.~10, pp. 126\,525--126\,535, 2022.

\bibitem{soares2019visual}
J.~C.~V. Soares, M.~Gattass, and M.~A. Meggiolaro, ``Visual slam in human populated environments: exploring the trade-off between accuracy and speed of yolo and mask r-cnn,'' in \emph{2019 19th International Conference on Advanced Robotics (ICAR)}.\hskip 1em plus 0.5em minus 0.4em\relax IEEE, 2019, pp. 135--140.

\bibitem{zhang2018semantic}
L.~Zhang, L.~Wei, P.~Shen, W.~Wei, G.~Zhu, and J.~Song, ``Semantic slam based on object detection and improved octomap,'' \emph{IEEE Access}, vol.~6, pp. 75\,545--75\,559, 2018.

\bibitem{islam2025advancing}
Q.~U. Islam, F.~Khozaei, I.~Baig, D.~Ignatyev \emph{et~al.}, ``Advancing autonomous slam systems: Integrating yolo object detection and enhanced loop closure techniques for robust environment mapping,'' \emph{Robotics and Autonomous Systems}, vol. 185, p. 104871, 2025.

\bibitem{wu2024dyn}
P.~Wu, P.~Tong, X.~Zhou, and X.~Yang, ``Dyn-darkslam: Yolo-based visual slam in low-light conditions,'' in \emph{2024 IEEE 25th China Conference on System Simulation Technology and its Application (CCSSTA)}.\hskip 1em plus 0.5em minus 0.4em\relax IEEE, 2024, pp. 346--351.

\bibitem{ronneberger2015u}
O.~Ronneberger, P.~Fischer, and T.~Brox, ``U-net: Convolutional networks for biomedical image segmentation,'' in \emph{Medical image computing and computer-assisted intervention--MICCAI 2015: 18th international conference, Munich, Germany, October 5-9, 2015, proceedings, part III 18}.\hskip 1em plus 0.5em minus 0.4em\relax Springer, 2015, pp. 234--241.

\bibitem{kendall2015bayesian}
A.~Kendall, V.~Badrinarayanan, and R.~Cipolla, ``Bayesian segnet: Model uncertainty in deep convolutional encoder-decoder architectures for scene understanding,'' \emph{arXiv preprint arXiv:1511.02680}, 2015.

\bibitem{badrinarayanan2017segnet}
V.~Badrinarayanan, A.~Kendall, and R.~Cipolla, ``Segnet: A deep convolutional encoder-decoder architecture for image segmentation,'' \emph{IEEE transactions on pattern analysis and machine intelligence}, vol.~39, no.~12, pp. 2481--2495, 2017.

\bibitem{zhao2017pyramid}
H.~Zhao, J.~Shi, X.~Qi, X.~Wang, and J.~Jia, ``Pyramid scene parsing network,'' in \emph{Proceedings of the IEEE conference on computer vision and pattern recognition}, 2017, pp. 2881--2890.

\bibitem{chen2018encoder}
L.-C. Chen, Y.~Zhu, G.~Papandreou, F.~Schroff, and H.~Adam, ``Encoder-decoder with atrous separable convolution for semantic image segmentation,'' in \emph{Proceedings of the European conference on computer vision (ECCV)}, 2018, pp. 801--818.

\bibitem{wang2020solo}
X.~Wang, T.~Kong, C.~Shen, Y.~Jiang, and L.~Li, ``Solo: Segmenting objects by locations,'' in \emph{Computer Vision--ECCV 2020: 16th European Conference, Glasgow, UK, August 23--28, 2020, Proceedings, Part XVIII 16}.\hskip 1em plus 0.5em minus 0.4em\relax Springer, 2020, pp. 649--665.

\bibitem{bolya2019yolact}
D.~Bolya, C.~Zhou, F.~Xiao, and Y.~J. Lee, ``Yolact: Real-time instance segmentation,'' in \emph{Proceedings of the IEEE/CVF international conference on computer vision}, 2019, pp. 9157--9166.

\bibitem{liu2018path}
S.~Liu, L.~Qi, H.~Qin, J.~Shi, and J.~Jia, ``Path aggregation network for instance segmentation,'' in \emph{Proceedings of the IEEE conference on computer vision and pattern recognition}, 2018, pp. 8759--8768.

\bibitem{chen2019hybrid}
K.~Chen, J.~Pang, J.~Wang, Y.~Xiong, X.~Li, S.~Sun, W.~Feng, Z.~Liu, J.~Shi, W.~Ouyang \emph{et~al.}, ``Hybrid task cascade for instance segmentation,'' in \emph{Proceedings of the IEEE/CVF conference on computer vision and pattern recognition}, 2019, pp. 4974--4983.

\bibitem{cheng2022masked}
B.~Cheng, I.~Misra, A.~G. Schwing, A.~Kirillov, and R.~Girdhar, ``Masked-attention mask transformer for universal image segmentation,'' in \emph{Proceedings of the IEEE/CVF conference on computer vision and pattern recognition}, 2022, pp. 1290--1299.

\bibitem{fang2021instances}
Y.~Fang, S.~Yang, X.~Wang, Y.~Li, C.~Fang, Y.~Shan, B.~Feng, and W.~Liu, ``Instances as queries,'' in \emph{Proceedings of the IEEE/CVF international conference on computer vision}, 2021, pp. 6910--6919.

\bibitem{bescos2018dynaslam}
B.~Bescos, J.~M. F{\'a}cil, J.~Civera, and J.~Neira, ``Dynaslam: Tracking, mapping, and inpainting in dynamic scenes,'' \emph{IEEE robotics and automation letters}, vol.~3, no.~4, pp. 4076--4083, 2018.

\bibitem{xu2019mid}
B.~Xu, W.~Li, D.~Tzoumanikas, M.~Bloesch, A.~Davison, and S.~Leutenegger, ``Mid-fusion: Octree-based object-level multi-instance dynamic slam,'' in \emph{2019 International Conference on Robotics and Automation (ICRA)}.\hskip 1em plus 0.5em minus 0.4em\relax IEEE, 2019, pp. 5231--5237.

\bibitem{oleynikova2017voxblox}
H.~Oleynikova, Z.~Taylor, M.~Fehr, R.~Siegwart, and J.~Nieto, ``Voxblox: Incremental 3d euclidean signed distance fields for on-board mav planning,'' in \emph{2017 IEEE/RSJ International Conference on Intelligent Robots and Systems (IROS)}.\hskip 1em plus 0.5em minus 0.4em\relax IEEE, 2017, pp. 1366--1373.

\bibitem{jiang2020pointgroup}
L.~Jiang, H.~Zhao, S.~Shi, S.~Liu, C.-W. Fu, and J.~Jia, ``Pointgroup: Dual-set point grouping for 3d instance segmentation,'' in \emph{Proceedings of the IEEE/CVF conference on computer vision and Pattern recognition}, 2020, pp. 4867--4876.

\bibitem{hou20193d}
J.~Hou, A.~Dai, and M.~Nie{\ss}ner, ``3d-sis: 3d semantic instance segmentation of rgb-d scans,'' in \emph{Proceedings of the IEEE/CVF conference on computer vision and pattern recognition}, 2019, pp. 4421--4430.

\bibitem{stuckler2012semantic}
J.~St{\"u}ckler, N.~Biresev, and S.~Behnke, ``Semantic mapping using object-class segmentation of rgb-d images,'' in \emph{2012 IEEE/RSJ International Conference on Intelligent Robots and Systems}.\hskip 1em plus 0.5em minus 0.4em\relax IEEE, 2012, pp. 3005--3010.

\bibitem{shi2021rgb}
W.~Shi, J.~Xu, D.~Zhu, G.~Zhang, X.~Wang, J.~Li, and X.~Zhang, ``Rgb-d semantic segmentation and label-oriented voxelgrid fusion for accurate 3d semantic mapping,'' \emph{IEEE transactions on circuits and systems for video technology}, vol.~32, no.~1, pp. 183--197, 2021.

\bibitem{kirillov2019panoptic}
A.~Kirillov, K.~He, R.~Girshick, C.~Rother, and P.~Doll{\'a}r, ``Panoptic segmentation,'' in \emph{Proceedings of the IEEE/CVF conference on computer vision and pattern recognition}, 2019, pp. 9404--9413.

\bibitem{mohan2021efficientps}
R.~Mohan and A.~Valada, ``Efficientps: Efficient panoptic segmentation,'' \emph{International Journal of Computer Vision}, vol. 129, no.~5, pp. 1551--1579, 2021.

\bibitem{de2020fast}
D.~De~Geus, P.~Meletis, and G.~Dubbelman, ``Fast panoptic segmentation network,'' \emph{IEEE Robotics and Automation Letters}, vol.~5, no.~2, pp. 1742--1749, 2020.

\bibitem{abati2024panoptic}
G.~F. Abati, J.~C.~V. Soares, V.~S. Medeiros, M.~A. Meggiolaro, and C.~Semini, ``Panoptic-slam: Visual slam in dynamic environments using panoptic segmentation,'' in \emph{2024 21st International Conference on Ubiquitous Robots (UR)}.\hskip 1em plus 0.5em minus 0.4em\relax IEEE, 2024, pp. 01--08.

\bibitem{li2025ps}
G.~Li, J.~Cai, C.~Huang, H.~Luo, and J.~Yu, ``Ps-slam: A visual slam for semantic mapping in dynamic outdoor environment using panoptic segmentation,'' \emph{IEEE Access}, 2025.

\bibitem{sunderhauf2016place}
N.~S{\"u}nderhauf, F.~Dayoub, S.~McMahon, B.~Talbot, R.~Schulz, P.~Corke, G.~Wyeth, B.~Upcroft, and M.~Milford, ``Place categorization and semantic mapping on a mobile robot,'' in \emph{2016 IEEE international conference on robotics and automation (ICRA)}.\hskip 1em plus 0.5em minus 0.4em\relax IEEE, 2016, pp. 5729--5736.

\bibitem{garg2017improving}
S.~Garg, A.~Jacobson, S.~Kumar, and M.~Milford, ``Improving condition-and environment-invariant place recognition with semantic place categorization,'' in \emph{2017 IEEE/RSJ International Conference on Intelligent Robots and Systems (IROS)}.\hskip 1em plus 0.5em minus 0.4em\relax IEEE, 2017, pp. 6863--6870.

\bibitem{arshad2023visem}
S.~Arshad and T.-H. Park, ``Visem: A visual and semantic information fusion based place recognition for long term autonomous navigation,'' in \emph{2023 IEEE 26th International Conference on Intelligent Transportation Systems (ITSC)}.\hskip 1em plus 0.5em minus 0.4em\relax IEEE, 2023, pp. 2151--2156.

\bibitem{woo2024context}
S.~Woo and S.-W. Kim, ``Context-based visual-language place recognition,'' \emph{arXiv preprint arXiv:2410.19341}, 2024.

\bibitem{krizhevsky2012imagenet}
A.~Krizhevsky, I.~Sutskever, and G.~E. Hinton, ``Imagenet classification with deep convolutional neural networks,'' \emph{Advances in neural information processing systems}, vol.~25, 2012.

\bibitem{simonyan2014very}
K.~Simonyan and A.~Zisserman, ``Very deep convolutional networks for large-scale image recognition,'' \emph{arXiv preprint arXiv:1409.1556}, 2014.

\bibitem{szegedy2015going}
C.~Szegedy, W.~Liu, Y.~Jia, P.~Sermanet, S.~Reed, D.~Anguelov, D.~Erhan, V.~Vanhoucke, and A.~Rabinovich, ``Going deeper with convolutions,'' in \emph{Proceedings of the IEEE conference on computer vision and pattern recognition}, 2015, pp. 1--9.

\bibitem{he2016deep}
K.~He, X.~Zhang, S.~Ren, and J.~Sun, ``Deep residual learning for image recognition,'' in \emph{Proceedings of the IEEE conference on computer vision and pattern recognition}, 2016, pp. 770--778.

\bibitem{deng2009imagenet}
J.~Deng, W.~Dong, R.~Socher, L.-J. Li, K.~Li, and L.~Fei-Fei, ``Imagenet: A large-scale hierarchical image database,'' in \emph{2009 IEEE conference on computer vision and pattern recognition}.\hskip 1em plus 0.5em minus 0.4em\relax Ieee, 2009, pp. 248--255.

\bibitem{zhou2017places}
B.~Zhou, A.~Lapedriza, A.~Khosla, A.~Oliva, and A.~Torralba, ``Places: A 10 million image database for scene recognition,'' \emph{IEEE transactions on pattern analysis and machine intelligence}, vol.~40, no.~6, pp. 1452--1464, 2017.

\bibitem{patterson2014sun}
G.~Patterson, C.~Xu, H.~Su, and J.~Hays, ``The sun attribute database: Beyond categories for deeper scene understanding,'' \emph{International Journal of Computer Vision}, vol. 108, pp. 59--81, 2014.

\bibitem{arandjelovic2016netvlad}
R.~Arandjelovic, P.~Gronat, A.~Torii, T.~Pajdla, and J.~Sivic, ``Netvlad: Cnn architecture for weakly supervised place recognition,'' in \emph{Proceedings of the IEEE conference on computer vision and pattern recognition}, 2016, pp. 5297--5307.

\bibitem{dosovitskiy2020image}
A.~Dosovitskiy, L.~Beyer, A.~Kolesnikov, D.~Weissenborn, X.~Zhai, T.~Unterthiner, M.~Dehghani, M.~Minderer, G.~Heigold, S.~Gelly \emph{et~al.}, ``An image is worth 16x16 words: Transformers for image recognition at scale,'' \emph{arXiv preprint arXiv:2010.11929}, 2020.

\bibitem{wang2022transvpr}
R.~Wang, Y.~Shen, W.~Zuo, S.~Zhou, and N.~Zheng, ``Transvpr: Transformer-based place recognition with multi-level attention aggregation,'' in \emph{Proceedings of the IEEE/CVF Conference on Computer Vision and Pattern Recognition}, 2022, pp. 13\,648--13\,657.

\bibitem{kostavelis2017semantic}
I.~Kostavelis and A.~Gasteratos, ``Semantic maps from multiple visual cues,'' \emph{Expert Systems with Applications}, vol.~68, pp. 45--57, 2017.

\bibitem{lin2018topological}
H.-Y. Lin, C.-W. Yao, K.-S. Cheng, and V.~L. Tran, ``Topological map construction and scene recognition for vehicle localization,'' \emph{Autonomous Robots}, vol.~42, pp. 65--81, 2018.

\bibitem{zimmerman2022long}
N.~Zimmerman, T.~Guadagnino, X.~Chen, J.~Behley, and C.~Stachniss, ``Long-term localization using semantic cues in floor plan maps,'' \emph{IEEE Robotics and Automation Letters}, vol.~8, no.~1, pp. 176--183, 2022.

\bibitem{suomela2024placenav}
L.~Suomela, J.~Kalliola, H.~Edelman, and J.-K. K{\"a}m{\"a}r{\"a}inen, ``Placenav: Topological navigation through place recognition,'' in \emph{2024 IEEE International Conference on Robotics and Automation (ICRA)}.\hskip 1em plus 0.5em minus 0.4em\relax IEEE, 2024, pp. 5205--5213.

\bibitem{muravyev2025prism}
K.~Muravyev, A.~Melekhin, D.~Yudin, and K.~Yakovlev, ``Prism-topomap: online topological mapping with place recognition and scan matching,'' \emph{IEEE Robotics and Automation Letters}, 2025.

\bibitem{long2020pspnet}
X.~Long, W.~Zhang, and B.~Zhao, ``Pspnet-slam: A semantic slam detect dynamic object by pyramid scene parsing network,'' \emph{IEEE access}, vol.~8, pp. 214\,685--214\,695, 2020.

\bibitem{ji2024drv}
Q.~Ji, Z.~Zhang, Y.~Chen, and E.~Zheng, ``Drv-slam: An adaptive real-time semantic visual slam based on instance segmentation toward dynamic environments,'' \emph{Ieee Access}, vol.~12, pp. 43\,827--43\,837, 2024.

\bibitem{zhou2019semantic}
H.~Zhou, J.~Fang, L.~Zhang, Y.~Xu, and Y.~Liu, ``Semantic scene completion from a single depth image,'' in \emph{Proceedings of the IEEE conference on computer vision and pattern recognition}, 2019, pp. 1746--1754.

\bibitem{civera2011towards}
J.~Civera, D.~G{\'a}lvez-L{\'o}pez, L.~Riazuelo, J.~D. Tard{\'o}s, and J.~M.~M. Montiel, ``Towards semantic slam using a monocular camera,'' in \emph{2011 IEEE/RSJ international conference on intelligent robots and systems}.\hskip 1em plus 0.5em minus 0.4em\relax IEEE, 2011, pp. 1277--1284.

\bibitem{vasudevan2008bayesian}
S.~Vasudevan and R.~Siegwart, ``Bayesian space conceptualization and place classification for semantic maps in mobile robotics,'' \emph{Robotics and Autonomous Systems}, vol.~56, no.~6, pp. 522--537, 2008.

\bibitem{kong2020semantic}
X.~Kong, X.~Yang, G.~Zhai, X.~Zhao, X.~Zeng, M.~Wang, Y.~Liu, W.~Li, and F.~Wen, ``Semantic graph based place recognition for 3d point clouds,'' in \emph{2020 IEEE/RSJ International Conference on Intelligent Robots and Systems (IROS)}.\hskip 1em plus 0.5em minus 0.4em\relax IEEE, 2020, pp. 8216--8223.

\bibitem{singh2024open}
K.~Singh and J.~J. Leonard, ``Open-set semantic uncertainty aware metric-semantic graph matching,'' \emph{arXiv preprint arXiv:2409.11555}, 2024.

\bibitem{tennakoon2023factor}
K.~B. Tennakoon, O.~De~Silva, A.~Jayasiri, G.~K. Mann, and R.~G. Gosine, ``Factor graph localization for mobile robots using google indoor street view and cnn-based place recognition,'' \emph{Drone Systems and Applications}, vol.~11, pp. 1--19, 2023.

\bibitem{patel2018semantic}
N.~Patel, P.~Krishnamurthy, and F.~Khorrami, ``Semantic segmentation guided slam using vision and lidar,'' in \emph{ISR 2018; 50th International Symposium on Robotics}.\hskip 1em plus 0.5em minus 0.4em\relax VDE, 2018, pp. 1--7.

\bibitem{nakajima2018fast}
Y.~Nakajima, K.~Tateno, F.~Tombari, and H.~Saito, ``Fast and accurate semantic mapping through geometric-based incremental segmentation,'' in \emph{2018 IEEE/RSJ International Conference on Intelligent Robots and Systems (IROS)}.\hskip 1em plus 0.5em minus 0.4em\relax IEEE, 2018, pp. 385--392.

\bibitem{du2020accurate}
Z.-J. Du, S.-S. Huang, T.-J. Mu, Q.~Zhao, R.~R. Martin, and K.~Xu, ``Accurate dynamic slam using crf-based long-term consistency,'' \emph{IEEE Transactions on Visualization and Computer Graphics}, vol.~28, no.~4, pp. 1745--1757, 2020.

\bibitem{jeon2022rgb}
H.~Jeon, C.~Han, D.~You, and J.~Oh, ``Rgb-d visual slam algorithm using scene flow and conditional random field in dynamic environments,'' in \emph{2022 22nd International Conference on Control, Automation and Systems (ICCAS)}.\hskip 1em plus 0.5em minus 0.4em\relax IEEE, 2022, pp. 129--134.

\bibitem{pronobis2017learning}
A.~Pronobis and R.~P. Rao, ``Learning deep generative spatial models for mobile robots,'' in \emph{2017 IEEE/RSJ International Conference on Intelligent Robots and Systems (IROS)}.\hskip 1em plus 0.5em minus 0.4em\relax IEEE, 2017, pp. 755--762.

\bibitem{zhang2024active}
R.~Zhang, H.~M. Bong, and G.~Beltrame, ``Active semantic mapping and pose graph spectral analysis for robot exploration,'' in \emph{2024 IEEE/RSJ International Conference on Intelligent Robots and Systems (IROS)}.\hskip 1em plus 0.5em minus 0.4em\relax IEEE, 2024, pp. 13\,787--13\,794.

\bibitem{bavle2020vps}
H.~Bavle, P.~De~La~Puente, J.~P. How, and P.~Campoy, ``Vps-slam: Visual planar semantic slam for aerial robotic systems,'' \emph{IEEE Access}, vol.~8, pp. 60\,704--60\,718, 2020.

\bibitem{qin2020avp}
T.~Qin, T.~Chen, Y.~Chen, and Q.~Su, ``Avp-slam: Semantic visual mapping and localization for autonomous vehicles in the parking lot,'' in \emph{2020 IEEE/RSJ International Conference on intelligent robots and systems (IROS)}.\hskip 1em plus 0.5em minus 0.4em\relax IEEE, 2020, pp. 5939--5945.

\bibitem{miller2022robust}
I.~D. Miller, R.~Soussan, B.~Coltin, T.~Smith, and V.~Kumar, ``Robust semantic mapping and localization on a free-flying robot in microgravity,'' in \emph{2022 International Conference on Robotics and Automation (ICRA)}.\hskip 1em plus 0.5em minus 0.4em\relax IEEE, 2022, pp. 4121--4127.

\bibitem{liu2019visual}
H.~Liu, H.~Ma, and L.~Zhang, ``Visual odometry based on semantic supervision,'' in \emph{2019 IEEE International Conference on Image Processing (ICIP)}.\hskip 1em plus 0.5em minus 0.4em\relax IEEE, 2019, pp. 2566--2570.

\bibitem{cui2019sof}
L.~Cui and C.~Ma, ``Sof-slam: A semantic visual slam for dynamic environments,'' \emph{IEEE access}, vol.~7, pp. 166\,528--166\,539, 2019.

\bibitem{an2017semantic}
L.~An, X.~Zhang, H.~Gao, and Y.~Liu, ``Semantic segmentation--aided visual odometry for urban autonomous driving,'' \emph{International Journal of Advanced Robotic Systems}, vol.~14, no.~5, p. 1729881417735667, 2017.

\bibitem{yuan2020sad}
X.~Yuan and S.~Chen, ``Sad-slam: A visual slam based on semantic and depth information,'' in \emph{2020 IEEE/RSJ International Conference on Intelligent Robots and Systems (IROS)}.\hskip 1em plus 0.5em minus 0.4em\relax IEEE, 2020, pp. 4930--4935.

\bibitem{yan2022dgs}
L.~Yan, X.~Hu, L.~Zhao, Y.~Chen, P.~Wei, and H.~Xie, ``Dgs-slam: A fast and robust rgbd slam in dynamic environments combined by geometric and semantic information,'' \emph{Remote Sensing}, vol.~14, no.~3, p. 795, 2022.

\bibitem{fan2022blitz}
Y.~Fan, Q.~Zhang, Y.~Tang, S.~Liu, and H.~Han, ``Blitz-slam: A semantic slam in dynamic environments,'' \emph{Pattern Recognition}, vol. 121, p. 108225, 2022.

\bibitem{sheng2020dynamic}
C.~Sheng, S.~Pan, W.~Gao, Y.~Tan, and T.~Zhao, ``Dynamic-dso: direct sparse odometry using objects semantic information for dynamic environments,'' \emph{Applied sciences}, vol.~10, no.~4, p. 1467, 2020.

\bibitem{li2022dystslam}
X.~Li, Y.~Shen, J.~Lu, Q.~Jiang, O.~Xie, Y.~Yang, and Q.~Zhu, ``Dystslam: an efficient stereo vision slam system in dynamic environment,'' \emph{Measurement Science and Technology}, vol.~34, no.~2, p. 025105, 2022.

\bibitem{zhong2022wf}
Y.~Zhong, S.~Hu, G.~Huang, L.~Bai, and Q.~Li, ``Wf-slam: A robust vslam for dynamic scenarios via weighted features,'' \emph{IEEE Sensors Journal}, vol.~22, no.~11, pp. 10\,818--10\,827, 2022.

\bibitem{zhang2021pld}
C.~Zhang, T.~Huang, R.~Zhang, and X.~Yi, ``Pld-slam: A new rgb-d slam method with point and line features for indoor dynamic scene,'' \emph{ISPRS International Journal of Geo-Information}, vol.~10, no.~3, p. 163, 2021.

\bibitem{wang2022ypd}
Y.~Wang, H.~Bu, X.~Zhang, and J.~Cheng, ``Ypd-slam: A real-time vslam system for handling dynamic indoor environments,'' \emph{Sensors}, vol.~22, no.~21, p. 8561, 2022.

\bibitem{yuan2023plds}
C.~Yuan, Y.~Xu, and Q.~Zhou, ``Plds-slam: point and line features slam in dynamic environment,'' \emph{Remote sensing}, vol.~15, no.~7, p. 1893, 2023.

\bibitem{lu2020dm}
X.~Lu, H.~Wang, S.~Tang, H.~Huang, and C.~Li, ``Dm-slam: Monocular slam in dynamic environments,'' \emph{Applied Sciences}, vol.~10, no.~12, p. 4252, 2020.

\bibitem{liu2021rds}
Y.~Liu and J.~Miura, ``Rds-slam: Real-time dynamic slam using semantic segmentation methods,'' \emph{Ieee Access}, vol.~9, pp. 23\,772--23\,785, 2021.

\bibitem{liu2021rdmo}
------, ``Rdmo-slam: Real-time visual slam for dynamic environments using semantic label prediction with optical flow,'' \emph{Ieee Access}, vol.~9, pp. 106\,981--106\,997, 2021.

\bibitem{ji2021towards}
T.~Ji, C.~Wang, and L.~Xie, ``Towards real-time semantic rgb-d slam in dynamic environments,'' in \emph{2021 IEEE international conference on robotics and automation (ICRA)}.\hskip 1em plus 0.5em minus 0.4em\relax IEEE, 2021, pp. 11\,175--11\,181.

\bibitem{hu2022cfp}
X.~Hu, Y.~Zhang, Z.~Cao, R.~Ma, Y.~Wu, Z.~Deng, and W.~Sun, ``Cfp-slam: A real-time visual slam based on coarse-to-fine probability in dynamic environments,'' in \emph{2022 IEEE/RSJ International Conference on Intelligent Robots and Systems (IROS)}.\hskip 1em plus 0.5em minus 0.4em\relax IEEE, 2022, pp. 4399--4406.

\bibitem{wei2023slam}
Y.~Wei, B.~Zhou, Y.~Duan, J.~Liu, and D.~An, ``Do-slam: research and application of semantic slam system towards dynamic environments based on object detection,'' \emph{Applied Intelligence}, vol.~53, no.~24, pp. 30\,009--30\,026, 2023.

\bibitem{he2023ovd}
J.~He, M.~Li, Y.~Wang, and H.~Wang, ``Ovd-slam: An online visual slam for dynamic environments,'' \emph{IEEE Sensors Journal}, vol.~23, no.~12, pp. 13\,210--13\,219, 2023.

\bibitem{islam2024mvs}
Q.~U. Islam, H.~Ibrahim, P.~K. Chin, K.~Lim, and M.~Z. Abdullah, ``Mvs-slam: Enhanced multiview geometry for improved semantic rgbd slam in dynamic environment,'' \emph{Journal of Field Robotics}, vol.~41, no.~1, pp. 109--130, 2024.

\bibitem{zhong2018detect}
F.~Zhong, S.~Wang, Z.~Zhang, and Y.~Wang, ``Detect-slam: Making object detection and slam mutually beneficial,'' in \emph{2018 IEEE winter conference on applications of computer vision (WACV)}.\hskip 1em plus 0.5em minus 0.4em\relax IEEE, 2018, pp. 1001--1010.

\bibitem{contreras2024dynanav}
M.~Contreras, N.~P. Bhatt, and E.~Hashemi, ``Dynanav-svo: Dynamic stereo visual odometry with semantic-aware perception for autonomous navigation,'' \emph{IEEE Transactions on Intelligent Vehicles}, 2024.

\bibitem{cheng2022sg}
S.~Cheng, C.~Sun, S.~Zhang, and D.~Zhang, ``Sg-slam: A real-time rgb-d visual slam toward dynamic scenes with semantic and geometric information,'' \emph{IEEE Transactions on Instrumentation and Measurement}, vol.~72, pp. 1--12, 2022.

\bibitem{zhang2025adaptive}
H.~Zhang, T.~N. Canh, C.~Li, and N.~Y. Chong, ``Adaptive prior scene-object slam for dynamic environments,'' in \emph{2025 IEEE International Conference on Real-time Computing and Robotics (RCAR)}.\hskip 1em plus 0.5em minus 0.4em\relax IEEE, 2025, pp. 1--6.

\bibitem{qian2023visual}
R.~Qian, H.~Guo, M.~Chen, G.~Gong, and H.~Cheng, ``A visual slam algorithm based on instance segmentation and background inpainting in dynamic scenes,'' in \emph{2023 38th Youth Academic Annual Conference of Chinese Association of Automation (YAC)}.\hskip 1em plus 0.5em minus 0.4em\relax IEEE, 2023, pp. 132--136.

\bibitem{bescos2020empty}
B.~Bescos, C.~Cadena, and J.~Neira, ``Empty cities: A dynamic-object-invariant space for visual slam,'' \emph{IEEE Transactions on Robotics}, vol.~37, no.~2, pp. 433--451, 2020.

\bibitem{goodfellow2020generative}
I.~Goodfellow, J.~Pouget-Abadie, M.~Mirza, B.~Xu, D.~Warde-Farley, S.~Ozair, A.~Courville, and Y.~Bengio, ``Generative adversarial networks,'' \emph{Communications of the ACM}, vol.~63, no.~11, pp. 139--144, 2020.

\bibitem{bescos2021dynaslam}
B.~Bescos, C.~Campos, J.~D. Tard{\'o}s, and J.~Neira, ``Dynaslam ii: Tightly-coupled multi-object tracking and slam,'' \emph{IEEE robotics and automation letters}, vol.~6, no.~3, pp. 5191--5198, 2021.

\bibitem{ballester2021dot}
I.~Ballester, A.~Font{\'a}n, J.~Civera, K.~H. Strobl, and R.~Triebel, ``Dot: Dynamic object tracking for visual slam,'' in \emph{2021 IEEE international conference on robotics and automation (ICRA)}.\hskip 1em plus 0.5em minus 0.4em\relax IEEE, 2021, pp. 11\,705--11\,711.

\bibitem{zhou2023pointslot}
P.~Zhou, Y.~Liu, and Z.~Meng, ``Pointslot: Real-time simultaneous localization and object tracking for dynamic environment,'' \emph{IEEE Robotics and Automation Letters}, vol.~8, no.~5, pp. 2645--2652, 2023.

\bibitem{zhang2020robust}
J.~Zhang, M.~Henein, R.~Mahony, and V.~Ila, ``Robust ego and object 6-dof motion estimation and tracking,'' in \emph{2020 IEEE/RSJ International Conference on Intelligent Robots and Systems (IROS)}.\hskip 1em plus 0.5em minus 0.4em\relax IEEE, 2020, pp. 5017--5023.

\bibitem{zhang2022motslam}
H.~Zhang, H.~Uchiyama, S.~Ono, and H.~Kawasaki, ``Motslam: Mot-assisted monocular dynamic slam using single-view depth estimation,'' in \emph{2022 IEEE/RSJ International Conference on Intelligent Robots and Systems (IROS)}.\hskip 1em plus 0.5em minus 0.4em\relax IEEE, 2022, pp. 4865--4872.

\bibitem{xing2022slam}
Z.~Xing, X.~Zhu, and D.~Dong, ``De-slam: Slam for highly dynamic environment,'' \emph{Journal of Field Robotics}, vol.~39, no.~5, pp. 528--542, 2022.

\bibitem{gonzalez2022twistslam}
M.~Gonzalez, E.~Marchand, A.~Kacete, and J.~Royan, ``Twistslam: Constrained slam in dynamic environment,'' \emph{IEEE Robotics and Automation Letters}, vol.~7, no.~3, pp. 6846--6853, 2022.

\bibitem{schorghuber2019slamantic}
M.~Schorghuber, D.~Steininger, Y.~Cabon, M.~Humenberger, and M.~Gelautz, ``Slamantic-leveraging semantics to improve vslam in dynamic environments,'' in \emph{Proceedings of the IEEE/CVF International Conference on Computer Vision Workshops}, 2019, pp. 0--0.

\bibitem{zhang2020vdo}
J.~Zhang, M.~Henein, R.~Mahony, and V.~Ila, ``Vdo-slam: A visual dynamic object-aware slam system,'' \emph{arXiv preprint arXiv:2005.11052}, 2020.

\bibitem{huang2020clustervo}
J.~Huang, S.~Yang, T.-J. Mu, and S.-M. Hu, ``Clustervo: Clustering moving instances and estimating visual odometry for self and surroundings,'' in \emph{Proceedings of the IEEE/CVF conference on computer vision and pattern recognition}, 2020, pp. 2168--2177.

\bibitem{chang2023ote}
Y.~Chang, J.~Hu, and S.~Xu, ``Ote-slam: An object tracking enhanced visual slam system for dynamic environments,'' \emph{Sensors}, vol.~23, no.~18, p. 7921, 2023.

\bibitem{zhang2022bytetrack}
Y.~Zhang, P.~Sun, Y.~Jiang, D.~Yu, F.~Weng, Z.~Yuan, P.~Luo, W.~Liu, and X.~Wang, ``Bytetrack: Multi-object tracking by associating every detection box,'' in \emph{European conference on computer vision}.\hskip 1em plus 0.5em minus 0.4em\relax Springer, 2022, pp. 1--21.

\bibitem{mascaro2024scene}
R.~Mascaro and M.~Chli, ``Scene representations for robotic spatial perception,'' \emph{Annual Review of Control, Robotics, and Autonomous Systems}, vol.~8, 2024.

\bibitem{sharma2021compositional}
A.~Sharma, W.~Dong, and M.~Kaess, ``Compositional and scalable object slam,'' in \emph{2021 IEEE International Conference on Robotics and Automation (ICRA)}.\hskip 1em plus 0.5em minus 0.4em\relax IEEE, 2021, pp. 11\,626--11\,632.

\bibitem{hosseinzadeh2019real}
M.~Hosseinzadeh, K.~Li, Y.~Latif, and I.~Reid, ``Real-time monocular object-model aware sparse slam,'' in \emph{2019 international conference on robotics and automation (ICRA)}.\hskip 1em plus 0.5em minus 0.4em\relax IEEE, 2019, pp. 7123--7129.

\bibitem{chen2022accurate}
K.~Chen, J.~Liu, Q.~Chen, Z.~Wang, and J.~Zhang, ``Accurate object association and pose updating for semantic slam,'' \emph{IEEE Transactions on Intelligent Transportation Systems}, vol.~23, no.~12, pp. 25\,169--25\,179, 2022.

\bibitem{kochanov2016scene}
D.~Kochanov, A.~O{\v{s}}ep, J.~St{\"u}ckler, and B.~Leibe, ``Scene flow propagation for semantic mapping and object discovery in dynamic street scenes,'' in \emph{2016 IEEE/RSJ International Conference on Intelligent Robots and Systems (IROS)}.\hskip 1em plus 0.5em minus 0.4em\relax IEEE, 2016, pp. 1785--1792.

\bibitem{mascaro2022volumetric}
R.~Mascaro, L.~Teixeira, and M.~Chli, ``Volumetric instance-level semantic mapping via multi-view 2d-to-3d label diffusion,'' \emph{IEEE Robotics and Automation Letters}, vol.~7, no.~2, pp. 3531--3538, 2022.

\bibitem{hu2023multi}
X.~Hu, ``Multi-level map construction for dynamic scenes,'' \emph{arXiv preprint arXiv:2308.04000}, 2023.

\bibitem{li2020textslam}
B.~Li, D.~Zou, D.~Sartori, L.~Pei, and W.~Yu, ``Textslam: Visual slam with planar text features,'' in \emph{2020 IEEE International Conference on Robotics and Automation (ICRA)}.\hskip 1em plus 0.5em minus 0.4em\relax IEEE, 2020, pp. 2102--2108.

\bibitem{grinvald2021tsdf++}
M.~Grinvald, F.~Tombari, R.~Siegwart, and J.~Nieto, ``Tsdf++: A multi-object formulation for dynamic object tracking and reconstruction,'' in \emph{2021 IEEE international conference on robotics and automation (ICRA)}.\hskip 1em plus 0.5em minus 0.4em\relax IEEE, 2021, pp. 14\,192--14\,198.

\bibitem{asgharivaskasi2023semantic}
A.~Asgharivaskasi and N.~Atanasov, ``Semantic octree mapping and shannon mutual information computation for robot exploration,'' \emph{IEEE Transactions on Robotics}, vol.~39, no.~3, pp. 1910--1928, 2023.

\bibitem{sun2022volumetric}
G.~Sun, X.~Zhang, Y.~Chu, Y.~Liu, X.~Zhang, and Y.~Zhuang, ``Volumetric instance-level semantic mapping via blendmask,'' in \emph{2022 IEEE/ASME International Conference on Advanced Intelligent Mechatronics (AIM)}.\hskip 1em plus 0.5em minus 0.4em\relax IEEE, 2022, pp. 374--379.

\bibitem{ran2021rs}
T.~Ran, L.~Yuan, J.~Zhang, D.~Tang, and L.~He, ``Rs-slam: A robust semantic slam in dynamic environments based on rgb-d sensor,'' \emph{IEEE Sensors Journal}, vol.~21, no.~18, pp. 20\,657--20\,664, 2021.

\bibitem{barsan2018robust}
I.~A. B{\^a}rsan, P.~Liu, M.~Pollefeys, and A.~Geiger, ``Robust dense mapping for large-scale dynamic environments,'' in \emph{2018 IEEE International Conference on Robotics and Automation (ICRA)}.\hskip 1em plus 0.5em minus 0.4em\relax IEEE, 2018, pp. 7510--7517.

\bibitem{li2018dense}
L.~Li, Z.~Liu, {\"U}.~{\"O}zg{\"\i}ner, J.~Lian, Y.~Zhou, and Y.~Zhao, ``Dense 3d semantic slam of traffic environment based on stereo vision,'' in \emph{2018 IEEE Intelligent Vehicles Symposium (IV)}.\hskip 1em plus 0.5em minus 0.4em\relax IEEE, 2018, pp. 965--970.

\bibitem{stuckler2015dense}
J.~St{\"u}ckler, B.~Waldvogel, H.~Schulz, and S.~Behnke, ``Dense real-time mapping of object-class semantics from rgb-d video,'' \emph{Journal of real-time image processing}, vol.~10, pp. 599--609, 2015.

\bibitem{reddy2015dynamic}
N.~D. Reddy, P.~Singhal, V.~Chari, and K.~M. Krishna, ``Dynamic body vslam with semantic constraints,'' in \emph{2015 IEEE/RSJ International Conference on Intelligent Robots and Systems (IROS)}.\hskip 1em plus 0.5em minus 0.4em\relax IEEE, 2015, pp. 1897--1904.

\bibitem{morreale2019dense}
L.~Morreale, A.~Romanoni, M.~Matteucci, and P.~di~Milano, ``Dense 3d visual mapping via semantic simplification,'' in \emph{2019 International Conference on Robotics and Automation (ICRA)}.\hskip 1em plus 0.5em minus 0.4em\relax IEEE, 2019, pp. 6891--6897.

\bibitem{seichter2022efficient}
D.~Seichter, S.~B. Fischedick, M.~K{\"o}hler, and H.-M. Gro{\ss}, ``Efficient multi-task rgb-d scene analysis for indoor environments,'' in \emph{2022 International joint conference on neural networks (IJCNN)}.\hskip 1em plus 0.5em minus 0.4em\relax IEEE, 2022, pp. 1--10.

\bibitem{prisacariu2017infinitam}
V.~A. Prisacariu, O.~K{\"a}hler, S.~Golodetz, M.~Sapienza, T.~Cavallari, P.~H. Torr, and D.~W. Murray, ``Infinitam v3: A framework for large-scale 3d reconstruction with loop closure,'' \emph{arXiv preprint arXiv:1708.00783}, 2017.

\bibitem{pham2019real}
Q.-H. Pham, B.-S. Hua, T.~Nguyen, and S.-K. Yeung, ``Real-time progressive 3d semantic segmentation for indoor scenes,'' in \emph{2019 IEEE Winter Conference on Applications of Computer Vision (WACV)}.\hskip 1em plus 0.5em minus 0.4em\relax IEEE, 2019, pp. 1089--1098.

\bibitem{yang2021tupper}
Z.~Yang and C.~Liu, ``Tupper-map: Temporal and unified panoptic perception for 3d metric-semantic mapping,'' in \emph{2021 IEEE/RSJ International Conference on Intelligent Robots and Systems (IROS)}.\hskip 1em plus 0.5em minus 0.4em\relax IEEE, 2021, pp. 1094--1101.

\bibitem{xiong2019upsnet}
Y.~Xiong, R.~Liao, H.~Zhao, R.~Hu, M.~Bai, E.~Yumer, and R.~Urtasun, ``Upsnet: A unified panoptic segmentation network,'' in \emph{Proceedings of the IEEE/CVF conference on computer vision and pattern recognition}, 2019, pp. 8818--8826.

\bibitem{kim2020video}
D.~Kim, S.~Woo, J.-Y. Lee, and I.~S. Kweon, ``Video panoptic segmentation,'' in \emph{Proceedings of the IEEE/CVF conference on computer vision and pattern recognition}, 2020, pp. 9859--9868.

\bibitem{schmid2022panoptic}
L.~Schmid, J.~Delmerico, J.~L. Sch{\"o}nberger, J.~Nieto, M.~Pollefeys, R.~Siegwart, and C.~Cadena, ``Panoptic multi-tsdfs: a flexible representation for online multi-resolution volumetric mapping and long-term dynamic scene consistency,'' in \emph{2022 International Conference on Robotics and Automation (ICRA)}.\hskip 1em plus 0.5em minus 0.4em\relax IEEE, 2022, pp. 8018--8024.

\bibitem{seichter2023panopticndt}
D.~Seichter, B.~Stephan, S.~B. Fischedick, S.~Mueller, L.~Rabes, and H.-M. Gross, ``Panopticndt: Efficient and robust panoptic mapping,'' in \emph{2023 IEEE/RSJ International Conference on Intelligent Robots and Systems (IROS)}.\hskip 1em plus 0.5em minus 0.4em\relax IEEE, 2023, pp. 7233--7240.

\bibitem{wu2024panorecon}
D.~Wu, Z.~Yan, and H.~Zha, ``Panorecon: Real-time panoptic 3d reconstruction from monocular video,'' in \emph{Proceedings of the IEEE/CVF Conference on Computer Vision and Pattern Recognition}, 2024, pp. 21\,507--21\,518.

\bibitem{zhou2024eprecon}
Z.~Zhou, Y.~Ma, J.~Fan, S.~Zhang, F.~Jing, and M.~Tan, ``Eprecon: An efficient framework for real-time panoptic 3d reconstruction from monocular video,'' in \emph{2022 International Conference on Robotics and Automation (ICRA)}.\hskip 1em plus 0.5em minus 0.4em\relax IEEE, 2024.

\bibitem{bernuy2015semantic}
F.~Bernuy and J.~Ruiz~del Solar, ``Semantic mapping of large-scale outdoor scenes for autonomous off-road driving,'' in \emph{Proceedings of the IEEE International Conference on Computer Vision Workshops}, 2015, pp. 35--41.

\bibitem{bernuy2018topological}
F.~Bernuy and J.~Ruiz-del Solar, ``Topological semantic mapping and localization in urban road scenarios,'' \emph{Journal of Intelligent \& Robotic Systems}, vol.~92, pp. 19--32, 2018.

\bibitem{tian2022vision}
W.~Tian, X.~Ren, X.~Yu, M.~Wu, W.~Zhao, and Q.~Li, ``Vision-based mapping of lane semantics and topology for intelligent vehicles,'' \emph{International Journal of Applied Earth Observation and Geoinformation}, vol. 111, p. 102851, 2022.

\bibitem{zhao2018indoor}
L.~Zhao, P.~Luo, Z.~Zhao, and L.~Dong, ``Indoor environment semantic topological mapping based on deep learning,'' in \emph{2018 IEEE International Conference on Real-time Computing and Robotics (RCAR)}.\hskip 1em plus 0.5em minus 0.4em\relax IEEE, 2018, pp. 520--525.

\bibitem{blochliger2018topomap}
F.~Blochliger, M.~Fehr, M.~Dymczyk, T.~Schneider, and R.~Siegwart, ``Topomap: Topological mapping and navigation based on visual slam maps,'' in \emph{2018 IEEE International Conference on Robotics and Automation (ICRA)}.\hskip 1em plus 0.5em minus 0.4em\relax IEEE, 2018, pp. 3818--3825.

\bibitem{chen2021topological}
Y.~Chen, J.~Zhang, and Y.~Lou, ``Topological and semantic map generation for mobile robot indoor navigation,'' in \emph{International Conference on Intelligent Robotics and Applications}.\hskip 1em plus 0.5em minus 0.4em\relax Springer, 2021, pp. 337--347.

\bibitem{kim2023topological}
N.~Kim, O.~Kwon, H.~Yoo, Y.~Choi, J.~Park, and S.~Oh, ``Topological semantic graph memory for image-goal navigation,'' in \emph{Conference on Robot Learning}.\hskip 1em plus 0.5em minus 0.4em\relax PMLR, 2023, pp. 393--402.

\bibitem{armeni20193d}
I.~Armeni, Z.-Y. He, J.~Gwak, A.~R. Zamir, M.~Fischer, J.~Malik, and S.~Savarese, ``3d scene graph: A structure for unified semantics, 3d space, and camera,'' in \emph{Proceedings of the IEEE/CVF international conference on computer vision}, 2019, pp. 5664--5673.

\bibitem{kim20193}
U.-H. Kim, J.-M. Park, T.-J. Song, and J.-H. Kim, ``3-d scene graph: A sparse and semantic representation of physical environments for intelligent agents,'' \emph{IEEE transactions on cybernetics}, vol.~50, no.~12, pp. 4921--4933, 2019.

\bibitem{wu2021scenegraphfusion}
S.-C. Wu, J.~Wald, K.~Tateno, N.~Navab, and F.~Tombari, ``Scenegraphfusion: Incremental 3d scene graph prediction from rgb-d sequences,'' in \emph{Proceedings of the IEEE/CVF Conference on Computer Vision and Pattern Recognition}, 2021, pp. 7515--7525.

\bibitem{mehan2024questmaps}
Y.~Mehan, K.~Gupta, R.~Jayanti, A.~Govil, S.~Garg, and M.~Krishna, ``Questmaps: Queryable semantic topological maps for 3d scene understanding,'' in \emph{2024 IEEE/RSJ International Conference on Intelligent Robots and Systems (IROS)}.\hskip 1em plus 0.5em minus 0.4em\relax IEEE, 2024, pp. 13\,311--13\,317.

\bibitem{sousa2022topological}
Y.~C. Sousa and H.~F. Bassani, ``Topological semantic mapping by consolidation of deep visual features,'' \emph{IEEE Robotics and Automation Letters}, vol.~7, no.~2, pp. 4110--4117, 2022.

\bibitem{yang2022automated}
B.~Yang, T.~Jiang, W.~Wu, Y.~Zhou, and L.~Dai, ``Automated semantics and topology representation of residential-building space using floor-plan raster maps,'' \emph{IEEE Journal of Selected Topics in Applied Earth Observations and Remote Sensing}, vol.~15, pp. 7809--7825, 2022.

\bibitem{cao2024context}
Z.~Cao, Y.~Sun, Z.~Ma, and M.~Zhou, ``A context-enhanced full-resolution floor plan segmentation network for topological semantic mapping,'' in \emph{2024 IEEE/RSJ International Conference on Intelligent Robots and Systems (IROS)}.\hskip 1em plus 0.5em minus 0.4em\relax IEEE, 2024, pp. 9761--9768.

\bibitem{lin2021topology}
S.~Lin, J.~Wang, M.~Xu, H.~Zhao, and Z.~Chen, ``Topology aware object-level semantic mapping towards more robust loop closure,'' \emph{IEEE Robotics and Automation Letters}, vol.~6, no.~4, pp. 7041--7048, 2021.

\bibitem{cao2024semantictopoloop}
Z.~Cao, Q.~Zhang, J.~Guang, S.~Wu, Z.~Hu, and J.~Liu, ``Semantictopoloop: Semantic loop closure with 3d topological graph based on quadric-level object map,'' \emph{IEEE Robotics and Automation Letters}, 2024.

\bibitem{fredriksson2023semantic}
S.~Fredriksson, A.~Saradagi, and G.~Nikolakopoulos, ``Semantic and topological mapping using intersection identification,'' \emph{IFAC-PapersOnLine}, vol.~56, no.~2, pp. 9251--9256, 2023.

\bibitem{kathirvel2025sent}
R.~S.~R. Kathirvel, Z.~A. Chavis, S.~J. Guy, and K.~Desingh, ``Sent map-semantically enhanced topological maps with foundation models,'' in \emph{ICRA 2025 Workshop on Foundation Models and Neuro-Symbolic AI for Robotics}.\hskip 1em plus 0.5em minus 0.4em\relax IEEE, 2025.

\bibitem{tomatis2003hybrid}
N.~Tomatis, I.~Nourbakhsh, and R.~Siegwart, ``Hybrid simultaneous localization and map building: a natural integration of topological and metric,'' \emph{Robotics and Autonomous systems}, vol.~44, no.~1, pp. 3--14, 2003.

\bibitem{kuipers2004local}
B.~Kuipers, J.~Modayil, P.~Beeson, M.~MacMahon, and F.~Savelli, ``Local metrical and global topological maps in the hybrid spatial semantic hierarchy,'' in \emph{IEEE International Conference on Robotics and Automation, 2004. Proceedings. ICRA'04. 2004}, vol.~5.\hskip 1em plus 0.5em minus 0.4em\relax IEEE, 2004, pp. 4845--4851.

\bibitem{drouilly2015hybrid}
R.~Drouilly, P.~Rives, and B.~Morisset, ``Hybrid metric-topological-semantic mapping in dynamic environments,'' in \emph{2015 IEEE/RSJ International Conference on Intelligent Robots and Systems (IROS)}.\hskip 1em plus 0.5em minus 0.4em\relax IEEE, 2015, pp. 5109--5114.

\bibitem{yang2017semantic}
S.~Yang, Y.~Huang, and S.~Scherer, ``Semantic 3d occupancy mapping through efficient high order crfs,'' in \emph{2017 IEEE/RSJ International Conference on Intelligent Robots and Systems (IROS)}.\hskip 1em plus 0.5em minus 0.4em\relax IEEE, 2017, pp. 590--597.

\bibitem{kohli2007p3}
P.~Kohli, M.~P. Kumar, and P.~H. Torr, ``P3 \& beyond: Solving energies with higher order cliques,'' in \emph{2007 IEEE conference on computer vision and pattern recognition}.\hskip 1em plus 0.5em minus 0.4em\relax IEEE, 2007, pp. 1--8.

\bibitem{malleson2018hybrid}
C.~Malleson, J.-Y. Guillemaut, and A.~Hilton, ``Hybrid modeling of non-rigid scenes from rgbd cameras,'' \emph{IEEE Transactions on Circuits and Systems for Video Technology}, vol.~29, no.~8, pp. 2391--2404, 2018.

\bibitem{luo2018hierarchical}
R.~C. Luo and M.~Chiou, ``Hierarchical semantic mapping using convolutional neural networks for intelligent service robotics,'' \emph{IEEE Access}, vol.~6, pp. 61\,287--61\,294, 2018.

\bibitem{wen2020hybrid}
S.~Wen, Y.~Zhao, X.~Liu, F.~Sun, H.~Lu, and Z.~Wang, ``Hybrid semi-dense 3d semantic-topological mapping from stereo visual-inertial odometry slam with loop closure detection,'' \emph{IEEE Transactions on Vehicular Technology}, vol.~69, no.~12, pp. 16\,057--16\,066, 2020.

\bibitem{yue2020hierarchical}
Y.~Yue, C.~Zhao, R.~Li, C.~Yang, J.~Zhang, M.~Wen, Y.~Wang, and D.~Wang, ``A hierarchical framework for collaborative probabilistic semantic mapping,'' in \emph{2020 IEEE international conference on robotics and automation (ICRA)}.\hskip 1em plus 0.5em minus 0.4em\relax IEEE, 2020, pp. 9659--9665.

\bibitem{deng2022hd}
Y.~Deng, M.~Wang, Y.~Yang, and Y.~Yue, ``Hd-ccsom: Hierarchical and dense collaborative continuous semantic occupancy mapping through label diffusion,'' in \emph{2022 IEEE/RSJ international conference on intelligent robots and systems (IROS)}.\hskip 1em plus 0.5em minus 0.4em\relax IEEE, 2022, pp. 2417--2422.

\bibitem{rosinol2021kimera}
A.~Rosinol, A.~Violette, M.~Abate, N.~Hughes, Y.~Chang, J.~Shi, A.~Gupta, and L.~Carlone, ``Kimera: From slam to spatial perception with 3d dynamic scene graphs,'' \emph{The International Journal of Robotics Research}, vol.~40, no. 12-14, pp. 1510--1546, 2021.

\bibitem{zhang2021building}
Y.~Zhang, G.~Tian, X.~Shao, S.~Liu, M.~Zhang, and P.~Duan, ``Building metric-topological map to efficient object search for mobile robot,'' \emph{IEEE Transactions on Industrial Electronics}, vol.~69, no.~7, pp. 7076--7087, 2021.

\bibitem{hughes2024foundations}
N.~Hughes, Y.~Chang, S.~Hu, R.~Talak, R.~Abdulhai, J.~Strader, and L.~Carlone, ``Foundations of spatial perception for robotics: Hierarchical representations and real-time systems,'' \emph{The International Journal of Robotics Research}, vol.~43, no.~10, pp. 1457--1505, 2024.

\bibitem{wald2020learning}
J.~Wald, H.~Dhamo, N.~Navab, and F.~Tombari, ``Learning 3d semantic scene graphs from 3d indoor reconstructions,'' in \emph{Proceedings of the IEEE/CVF Conference on Computer Vision and Pattern Recognition}, 2020, pp. 3961--3970.

\bibitem{Schmid2024Khronos}
L.~Schmid, M.~Abate, Y.~Chang, and L.~Carlone, ``Khronos: A unified approach for spatio-temporal metric-semantic slamin dynamic environments,'' in \emph{Proc. of Robotics: Science and Systems}, 2024.

\bibitem{yang2020graduated}
H.~Yang, P.~Antonante, V.~Tzoumas, and L.~Carlone, ``Graduated non-convexity for robust spatial perception: From non-minimal solvers to global outlier rejection,'' \emph{IEEE Robotics and Automation Letters}, vol.~5, no.~2, pp. 1127--1134, 2020.

\bibitem{gtsam}
\BIBentryALTinterwordspacing
F.~Dellaert and G.~Contributors, ``borglab/gtsam,'' May 2022. [Online]. Available: \url{https://github.com/borglab/gtsam)}
\BIBentrySTDinterwordspacing

\bibitem{maggio2024clio}
D.~Maggio, Y.~Chang, N.~Hughes, M.~Trang, D.~Griffith, C.~Dougherty, E.~Cristofalo, L.~Schmid, and L.~Carlone, ``Clio: Real-time task-driven open-set 3d scene graphs,'' \emph{IEEE Robotics and Automation Letters}, 2024.

\bibitem{zhao2023fast}
X.~Zhao, W.~Ding, Y.~An, Y.~Du, T.~Yu, M.~Li, M.~Tang, and J.~Wang, ``Fast segment anything,'' \emph{arXiv preprint arXiv:2306.12156}, 2023.

\bibitem{radford2021learning}
A.~Radford, J.~W. Kim, C.~Hallacy, A.~Ramesh, G.~Goh, S.~Agarwal, G.~Sastry, A.~Askell, P.~Mishkin, J.~Clark \emph{et~al.}, ``Learning transferable visual models from natural language supervision,'' in \emph{International conference on machine learning}.\hskip 1em plus 0.5em minus 0.4em\relax PmLR, 2021, pp. 8748--8763.

\bibitem{hoang2020object}
D.-C. Hoang, A.~J. Lilienthal, and T.~Stoyanov, ``Object-rpe: Dense 3d reconstruction and pose estimation with convolutional neural networks,'' \emph{Robotics and Autonomous Systems}, vol. 133, p. 103632, 2020.

\bibitem{kabalar2023towards}
J.~Kabalar, S.-C. Wu, J.~Wald, K.~Tateno, N.~Navab, and F.~Tombari, ``Towards long-term retrieval-based visual localization in indoor environments with changes,'' \emph{IEEE Robotics and Automation Letters}, vol.~8, no.~4, pp. 1975--1982, 2023.

\bibitem{bochinski2017high}
E.~Bochinski, V.~Eiselein, and T.~Sikora, ``High-speed tracking-by-detection without using image information,'' in \emph{2017 14th IEEE international conference on advanced video and signal based surveillance (AVSS)}.\hskip 1em plus 0.5em minus 0.4em\relax IEEE, 2017, pp. 1--6.

\bibitem{grinvald2019volumetric}
M.~Grinvald, F.~Furrer, T.~Novkovic, J.~J. Chung, C.~Cadena, R.~Siegwart, and J.~Nieto, ``Volumetric instance-aware semantic mapping and 3d object discovery,'' \emph{IEEE Robotics and Automation Letters}, vol.~4, no.~3, pp. 3037--3044, 2019.

\bibitem{antonello2018multi}
M.~Antonello, D.~Wolf, J.~Prankl, S.~Ghidoni, E.~Menegatti, and M.~Vincze, ``Multi-view 3d entangled forest for semantic segmentation and mapping,'' in \emph{2018 IEEE International Conference on Robotics and Automation (ICRA)}.\hskip 1em plus 0.5em minus 0.4em\relax IEEE, 2018, pp. 1855--1862.

\bibitem{nakajima_fast_2018}
\BIBentryALTinterwordspacing
Y.~Nakajima, K.~Tateno, F.~Tombari, and H.~Saito, ``Fast and {Accurate} {Semantic} {Mapping} through {Geometric}-based {Incremental} {Segmentation},'' in \emph{2018 {IEEE}/{RSJ} {International} {Conference} on {Intelligent} {Robots} and {Systems} ({IROS})}, Oct. 2018, pp. 385--392, iSSN: 2153-0866. [Online]. Available: \url{https://ieeexplore.ieee.org/document/8593993/}
\BIBentrySTDinterwordspacing

\bibitem{qian2022pocd}
J.~Qian, V.~Chatrath, J.~Yang, J.~Servos, A.~P. Schoellig, and S.~L. Waslander, ``Pocd: Probabilistic object-level change detection and volumetric mapping in semi-static scenes,'' \emph{arXiv preprint arXiv:2205.01202}, 2022.

\bibitem{lu2024semantics}
L.~Lu, Y.~Zhang, P.~Zhou, J.~Qi, Y.~Pan, C.~Fu, and J.~Pan, ``Semantics-aware receding horizon planner for object-centric active mapping,'' \emph{IEEE Robotics and Automation Letters}, vol.~9, no.~4, pp. 3838--3845, 2024.

\bibitem{bao2022semantic}
Y.~Bao, Z.~Yang, Y.~Pan, and R.~Huan, ``Semantic-direct visual odometry,'' \emph{IEEE Robotics and Automation Letters}, vol.~7, no.~3, pp. 6718--6725, 2022.

\bibitem{engel2014lsd}
J.~Engel, T.~Sch{\"o}ps, and D.~Cremers, ``Lsd-slam: Large-scale direct monocular slam,'' in \emph{European conference on computer vision}.\hskip 1em plus 0.5em minus 0.4em\relax Springer, 2014, pp. 834--849.

\bibitem{li2023textslam}
B.~Li, D.~Zou, Y.~Huang, X.~Niu, L.~Pei, and W.~Yu, ``Textslam: Visual slam with semantic planar text features,'' \emph{IEEE Transactions on Pattern Analysis and Machine Intelligence}, vol.~46, no.~1, pp. 593--610, 2023.

\bibitem{li2019semantic}
J.~Li, D.~Meger, and G.~Dudek, ``Semantic mapping for view-invariant relocalization,'' in \emph{2019 International Conference on Robotics and Automation (ICRA)}.\hskip 1em plus 0.5em minus 0.4em\relax IEEE, 2019, pp. 7108--7115.

\bibitem{kuhn1955hungarian}
H.~W. Kuhn, ``The hungarian method for the assignment problem,'' \emph{Naval research logistics quarterly}, vol.~2, no. 1-2, pp. 83--97, 1955.

\bibitem{mu2016slam}
B.~Mu, S.-Y. Liu, L.~Paull, J.~Leonard, and J.~P. How, ``Slam with objects using a nonparametric pose graph,'' in \emph{2016 IEEE/RSJ International Conference on Intelligent Robots and Systems (IROS)}.\hskip 1em plus 0.5em minus 0.4em\relax IEEE, 2016, pp. 4602--4609.

\bibitem{zhang2021fairmot}
Y.~Zhang, C.~Wang, X.~Wang, W.~Zeng, and W.~Liu, ``Fairmot: On the fairness of detection and re-identification in multiple object tracking,'' \emph{International journal of computer vision}, vol. 129, no.~11, pp. 3069--3087, 2021.

\bibitem{reynolds2015gaussian}
D.~Reynolds, ``Gaussian mixture models,'' in \emph{Encyclopedia of biometrics}.\hskip 1em plus 0.5em minus 0.4em\relax Springer, 2015, pp. 827--832.

\bibitem{wang2021dsp}
J.~Wang, M.~R{\"u}nz, and L.~Agapito, ``Dsp-slam: Object oriented slam with deep shape priors,'' in \emph{2021 International Conference on 3D Vision (3DV)}.\hskip 1em plus 0.5em minus 0.4em\relax IEEE, 2021, pp. 1362--1371.

\bibitem{wu2023object}
Y.~Wu, Y.~Zhang, D.~Zhu, Z.~Deng, W.~Sun, X.~Chen, and J.~Zhang, ``An object slam framework for association, mapping, and high-level tasks,'' \emph{IEEE Transactions on Robotics}, vol.~39, no.~4, pp. 2912--2932, 2023.

\bibitem{strecke2019fusion}
M.~Strecke and J.~Stuckler, ``Em-fusion: Dynamic object-level slam with probabilistic data association,'' in \emph{Proceedings of the IEEE/CVF International Conference on Computer Vision}, 2019, pp. 5865--5874.

\bibitem{parkison2018semantic}
S.~A. Parkison, L.~Gan, M.~G. Jadidi, and R.~M. Eustice, ``Semantic iterative closest point through expectation-maximization.'' in \emph{BMVC}, vol.~1, 2018, p.~2.

\bibitem{zhang2019hierarchical}
J.~Zhang, M.~Gui, Q.~Wang, R.~Liu, J.~Xu, and S.~Chen, ``Hierarchical topic model based object association for semantic slam,'' \emph{IEEE transactions on visualization and computer graphics}, vol.~25, no.~11, pp. 3052--3062, 2019.

\bibitem{iqbal2018localization}
A.~Iqbal and N.~R. Gans, ``Localization of classified objects in slam using nonparametric statistics and clustering,'' in \emph{2018 IEEE/RSJ International Conference on Intelligent Robots and Systems (IROS)}.\hskip 1em plus 0.5em minus 0.4em\relax IEEE, 2018, pp. 161--168.

\bibitem{shekhar2012word}
R.~Shekhar and C.~Jawahar, ``Word image retrieval using bag of visual words,'' in \emph{2012 10th IAPR International Workshop on Document Analysis Systems}.\hskip 1em plus 0.5em minus 0.4em\relax IEEE, 2012, pp. 297--301.

\bibitem{merrill2019calc2}
N.~Merrill and G.~Huang, ``Calc2. 0: Combining appearance, semantic and geometric information for robust and efficient visual loop closure,'' in \emph{2019 IEEE/RSJ International Conference on Intelligent Robots and Systems (IROS)}.\hskip 1em plus 0.5em minus 0.4em\relax IEEE, 2019, pp. 4554--4561.

\bibitem{arshad2021semantics}
S.~Arshad and G.-W. Kim, ``Semantics aware loop closure detection in visual slam,'' in \emph{2021 21st International Conference on Control, Automation and Systems (ICCAS)}.\hskip 1em plus 0.5em minus 0.4em\relax IEEE, 2021, pp. 21--24.

\bibitem{hu2019loop}
M.~Hu, S.~Li, J.~Wu, J.~Guo, H.~Li, and X.~Kang, ``Loop closure detection for visual slam fusing semantic information,'' in \emph{2019 Chinese Control Conference (CCC)}.\hskip 1em plus 0.5em minus 0.4em\relax IEEE, 2019, pp. 4136--4141.

\bibitem{yuan2021sv}
Z.~Yuan, K.~Xu, B.~Deng, X.~Zhou, P.~Chen, and Y.~Ma, ``Sv-loop: Semantic-visual loop closure detection with panoptic segmentation,'' in \emph{2021 IEEE 6th International Conference on Signal and Image Processing (ICSIP)}.\hskip 1em plus 0.5em minus 0.4em\relax IEEE, 2021, pp. 245--250.

\bibitem{tsintotas2018seqslam}
K.~A. Tsintotas, L.~Bampis, S.~Rallis, and A.~Gasteratos, ``Seqslam with bag of visual words for appearance based loop closure detection,'' in \emph{International Conference on Robotics in Alpe-Adria Danube Region}.\hskip 1em plus 0.5em minus 0.4em\relax Springer, 2018, pp. 580--587.

\bibitem{papapetros2023semantic}
I.~T. Papapetros, K.~M. Oikonomou, I.~Kansizoglou, K.~A. Tsintotas, and A.~Gasteratos, ``Semantic-based visual vocabulary for loop closure detection,'' in \emph{2023 IEEE International Conference on Imaging Systems and Techniques (IST)}.\hskip 1em plus 0.5em minus 0.4em\relax IEEE, 2023, pp. 1--5.

\bibitem{yu2023semanticloop}
J.~Yu and S.~Shen, ``Semanticloop: Loop closure with 3d semantic graph matching,'' \emph{IEEE Robotics and Automation Letters}, vol.~8, no.~2, pp. 568--575, 2023.

\bibitem{qian2022towards}
Z.~Qian, J.~Fu, and J.~Xiao, ``Towards accurate loop closure detection in semantic slam with 3d semantic covisibility graphs,'' \emph{IEEE Robotics and Automation Letters}, vol.~7, no.~2, pp. 2455--2462, 2022.

\bibitem{xiao2023semantic}
D.~Xiao, S.~Li, and Z.~Xuanyuan, ``Semantic loop closure detection for intelligent vehicles using panoramas,'' \emph{IEEE Transactions on Intelligent Vehicles}, vol.~8, no.~10, pp. 4395--4405, 2023.

\bibitem{kim2025semantic}
J.~Kim and J.~Kim, ``Semantic loop closure for reducing false matches in slam,'' in \emph{2025 22nd International Conference on Ubiquitous Robots (UR)}.\hskip 1em plus 0.5em minus 0.4em\relax IEEE, 2025, pp. 34--39.

\bibitem{singh2021hierarchical}
G.~Singh, M.~Wu, S.-K. Lam, and D.~Van~Minh, ``Hierarchical loop closure detection for long-term visual slam with semantic-geometric descriptors,'' in \emph{2021 IEEE International Intelligent Transportation Systems Conference (ITSC)}.\hskip 1em plus 0.5em minus 0.4em\relax IEEE, 2021, pp. 2909--2916.

\bibitem{chen2021semantic}
H.~Chen, G.~Zhang, and Y.~Ye, ``Semantic loop closure detection with instance-level inconsistency removal in dynamic industrial scenes,'' \emph{IEEE Transactions on Industrial Informatics}, vol.~17, no.~3, pp. 2030--2040, 2021.

\bibitem{wang2025prior}
Y.~Wang, W.~Bai, Z.~Zhang, H.~Wang, H.~Sun, and Q.~Cao, ``Prior-slam: Enabling visual slam for loop closure under large viewpoint variations,'' \emph{IEEE Transactions on Robotics}, 2025.

\bibitem{agrawal2022slam}
A.~Agrawal, D.~Agarwal, M.~Arora, R.~Mahajan, S.~Beohar, L.~Kenye, and R.~Kala, ``Slam and map learning using hybrid semantic graph optimization,'' in \emph{2022 30th Mediterranean Conference on Control and Automation (MED)}.\hskip 1em plus 0.5em minus 0.4em\relax IEEE, 2022, pp. 731--736.

\bibitem{yang2020d3vo}
N.~Yang, R.~Wang, J.~Stueckler, and D.~Cremers, ``D3vo: Deep depth, deep pose and deep uncertainty for monocular visual odometry,'' \emph{Proceedings of the IEEE/CVF Conference on Computer Vision and Pattern Recognition}, pp. 1281--1292, 2020.

\bibitem{wimbauer2021monorec}
F.~Wimbauer, N.~Yang, L.~Von~Stumberg, N.~Zeller, and D.~Cremers, ``Monorec: Semi-supervised dense reconstruction in dynamic environments from a single moving camera,'' in \emph{Proceedings of the IEEE/CVF Conference on Computer Vision and Pattern Recognition}, 2021, pp. 6112--6122.

\bibitem{radwan2018vlocnet++}
N.~Radwan, A.~Valada, and W.~Burgard, ``Vlocnet++: Deep multitask learning for semantic visual localization and odometry,'' \emph{IEEE Robotics and Automation Letters}, vol.~3, no.~4, pp. 4407--4414, 2018.

\bibitem{kim2020simvodis}
U.-H. Kim, S.-H. Kim, and J.-H. Kim, ``Simvodis: Simultaneous visual odometry, object detection, and instance segmentation,'' \emph{IEEE Transactions on Pattern Analysis and Machine Intelligence}, vol.~44, no.~1, pp. 428--441, 2020.

\bibitem{kim2022simvodis++}
------, ``Simvodis++: Neural semantic visual odometry in dynamic environments,'' \emph{IEEE Robotics and Automation Letters}, vol.~7, no.~2, pp. 4244--4251, 2022.

\bibitem{laina2025findanything}
S.~B. Laina, S.~Boche, S.~Papatheodorou, S.~Schaefer, J.~Jung, and S.~Leutenegger, ``Findanything: Open-vocabulary and object-centric mapping for robot exploration in any environment,'' \emph{arXiv preprint arXiv:2504.08603}, 2025.

\bibitem{li2024resolving}
H.~Li, S.~Yu, S.~Zhang, and G.~Tan, ``Resolving loop closure confusion in repetitive environments for visual slam through ai foundation models assistance,'' in \emph{2024 IEEE International Conference on Robotics and Automation (ICRA)}.\hskip 1em plus 0.5em minus 0.4em\relax IEEE, 2024, pp. 6657--6663.

\bibitem{tosi2024nerfs}
F.~Tosi, Y.~Zhang, Z.~Gong, E.~Sandstr{\"o}m, S.~Mattoccia, M.~R. Oswald, and M.~Poggi, ``How nerfs and 3d gaussian splatting are reshaping slam: a survey,'' \emph{arXiv preprint arXiv:2402.13255}, vol.~4, p.~1, 2024.

\bibitem{mildenhall2021nerf}
B.~Mildenhall, P.~P. Srinivasan, M.~Tancik, J.~T. Barron, R.~Ramamoorthi, and R.~Ng, ``Nerf: Representing scenes as neural radiance fields for view synthesis,'' \emph{Communications of the ACM}, vol.~65, no.~1, pp. 99--106, 2021.

\bibitem{czarnowski2020deepfactors}
J.~Czarnowski, T.~Laidlow, R.~Clark, and A.~J. Davison, ``Deepfactors: Real-time probabilistic dense monocular slam,'' \emph{IEEE Robotics and Automation Letters}, vol.~5, no.~2, pp. 721--728, 2020.

\bibitem{kong2023vmap}
X.~Kong, S.~Liu, M.~Taher, and A.~J. Davison, ``vmap: Vectorised object mapping for neural field slam,'' in \emph{Proceedings of the IEEE/CVF Conference on Computer Vision and Pattern Recognition}, 2023, pp. 952--961.

\bibitem{han2023ro}
X.~Han, H.~Liu, Y.~Ding, and L.~Yang, ``Ro-map: Real-time multi-object mapping with neural radiance fields,'' \emph{IEEE Robotics and Automation Letters}, vol.~8, no.~9, pp. 5950--5957, 2023.

\bibitem{kruzhkov2022meslam}
E.~Kruzhkov, A.~Savinykh, P.~Karpyshev, M.~Kurenkov, E.~Yudin, A.~Potapov, and D.~Tsetserukou, ``Meslam: Memory efficient slam based on neural fields,'' in \emph{2022 IEEE International Conference on Systems, Man, and Cybernetics (SMC)}.\hskip 1em plus 0.5em minus 0.4em\relax IEEE, 2022, pp. 430--435.

\bibitem{wang2023co}
H.~Wang, J.~Wang, and L.~Agapito, ``Co-slam: Joint coordinate and sparse parametric encodings for neural real-time slam,'' in \emph{Proceedings of the IEEE/CVF Conference on Computer Vision and Pattern Recognition}, 2023, pp. 13\,293--13\,302.

\bibitem{johari2023eslam}
M.~M. Johari, C.~Carta, and F.~Fleuret, ``Eslam: Efficient dense slam system based on hybrid representation of signed distance fields,'' in \emph{Proceedings of the IEEE/CVF conference on computer vision and pattern recognition}, 2023, pp. 17\,408--17\,419.

\bibitem{xiang2023nisb}
B.~Xiang, Y.~Sun, Z.~Xie, X.~Yang, and Y.~Wang, ``Nisb-map: Scalable mapping with neural implicit spatial block,'' \emph{IEEE Robotics and Automation Letters}, vol.~8, no.~8, pp. 4761--4768, 2023.

\bibitem{rosinol2023nerf}
A.~Rosinol, J.~J. Leonard, and L.~Carlone, ``Nerf-slam: Real-time dense monocular slam with neural radiance fields,'' in \emph{2023 IEEE/RSJ International Conference on Intelligent Robots and Systems (IROS)}.\hskip 1em plus 0.5em minus 0.4em\relax IEEE, 2023, pp. 3437--3444.

\bibitem{matsuki2023imode}
H.~Matsuki, E.~Sucar, T.~Laidow, K.~Wada, R.~Scona, and A.~J. Davison, ``imode: Real-time incremental monocular dense mapping using neural field,'' in \emph{2023 IEEE International Conference on Robotics and Automation (ICRA)}.\hskip 1em plus 0.5em minus 0.4em\relax IEEE, 2023, pp. 4171--4177.

\bibitem{zhu2024nicer}
Z.~Zhu, S.~Peng, V.~Larsson, Z.~Cui, M.~R. Oswald, A.~Geiger, and M.~Pollefeys, ``Nicer-slam: Neural implicit scene encoding for rgb slam,'' in \emph{2024 International Conference on 3D Vision (3DV)}.\hskip 1em plus 0.5em minus 0.4em\relax IEEE, 2024, pp. 42--52.

\bibitem{liso2024loopy}
L.~Liso, E.~Sandstr{\"o}m, V.~Yugay, L.~Van~Gool, and M.~R. Oswald, ``Loopy-slam: Dense neural slam with loop closures,'' in \emph{Proceedings of the IEEE/CVF conference on computer vision and pattern recognition}, 2024, pp. 20\,363--20\,373.

\bibitem{li2024gs3lam}
L.~Li, L.~Zhang, Z.~Wang, and Y.~Shen, ``Gs3lam: Gaussian semantic splatting slam,'' in \emph{Proceedings of the 32nd ACM International Conference on Multimedia}, 2024, pp. 3019--3027.

\bibitem{liu2025sdd}
H.~Liu, L.~Wang, H.~Luo, F.~Zhao, R.~Chen, Y.~Chen, M.~Xiao, J.~Yan, and D.~Luo, ``Sdd-slam: Semantic-driven dynamic slam with gaussian splatting,'' \emph{IEEE Robotics and Automation Letters}, 2025.

\bibitem{xin2025large}
Z.~Xin, C.~Wu, P.~Huang, Y.~Zhang, Y.~Mao, and G.~Huang, ``Large-scale gaussian splatting slam,'' \emph{arXiv preprint arXiv:2505.09915}, 2025.

\bibitem{yugay2025gaussian}
V.~Yugay, T.~Kersten, L.~Carlone, T.~Gevers, M.~R. Oswald, and L.~Schmid, ``Gaussian mapping for evolving scenes,'' \emph{arXiv preprint arXiv:2506.06909}, 2025.

\bibitem{cui2024neural}
J.~Cui, J.~Zhang, L.~Kneip, and S.~Schwertfeger, ``Neural surfel reconstruction: Addressing loop closure challenges in large-scale 3d neural scene mapping,'' \emph{Sensors (Basel, Switzerland)}, vol.~24, no.~21, p. 6919, 2024.

\bibitem{chaplot2020neural}
D.~S. Chaplot, R.~Salakhutdinov, A.~Gupta, and S.~Gupta, ``Neural topological slam for visual navigation,'' in \emph{Proceedings of the IEEE/CVF conference on computer vision and pattern recognition}, 2020, pp. 12\,875--12\,884.

\bibitem{tchuiev2020distributed}
V.~Tchuiev and V.~Indelman, ``Distributed consistent multi-robot semantic localization and mapping,'' \emph{IEEE Robotics and Automation Letters}, vol.~5, no.~3, pp. 4649--4656, 2020.

\bibitem{hu2023cosar}
K.~Hu, L.~Zhan, L.~Zou, Y.~Han, T.~Bi, and G.-M. Muntean, ``Cosar: Multi-robot collaborative semantic mapping over wireless networks,'' in \emph{2023 IEEE International Symposium on Broadband Multimedia Systems and Broadcasting (BMSB)}.\hskip 1em plus 0.5em minus 0.4em\relax IEEE, 2023, pp. 1--6.

\bibitem{liu2022active}
X.~Liu, A.~Prabhu, F.~Cladera, I.~D. Miller, L.~Zhou, C.~J. Taylor, and V.~Kumar, ``Active metric-semantic mapping by multiple aerial robots,'' \emph{arXiv preprint arXiv:2209.08465}, 2022.

\bibitem{asgharivaskasi_riemannian_2025}
\BIBentryALTinterwordspacing
A.~Asgharivaskasi, F.~Girke, and N.~Atanasov, ``\BIBforeignlanguage{en}{Riemannian {Optimization} for {Active} {Mapping} {With} {Robot} {Teams}},'' \emph{\BIBforeignlanguage{en}{IEEE Transactions on Robotics}}, vol.~41, pp. 1077--1097, 2025. [Online]. Available: \url{https://ieeexplore.ieee.org/document/10829726/}
\BIBentrySTDinterwordspacing

\bibitem{aguilar2025multi}
G.~Aguilar, I.~Becerra, and R.~Murrieta-Cid, ``Multi-robot exploration and semantic map building: Heterogeneous terrestrial robots and a drone,'' \emph{Inteligencia Artificial}, vol.~28, no.~76, pp. 166--185, 2025.

\bibitem{dai2017scannet}
A.~Dai, A.~X. Chang, M.~Savva, M.~Halber, T.~Funkhouser, and M.~Nie{\ss}ner, ``Scannet: Richly-annotated 3d reconstructions of indoor scenes,'' in \emph{Proceedings of the IEEE conference on computer vision and pattern recognition}, 2017, pp. 5828--5839.

\bibitem{Matterport3D}
A.~Chang, A.~Dai, T.~Funkhouser, M.~Halber, M.~Niessner, M.~Savva, S.~Song, A.~Zeng, and Y.~Zhang, ``Matterport3d: Learning from rgb-d data in indoor environments,'' \emph{International Conference on 3D Vision (3DV)}, 2017.

\bibitem{Phil2018active}
P.~Ammirato, A.~C. Berg, and J.~Košecká, ``Active vision dataset benchmark,'' in \emph{2018 IEEE/CVF Conference on Computer Vision and Pattern Recognition Workshops (CVPRW)}, 2018, pp. 2127--21\,273.

\bibitem{geiger2012cvpr}
A.~Geiger, P.~Lenz, and R.~Urtasun, ``{Are we ready for Autonomous Driving? The KITTI Vision Benchmark Suite},'' in \emph{Proc.~of the IEEE Conf.~on Computer Vision and Pattern Recognition (CVPR)}, 2012, pp. 3354--3361.

\bibitem{Menze2015CVPR}
M.~Menze and A.~Geiger, ``Object scene flow for autonomous vehicles,'' in \emph{Conference on Computer Vision and Pattern Recognition (CVPR)}, 2015.

\bibitem{Liao2022PAMI}
Y.~Liao, J.~Xie, and A.~Geiger, ``{KITTI}-360: A novel dataset and benchmarks for urban scene understanding in 2d and 3d,'' \emph{Pattern Analysis and Machine Intelligence (PAMI)}, 2022.

\bibitem{caesar2020nuscenes}
H.~Caesar, V.~Bankiti, A.~H. Lang, S.~Vora, V.~E. Liong, Q.~Xu, A.~Krishnan, Y.~Pan, G.~Baldan, and O.~Beijbom, ``nuscenes: A multimodal dataset for autonomous driving,'' in \emph{Proceedings of the IEEE/CVF conference on computer vision and pattern recognition}, 2020, pp. 11\,621--11\,631.

\bibitem{liu2023active}
X.~Liu, A.~Prabhu, F.~Cladera, I.~D. Miller, L.~Zhou, C.~J. Taylor, and V.~Kumar, ``Active metric-semantic mapping by multiple aerial robots,'' \emph{arXiv preprint arXiv:2209.08465}, 2023.

\bibitem{tao20243d}
Y.~Tao, X.~Liu, I.~Spasojevic, S.~Agarwal, and V.~Kumar, ``3d active metric-semantic slam,'' \emph{IEEE Robotics and Automation Letters}, vol.~9, no.~3, pp. 2989--2996, 2024.

\bibitem{georgakis2021learning}
G.~Georgakis, B.~Bucher, K.~Schmeckpeper, S.~Singh, and K.~Daniilidis, ``Learning to map for active semantic goal navigation,'' \emph{arXiv preprint arXiv:2106.15648}, 2022.

\bibitem{tian2024rasls}
C.~Tian, S.~Tian, Y.~Kang, H.~Wang, J.~Tie, and S.~Xu, ``Rasls: Reinforcement learning active slam approach with layout semantic,'' in \emph{2024 International Joint Conference on Neural Networks (IJCNN)}.\hskip 1em plus 0.5em minus 0.4em\relax IEEE, 2024, pp. 1--8.

\bibitem{ravichandran2022hierarchical}
Z.~Ravichandran, L.~Peng, N.~Hughes, J.~D. Griffith, and L.~Carlone, ``Hierarchical representations and explicit memory: Learning effective navigation policies on 3d scene graphs using graph neural networks,'' in \emph{2022 International Conference on Robotics and Automation (ICRA)}.\hskip 1em plus 0.5em minus 0.4em\relax IEEE, 2022, pp. 9272--9279.

\bibitem{baxter2020toward}
D.~P. Baxter \emph{et~al.}, ``Toward robust active semantic slam via max-mixtures,'' Ph.D. dissertation, Massachusetts Institute of Technology, 2020.

\bibitem{vodisch2022continual}
N.~V{\"o}disch, D.~Cattaneo, W.~Burgard, and A.~Valada, ``Continual slam: Beyond lifelong simultaneous localization and mapping through continual learning,'' in \emph{The international symposium of robotics research}.\hskip 1em plus 0.5em minus 0.4em\relax Springer, 2022, pp. 19--35.

\bibitem{li2024learn}
B.~Li, Z.~Yan, D.~Wu, H.~Jiang, and H.~Zha, ``Learn to memorize and to forget: A continual learning perspective of dynamic slam,'' in \emph{European Conference on Computer Vision}.\hskip 1em plus 0.5em minus 0.4em\relax Springer, 2024, pp. 41--57.

\bibitem{vodisch2023covio}
N.~V{\"o}disch, D.~Cattaneo, W.~Burgard, and A.~Valada, ``Covio: Online continual learning for visual-inertial odometry,'' in \emph{Proceedings of the ieee/cvf conference on computer vision and pattern recognition}, 2023, pp. 2464--2473.

\bibitem{huang2025bye}
C.~Huang, S.~Yan, and W.~Burgard, ``Bye: Build your encoder with one sequence of exploration data for long-term dynamic scene understanding,'' \emph{IEEE Robotics and Automation Letters}, 2025.

\end{thebibliography}


\section*{Biography Section}

\vspace{11pt}

\vspace{-33pt}
\begin{IEEEbiography}[{\includegraphics[width=1in,height=1.25in,clip,keepaspectratio]{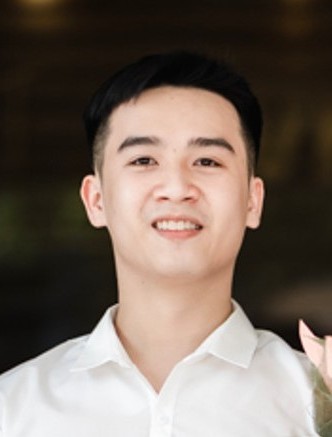}}]{Thanh Nguyen Canh} (\textit{Graduate Student Member,~IEEE}) received his Engineering Degree in Robotics Engineering at the University of Engineering and Technology, Viet Nam National University (VNU) in 2022, and his M.Sc. degree from Japan Advanced Institute of Science and Technology (JAIST), Japan in 2024. Currently, he is pursuing a Ph.D. degree at JAIST. His research interests include Robotics, SLAM, AI, and robot vision.
\end{IEEEbiography}

\begin{IEEEbiography}[{\includegraphics[width=1in,height=1.25in,clip,keepaspectratio]{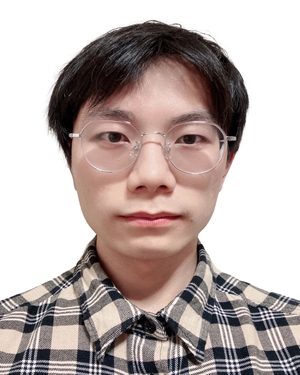}}]{Haolan Zhang} received the B.Sc. degree in automation engineering from Dalian Polytechnic University, Dalian, China, in 2021 and his M.Sc. Degree from Gunma University, Kiryu, Japan in 2024. Currently, he is pursuing a Ph.D. degree at Japan Advanced Institute of Science and Technology (JAIST), Japan. His current research interests include SLAM, Computer Vision.
\end{IEEEbiography}

\begin{IEEEbiography}[{\includegraphics[width=1in,height=1.25in,clip,keepaspectratio]{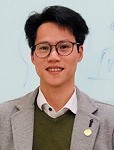}}]{Xiem HoangVan} is an Associate Professor at the Faculty of Electronics and Telecommunications, VNU-University of Engineering and Technology, Vietnam. He received his Ph.D. degree from Lisbon University, Portugal, in 2015, his M.Sc. degree from Sungkyunkwan University, South Korea, in 2011, all in Electrical and Computer Engineering.  His research interests are machine learning, image, and video communications. Prof. Xiem has published about 100 papers on robotics, image, and video processing and regularly reviews for many renowned IEEE, IET, and EURASIP journals and serves as a technical committee member for international conferences and funding agencies worldwide. 
 
 He has received several technical awards for his contributions to image and video coding, including 5 Best Paper awards, i.e., at Picture Coding Symposium 2015 (Australia), the International Workshop on Advanced Image Technology 2018 (Thailand), REC-ECIT 2022, IEEE-RIVE 2023, and ATC 2024. He is a recipient of the Fraunhofer Portugal award 2015, and  Golden Globe Award for Young Scientists (under 35 years old) in Science and Technology 2019, and the VNU Top Young Scientist Award 2019. 
\end{IEEEbiography}

\vspace{-33pt}
\begin{IEEEbiography}[{\includegraphics[width=1in,height=1.25in,clip,keepaspectratio]{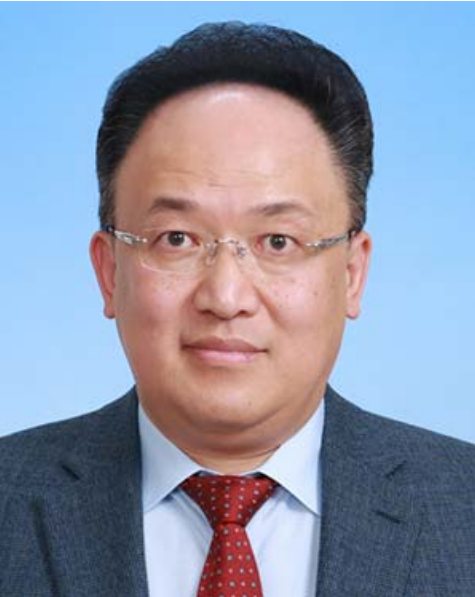}}]{Nak Young Chong} (\textit{Senior Member,~IEEE}) received the B.S., M.S., and Ph.D. degrees in mechanical engineering from Hanyang University, Seoul, Korea, in 1987, 1989, and 1994, respectively. From 1994 to 2003, he was with Daewoo Heavy Industries, Geoje,  Korea; Korea Institute of Science and Technology, Seoul, Korea; and Mechanical Engineering Laboratory and National Institute of Advanced Industrial Science and Technology, Tsukuba, Japan. In 2003, he joined a Faculty Member at Japan Advanced Institute of Science and Technology, Ishikawa, Japan, where he served as a Councilor, Director of the Center for Intelligent Robotics, and the Chair Professor of Intelligent Robotics Group, and is currently a Professor of information science. He was a Visiting Scholar with Northwestern University, Evanston, IL, USA; Georgia Institute of Technology, GA, USA; University of Genoa, Genoa, Italy; and Carnegie Mellon University, Pittsburgh, PA, USA, and serves/served as an Associate Faculty with the University of Nevada, Las Vegas, NV, USA; Kyung Hee University, Yongin, Korea; and Hanyang University, Ansan, Korea. He served as the Program (Co)-chair for JCK Robotics 2009, ICAM 2010, IEEE Ro-Man 2011/2013/2022, IEEE CASE 2012, URAI 2013/2014, DARS 2014, ICCAS 2016, and IEEE ARM 2019. He was the General (Co)-chair of URAI 2017 and UR 2020. He also served as the President of Korea Robotics Society, the Co-chair for IEEE RAS Networked Robots TC, and the Fujitsu Scientific System WG. He serves/served as a Senior Editor for \textit{IEEE Robotics and Automation Letters}, \textit{Intelligent Service Robotics}, and \textit{International Journal of Advanced Robotic Systems}; and as an Associate Editor for IEEE T{\scriptsize RANSACTIONS ON} R{\scriptsize OBOTICS}. 
\end{IEEEbiography}

\vfill

\end{document}